%% file: main.tex
\documentclass[twoside,11pt]{fairmeta_seamless}
\newif\ifdraft
\draftfalse

\usepackage[T1, T2A]{fontenc} 
\usepackage{hyperref}       
\usepackage{url}            
\usepackage{nicefrac}       
\usepackage{multicol}
\usepackage{microtype}      
\usepackage{geometry}
\usepackage{graphicx}%
\graphicspath{{figures/}}
\usepackage{multirow}%
\usepackage{amsmath}
\usepackage{amssymb}
\usepackage{amsfonts}
\usepackage{stmaryrd}
\usepackage{mathrsfs}%
\usepackage{xcolor}%
\usepackage{textcomp}%
\usepackage{manyfoot}%
\usepackage{booktabs}%
\usepackage{algorithm}%
\usepackage{tabularx}%
\usepackage{algorithmicx}%
\usepackage{algpseudocode}%
\usepackage{listings}%
\usepackage{algpseudocode}
\usepackage[inline]{enumitem}
\usepackage{xspace}
\usepackage{tikz}
\usepackage{makecell}
\usepackage{longtable}
\usepackage{changepage}
\usepackage{color}
\usepackage{colortbl}
\usepackage{multirow}
\usepackage{pifont}
\usepackage{subcaption}
\usepackage{physics}
\usepackage[russian,english]{babel}

\usepackage{setspace}
\usepackage{changepage} 
\usepackage{float}

\usepackage{tablefootnote}
\usepackage{threeparttable}
\usepackage{placeins}

\input{macros}

\ifdraft
    \newcommand{\hongyu}[1]{\textcolor{blue}{[Hongyu: #1]}}
    \newcommand{\juan}[1]{\textcolor{green}{[Juan: #1]}}
    \newcommand{\TODO}[1]{\textcolor{red}{TODO: #1}}
    \newcommand{\todo}[1]{%
      \begin{yellowtodo}
        #1
      \end{yellowtodo}
    }
\else
    \newcommand{\hongyu}[1]{}
    \newcommand{\juan}[1]{}
    \newcommand{\TODO}[1]{}
    \newcommand{\todo}[1]{}
\fi

\author[]{Vasu Agrawal}
\author[]{Akinniyi Akinyemi}
\author[]{Kathryn Alvero}
\author[*]{Morteza Behrooz}
\author[]{Julia Buffalini}
\author[]{Fabio Maria Carlucci}
\author[]{Joy Chen}
\author[]{Junming Chen}
\author[]{Zhang Chen}
\author[]{Shiyang Cheng}
\author[]{Praveen Chowdary}
\author[]{Joe Chuang}
\author[]{Antony D'Avirro}
\author[]{Jon Daly}
\author[*]{Ning Dong}
\author[]{Mark Duppenthaler}
\author[]{Cynthia Gao}
\author[\dagger]{Jeff Girard}
\author[]{Martin Gleize}
\author[]{Sahir Gomez}
\author[]{Hongyu Gong}
\author[]{Srivathsan Govindarajan}
\author[]{Brandon Han}
\author[]{Sen He}
\author[]{Denise Hernandez}
\author[]{Yordan Hristov}
\author[]{Rongjie Huang}
\author[]{Hirofumi Inaguma}
\author[]{Somya Jain}
\author[]{Raj Janardhan}
\author[]{Qingyao Jia}
\author[]{Christopher Klaiber}
\author[]{Dejan Kovachev}
\author[]{Moneish Kumar}
\author[]{Hang Li}
\author[]{Yilei Li}
\author[]{Pavel Litvin}
\author[]{Wei Liu}
\author[]{Guangyao Ma}
\author[]{Jing Ma}
\author[]{Martin Ma}
\author[]{Xutai Ma}
\author[]{Lucas Mantovani}
\author[]{Sagar Miglani}
\author[]{Sreyas Mohan}
\author[]{Louis-Philippe Morency}
\author[]{Evonne Ng}
\author[]{Kam-Woh Ng}
\author[]{Tu Anh Nguyen}
\author[]{Amia Oberai}
\author[]{Benjamin Peloquin}
\author[]{Juan Pino}
\author[]{Jovan Popovi\'{c}}
\author[]{Omid Poursaeed}
\author[]{Fabian Prada}
\author[]{Alice Rakotoarison}
\author[]{Rakesh Ranjan}
\author[]{Alexander Richard}
\author[]{Christophe Ropers}
\author[]{Safiyyah Saleem}
\author[*]{Vasu Sharma}
\author[]{Alex Shcherbyna}
\author[]{Jie Shen}
\author[]{Anastasis Stathopoulos}
\author[]{Anna Sun}
\author[]{Paden Tomasello}
\author[]{Tuan Tran}
\author[]{Arina Turkatenko}
\author[]{Bo Wan}
\author[]{Chao Wang}
\author[]{Jeff Wang}
\author[]{Mary Williamson}
\author[]{Carleigh Wood}
\author[]{Tao Xiang}
\author[]{Yilin Yang}
\author[]{Zhiyuan Yao}
\author[]{Chen Zhang}
\author[]{Jiemin Zhang}
\author[]{Xinyue Zhang}
\author[]{Jason Zheng}
\author[]{Pavlo Zhyzheria}
\author[*]{Jan Zikes}
\author[]{Michael Zollhoefer}

\affiliation[]{Meta}
\affiliation[\dagger]{University of Kansas}
\affiliation[*]{Work was done when the author was affiliated with Meta.}
\contribution[]{Authors are listed in alphabetical order.}

\correspondence{Louis-Philippe Morency at \email{lpmorency@meta.com}, Hongyu Gong at \email{hygong@meta.com}}
\github{\url{https://github.com/facebookresearch/seamless_interaction}}
\hf{\url{https://huggingface.co/datasets/facebook/seamless-interaction}}

\title{Seamless Interaction: Dyadic Audiovisual Motion Modeling and Large-Scale Dataset}

\abstract{
Human communication involves a complex interplay of verbal and nonverbal signals, essential for conveying meaning and achieving interpersonal goals. To develop socially intelligent AI technologies, it is crucial to develop models that can both comprehend and generate dyadic behavioral dynamics. To this end, we introduce the \mosaic Dataset, a large-scale collection of over 4,000 hours of face-to-face interaction footage from over 4,000 participants in diverse contexts. This dataset enables the development of AI technologies that understand dyadic embodied dynamics, unlocking breakthroughs in virtual agents, telepresence experiences, and multimodal content analysis tools. We also develop a suite of models that utilize the dataset to generate dyadic motion gestures and facial expressions aligned with human speech. These models can take as input both the speech and visual behavior of their interlocutors. We present a variant with speech from an LLM model and integrations with 2D and 3D rendering methods, bringing us closer to interactive virtual agents. Additionally, we describe controllable variants of our motion models that can adapt emotional responses and expressivity levels, as well as generating more semantically-relevant gestures. Finally, we discuss methods for assessing the quality of these dyadic motion models, which are demonstrating the potential for more intuitive and responsive human-AI interactions.
}

\begin{document}

\maketitle

{\hypersetup{hidelinks} 
\clearpage
\setcounter{tocdepth}{2}
\tableofcontents
\clearpage
}

\input{intro}

\input{data/arxiv}
\input{repr/arxiv}
\input{motion_model/arxiv}

\input{llama_integration/arxiv}
\input{visualization/arxiv}
\input{eval/arxiv}

\input{rai/arxiv}
\input{related_work/arxiv}
\input{conclusion}

\input{ack}

\bibliography{bibliography}
\newpage

\input{appendix/arxiv}

\end{document}

%% file: macros.tex
\newcolumntype{H}{>{\setbox0=\hbox\bgroup}c<{\egroup}@{}}
\newcolumntype{P}[1]{>{\centering\arraybackslash}p{#1}}

\usepackage{xcolor}
\usepackage{tcolorbox}
\newtcolorbox{yellowtodo}[1][]{
  colback=yellow!20!white,
  colframe=yellow!80!black,
  fonttitle=\bfseries,
  title=TODO,
  #1
}

\newcommand{\mosaic}{\textsc{Seamless Interaction}\xspace}
\newcommand{\Naturalistic}{\textit{Naturalistic}\xspace}
\newcommand{\Improvised}{\textit{Improvised}\xspace}

\definecolor{alizarin}{rgb}{0.82, 0.1, 0.26}
\definecolor{asparagus}{rgb}{0.53, 0.66, 0.42}
\definecolor{cerulean}{rgb}{0.0, 0.48, 0.65}
\newcommand{\greencheck}{{\color{asparagus}\ding{51}}}
\newcommand{\redxmark}{{\color{alizarin}\ding{55}}}

\newcommand{\silero}{\textsc{Silero VAD}\xspace}
\newcommand{\whisperx}{\textsc{WhisperX}\xspace}
\newcommand{\hamer}{HaMeR\xspace}
\newcommand{\hmr}{HMR\xspace}
\newcommand{\smpl}{SMPL\hyp{}H\xspace}

\newcommand{\adapter}{Adapter\xspace}

%% file: intro.tex
\section{Introduction}
\label{sec:intro}

Human face-to-face communication is a dynamic and intricate dance in which individuals continuously adapt their verbal and nonverbal behaviors to convey meaning and pursue interpersonal goals. Successfully navigating such interactions requires an individual to skillfully produce and interpret a complex blend of signals---words, tone, gestures, posture, and more---within the ever-changing context of the conversation. To develop AI technologies that are socially intelligent and effective, it is essential to build foundation models that understand the \textit{dyadic embodied dynamics} of social interactions---essentially, the interactive and physical nuances exchanged between pairs of individuals in conversation. By capturing these subtleties, we can unlock new technological breakthroughs, such as intuitive and responsive virtual agents, immersive and naturalistic telepresence experiences in AR/VR settings, and sophisticated multimodal content analysis tools.

Achieving this goal requires large-scale, meticulously crafted datasets that include synchronized, multimodal recordings of face-to-face interactions across diverse contexts. The \mosaic dataset meets this need by providing over 4,000 hours of dyadic footage from more than 4,000 participants, accompanied by extensive metadata. To capture the full spectrum of communication styles and interpersonal goals, interaction prompts were designed based on a categories rooted in contemporary psychological theory, and two distinct types of interactions were recorded. First, to reflect naturalistic communication, prompts related to common and comfortable interaction styles were given to pairs of untrained research participants. Second, to explore rare and challenging interaction styles, trained actors improvised a subset of the prompts while drawing from their own relevant experiences. The dataset not only facilitates the training of advanced computational models but also offers methodologies to evaluate the quality of generated dyadic interactions both objectively, through automatically computed metrics, and subjectively, via human judgments.

In addition to describing the \mosaic dataset, we present a suite of research models that use this data to understand and generate dyadic embodied dynamics. 
Our primary model processes \textit{dyadic audio} (i.e., audio from both participants in a conversation) and generates corresponding facial expressions and body movements for each avatar. By considering both participants simultaneously, our models learn to generate not only naturalistic speaking behaviors (e.g., relevant hand gestures), but also appropriate listening behaviors (e.g., timely head nods and smiles). We also present the capability of adding the visual behavior of one person as input to the dyadic motion model to generate the facial and body motion of the other person's avatar, highlighting the potential of avatars to react intelligently to visual cues, e.g., with smile mirroring and joint gaze attention. 

\Cref{tab:model_comparison} compares capabilities of our proposed modeling approach with related work on motion generation. Previous work has explored the capability of \textit{monadic audio-driven} (aka, audio2motion) where speech of one speaker can drive the animation of the avatar. \textit{Dyadic audio-driven} goes beyond that by taking input of two audio streams (e.g., a podcast of two people talking) and generates not only speaking gestures, but also listening and turn-taking cues. \textit{Reactivity to visual input} means that the motion model is also reactive the visual behaviors of the other interlocutors, such as smile mirroring or mutual gaze. While most prior work either generates face motion or body motion separately (usually, only one of them), \textit{Face + body modeling} means that the model is able to synchronize the face and body motion. Finally, \textit{2D and 3D renderings} means that our proposed model is able to generate intermediate representations (aka, codes) that can be rendered either in 2d videos, or in 3D (in our case, building from Meta's Codec Avatar 3D rendering technologies).

\begin{table}[h]
\centering
\resizebox{\textwidth}{!}{
\begin{tabular}{lccccc}
\toprule
 & \begin{tabular}[c]{@{}c@{}}\textbf{Monadic} \\ \textbf{audio-driven}\end{tabular} & \begin{tabular}[c]{@{}c@{}}\textbf{Dyadic} \\ \textbf{audio-driven}\end{tabular} & \begin{tabular}[c]{@{}c@{}}\textbf{Reactivity to} \\ \textbf{visual input} \end{tabular} & \begin{tabular}[c]{@{}c@{}} \textbf{Face+Body} \\ \textbf{modeling}\end{tabular} & \begin{tabular}[c]{@{}c@{}}\textbf{2D and 3D} \\ \textbf{renderings}\end{tabular} \\ \midrule
VASA-1 \citep{vasa} & \greencheck & \redxmark & \redxmark & \redxmark & \redxmark \\
HeyGen \citep{heygen} & \greencheck & \redxmark & \redxmark & \redxmark & \redxmark \\
OmniHuman-1 \citep{omnihuman} & \greencheck & \redxmark & \redxmark & \greencheck & \redxmark \\
IFNP \citep{zhu2025infp} & \greencheck & \greencheck & \redxmark & \redxmark & \redxmark \\
ConvoFusion \citep{mughal2024convofusion} & \greencheck & \greencheck & \redxmark & \redxmark & \redxmark \\
Veo 3 \citep{veo3} & \redxmark & \redxmark & \redxmark & \greencheck & \redxmark \\ 
Ours & \greencheck & \greencheck & \greencheck & \greencheck & \greencheck \\ \bottomrule
\end{tabular}}
\caption{Comparing capabilities of our proposed modeling approach with other related work.}
\label{tab:model_comparison}
\end{table}

Finally, we describe controllable variants of our motion models that can adapt emotional responses and expressivity levels, as well as generating more semantically-relevant gestures. When using speech input from a speech LLM model, we can leverage the speech LLM model's ability to grasp conversational context, enabling the extra controllability of generation of gestures and expressions that suit each moment of the conversation. We refer to this last approach as \textit{LLM-guided codebook generation}. All our proposed Dyadic Motion Models output intermediate representations (aka, ``codes''), one for face and another for body pose, which can be used to render either 2D videos and 3D Codec Avatars.

%% file: data/arxiv.tex
\section{\mosaic Dataset}

The \mosaic dataset is a comprehensive collection of in-person, dyadic interactions designed to advance research in social AI. As released, it includes over 4,000 hours of interactions from more than 4,000 participants, featuring nearly 1,300 conversational and activity-based prompts. The dataset is anchored in contemporary psychological theory and includes a wide range of conversational topics, interpersonal stances, and participant relationships across both \Naturalistic and \Improvised content (defined below). Rich contextualization is provided to interactions via detailed annotations and metadata.

This work continues a long tradition of high-quality, human-centric, and multi-modal datasets, such as the Switchboard \citep{godfrey1992switchboard} and Fisher \citep{cieri2004fisher} speech-based conversational corpora and the AMI Meeting \citep{carletta2005ami}, IEMOCAP \citep{busso2008iemocap}, and CANDOR \citep{candor} audiovisual interaction datasets (see comparison with related work in Table \ref{tab:data_comparison}). Critically, however, our new dataset innovates in several key aspects:

\paragraph{In-person recordings.} By collecting the vast majority of interactions in-person, we preserve the natural dynamics of face-to-face communication. Indeed, previous research has found that remote interactions (e.g., audio and video conferencing) often differ substantially from in-person interactions in terms of turn-taking \citep{tian2024Corpus}, gaze patterns \citep{horstmann2022perception}, and feelings of closeness and satisfaction \citep{mallen2003Online}. In-person recordings also allow us to ensure the quality of recording equipment and conditions, as well as to avoid technical artifacts caused by hardware and network latency.

\paragraph{Various relationship types.} The history shared between two people (or the lack thereof) is a key contextual factor influencing their interactions \citep[e.g.,][]{knapp2013,hopwood2020}. For instance, familiar dyads tend to use more nonverbal cues and shared language and may be more comfortable with overlapping speech, deeper conversational topics, and collaborative conflict resolution strategies. In contrast, strangers tend to rely more on verbal communication and formal language and may follow more structured turn-taking rules, stick to surface-level topics, and use more avoidant conflict resolution strategies. By including unfamiliar dyads as well as several types of familiar dyads (e.g., friends, family, partners, coworkers), we are able to capture a wider range of relational contexts and further reveal their influence on dyadic communication.


\paragraph{Naturalistic and Improvised content.} Our goal was to capture a comprehensive sample of behaviors that accurately represents the full spectrum of interpersonal interactions. We used contemporary psychological theory to guide our representation of this spectrum. However, certain behaviors crucial for social AI systems to understand are infrequent when interactions are recorded in a laboratory setting. This rarity can be attributed to these behaviors being less socially accepted (e.g., bullying), more context-specific (e.g., competitive negotiation tactics), or associated with mental health challenges (e.g., paranoia), making them challenging to replicate in a lab environment. Some behaviors would also be ethically problematic to expose participants to in order to gauge their genuine reactions. To address this challenge, we include both untrained participants to complete more common and comfortable \Naturalistic interactions and recruit professional actors with improvisational experience to safely complete more rare and challenging \Improvised interactions. The majority of \Improvised interaction involve two actors. A minority involve a single actor interacting with a non-actor.

The \mosaic dataset was collected from June 2024 to May 2025 in partnership with vendors at various locations across the United States, spanning six states and ten cities (see \Cref{appendix:collection_geo}). Our aim is to catalyze the next generation of research into dyadic interaction and social AI.

\begin{table}[h]
\centering
\resizebox{\textwidth}{!}{
\begin{tabular}{lccccccc}
\toprule
 & \begin{tabular}[c]{@{}c@{}}\textbf{Volume} \\ \textbf{(hours)}\end{tabular} & \begin{tabular}[c]{@{}c@{}}\textbf{Participants} \\ \textbf{(total unique)}\end{tabular} & \begin{tabular}[c]{@{}c@{}}\textbf{Audio-}\\ \textbf{visual}\end{tabular} & \textbf{In-person} & \begin{tabular}[c]{@{}c@{}}\textbf{High quality} \\ \textbf{face+body}\end{tabular} & \begin{tabular}[c]{@{}c@{}}$\mathbf{\ge2}$\\ \textbf{participants}\end{tabular} & \textbf{Annotations} \\ \midrule
Ours & 4,065 & 4,284 & \greencheck & \greencheck & \greencheck & \greencheck & \greencheck \\
CANDOR \citep{candor} & 850 & 1,454 & \greencheck & \redxmark & \redxmark & \greencheck & \greencheck \\
AMI \citep{carletta2005ami} & ~100 & ~100 & \greencheck & \greencheck & \greencheck & \greencheck & \greencheck\\
BEAT \citep{liu2022beatlargescalesemanticemotional} & 76 & 30 & \greencheck & \greencheck & \greencheck & \greencheck & \greencheck \\
IEMOCAP \citep{busso2008iemocap} & 12.5 & 10 & \greencheck & \greencheck & \greencheck & \greencheck & \greencheck \\ 
InterAct \citep{huang2024interactcapturemodellingrealistic} & ~10 & ~400 & \greencheck & \greencheck & \greencheck & \greencheck & \redxmark \\ 
Talking with hands \citep{lee2019talking} & ~50 & ~50 & \greencheck & \greencheck & \greencheck & \greencheck & \redxmark \\ 
MELD \citep{poria2018meld} & ~13 & ~250 & \greencheck & \greencheck & \greencheck & \greencheck & \greencheck \\ 
Ted Talks \citep{siarohin2021motionrepresentationsarticulatedanimation} & ~100 & ~3,000 & \greencheck & \redxmark & \redxmark & \redxmark & \redxmark \\ 
CMU MOSEI \citep{bagher-zadeh-etal-2018-multimodal} & ~70 & ~1,000 & \greencheck & \redxmark & \redxmark & \redxmark & \greencheck \\ 
DyConv \citep{zhu2024infpaudiodriveninteractivehead} & ~200 & -- & \greencheck & \redxmark & \redxmark & \greencheck & \redxmark \\ 
Switchboard \citep{godfrey1992switchboard} & 300 & 543 & \redxmark & \redxmark & \greencheck & \greencheck & \greencheck \\ 
Fisher (Part 1\&2) \citep{cieri2004fisher} & 2,000 & 3,000 & \redxmark & \redxmark & \greencheck & \greencheck & \greencheck \\ 
\bottomrule
\end{tabular}}
\caption{Dataset characteristics in comparison with existing work. Counts and volumes are either as directly reported in papers or estimated from relevant reported quantities. "Volume" for dyadic or multi-party collection indicates "conversation" or "interaction" time, rather than total footage. "In-person" means the participants are in the same room. "High quality face+body" indicates that complete information of all participants (face and body) is maintained throughout all recordings. "Annotations" broadly indicates additional human-labeled data.}
\label{tab:data_comparison}
\end{table}

\begin{table}[h]
    \centering
    \begin{tabular}{lccccc}
        \toprule
        \textbf{Part} & \textbf{Hours} & \textbf{Interactions} & \textbf{Sessions} & \textbf{Participants} & \textbf{Prompts} \\
        \midrule
        Overall & 4,065.04 & 64,739 & 5,098 & 4,284 & 1,283 \\
        Naturalistic & 2,745.43 & 47,333 & 3,363 & 3,525 & 675 \\
        Improvised & 1,319.61 & 17,406 & 1,735 & 1,011 & 845 \\
        \bottomrule
    \end{tabular}
    \caption{Overall corpus statistics and major partitions (\Naturalistic and \Improvised.)}
    \label{tab:stats_overall}
\end{table}

The number of dyadic hours, interactions, sessions, participants, and prompts in the dataset is summarized in \Cref{tab:stats_overall} (overall and in the \Naturalistic and \Improvised portions). Beyond its large volume of dyadic interactions, a principle strength of the \mosaic dataset is the accompanying metadata for each interaction. For a given interaction we have:
\begin{itemize}
    \item Prompt text presented to participants (see example in \Cref{tab:prompt_example_naturalistic}).
    \item Interaction type information: one of the three \textit{games/activities} or IPC-based conversations (see \Cref{section:activites_prompts}).
    \item Personality information for a subset of participants (see \Cref{section:personality}).
    \item Relationship information of the participants in a given session (see \Cref{section:participant_relationship}).
    \item Annotations of \textit{Internal State}, \textit{Internal State Rationale} and \textit{Visual behavior} for a subset of interactions (see \Cref{section:annotations}).
\end{itemize}

\subsection{Interactions, Sessions, Scripts}
\label{section:interactions_session_scripts}
As released, the \mosaic dataset is a collection of nearly 70,000 short, 2-10 min long interactions between two people. This design allows us to collect a large variety of conversational topics and interpersonal stances - by breaking up larger 1-hr long recording sessions between dyads into thousands of shorter interactions. This approach naturally allows for continuity between interactions over the course of the 1-hr long recording session for a given dyad, while also orienting participants towards a range of different topics and implied socio-affective states.

\paragraph{Sessions.} Each dyadic session is 1-hr long, capturing the behavior of two participants conversing and engaging in activities. A trained moderator guides each session, signaling the start and end of each interaction, providing clarification and encouragement to the participants (see moderator responsibilities in \Cref{subsubsec:moderators}).

\paragraph{Interactions.} A given session is divided into between 5-20 (2-10 min long) interactions. An interaction is made up of  \textit{active time} - any period during which the participants are speaking, acting, or otherwise behaving in direct response to a prompt from the script. Non-interaction time, also referred to as \textit{meta time}, is any period during which participants are reading the prompts, engaging with the moderator, being distracted, or otherwise have "broken character" from the prompt and are discussing aspects of the session, dataset, or situation itself. While there is some degree of fuzziness in the boundaries of \textit{active} and \textit{meta time}, either due to the nature of the interaction or due to time-stamping error, a given session contains alternating \textit{active} and \textit{meta time} chronologically with regard to the prompts. On average each 1-hr session contains approximately 80-90\% \textit{active time}. The current release of the \mosaic dataset contains only \textit{active} time segments. 

\begin{figure}[t]
    \centering
    \includegraphics[width=0.9\linewidth]{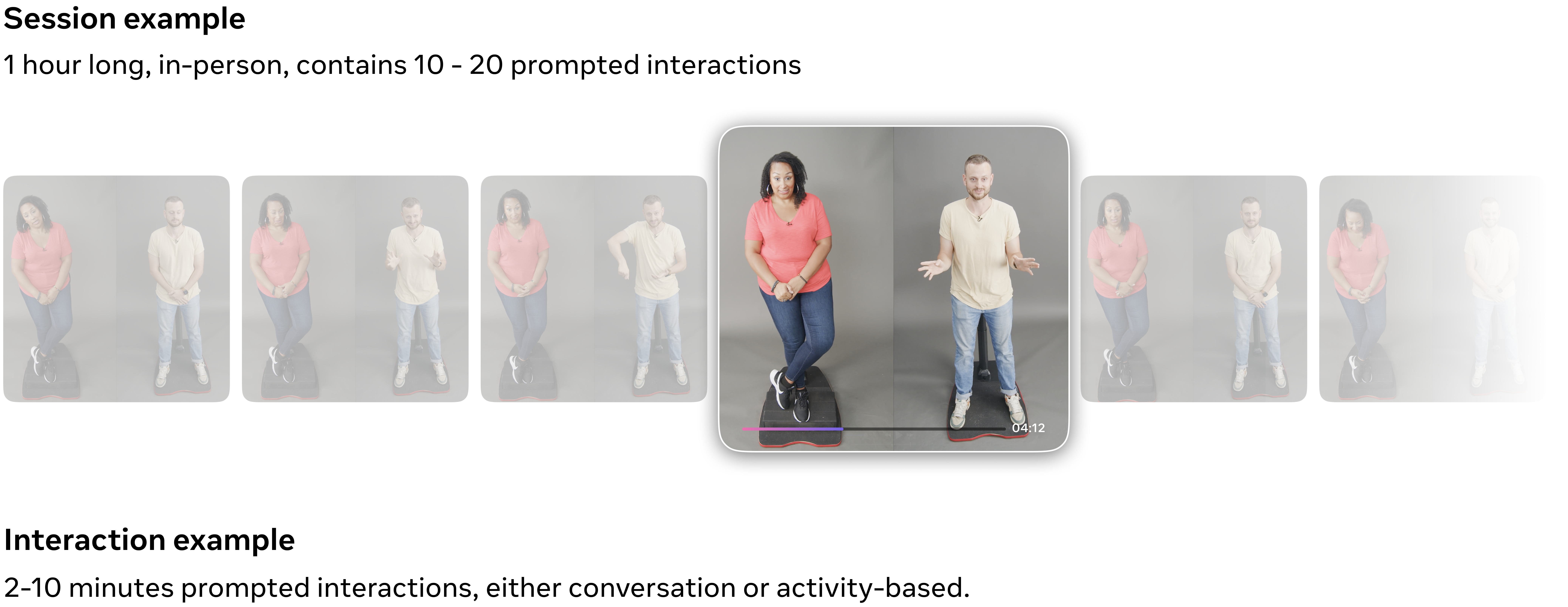}
    \caption{An illustration of how a \textit{Session} is divided into 10-20 shorter interactions.}
    \label{fig:session-example}
\end{figure}
\begin{figure}[t]
    \centering
    \includegraphics[width=0.9\linewidth]{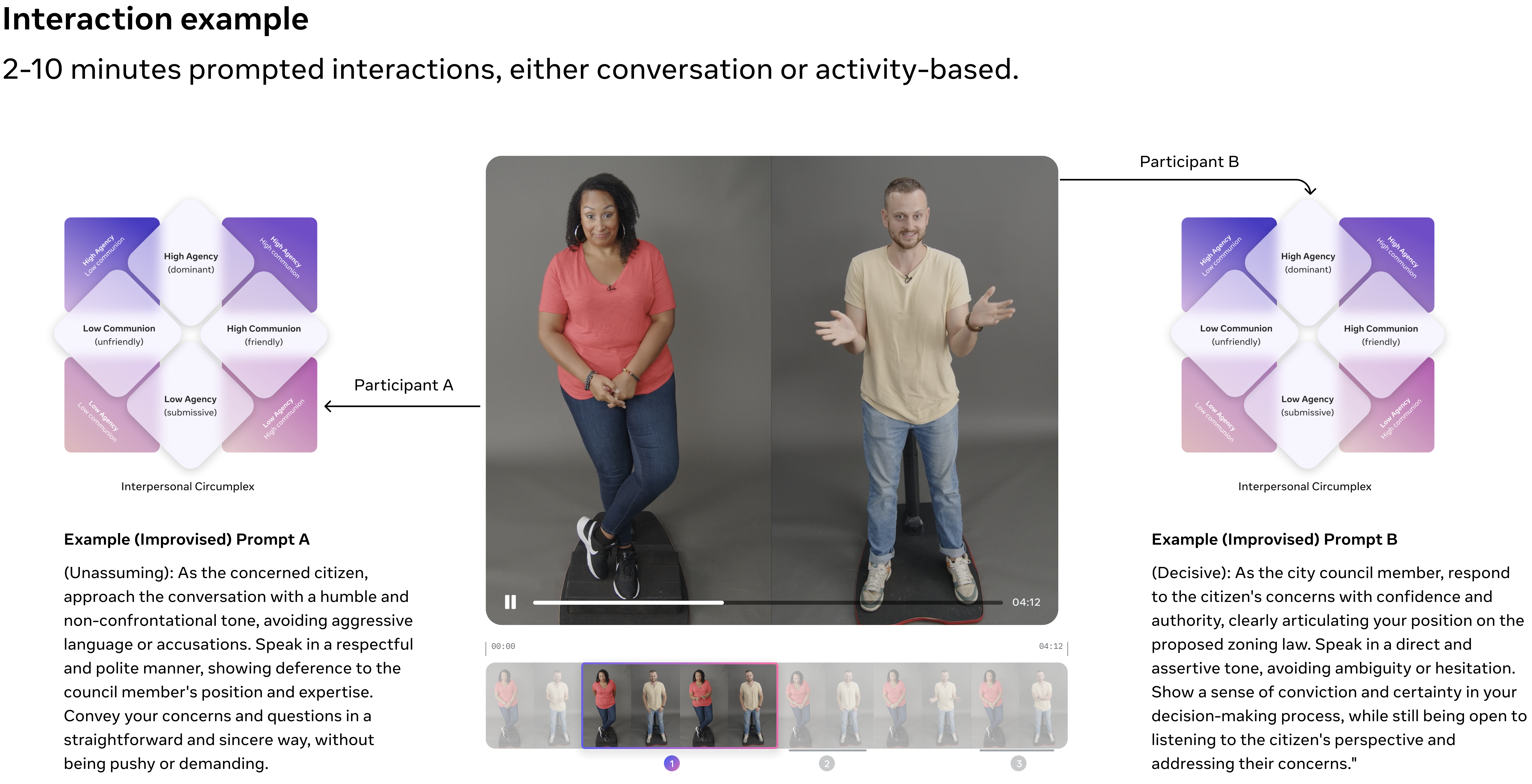}
    \caption{An illustration of how an \textit{Interaction} is anchored in the IPC via prompts.}
    \label{fig:interaction-example}
\end{figure}

\paragraph{Scripts.} The collection of interactions that make up a session are sampled uniformly across the IPC octant space. A particular organization of interaction prompts is called a \textit{Script}, where each \textit{Script} is tailored to whether the dyad belongs to the \Naturalistic or \Improvised part, and within the \Naturalistic pairs, whether they are familiar or stranger.

For every interaction both participants are given separate prompts. The prompts are curated according to whether it is a \Naturalistic or \Improvised session, the relationship (familiar or stranger), and each is designed to anchor the interaction in the IPC \citep{pincus2013,wright2023}.

\subsection{Building Instruction Prompts with Interpersonal Theory}
\label{section:taxonomy_and_ipc}

Contemporary Integrative Interpersonal Theory (CIIT) refines over 70 years of multidisciplinary research on human interaction and relationships; it has wide relevance and influence across social, medical, and organizational settings, as well as substantial empirical support \citep{pincus2013,wright2023}. It holds that essential features of personality and mental health are reflected in interpersonal situations, including face-to-face and digital interactions, as well as mental representations (e.g., memories, expectations, and fantasies) of such interactions. Furthermore, it proposes that these features can be concisely organized using different combinations of the dimensions of agency and communion. 

\begin{figure}[t]
\begin{center}
\includegraphics[width=0.55\textwidth]{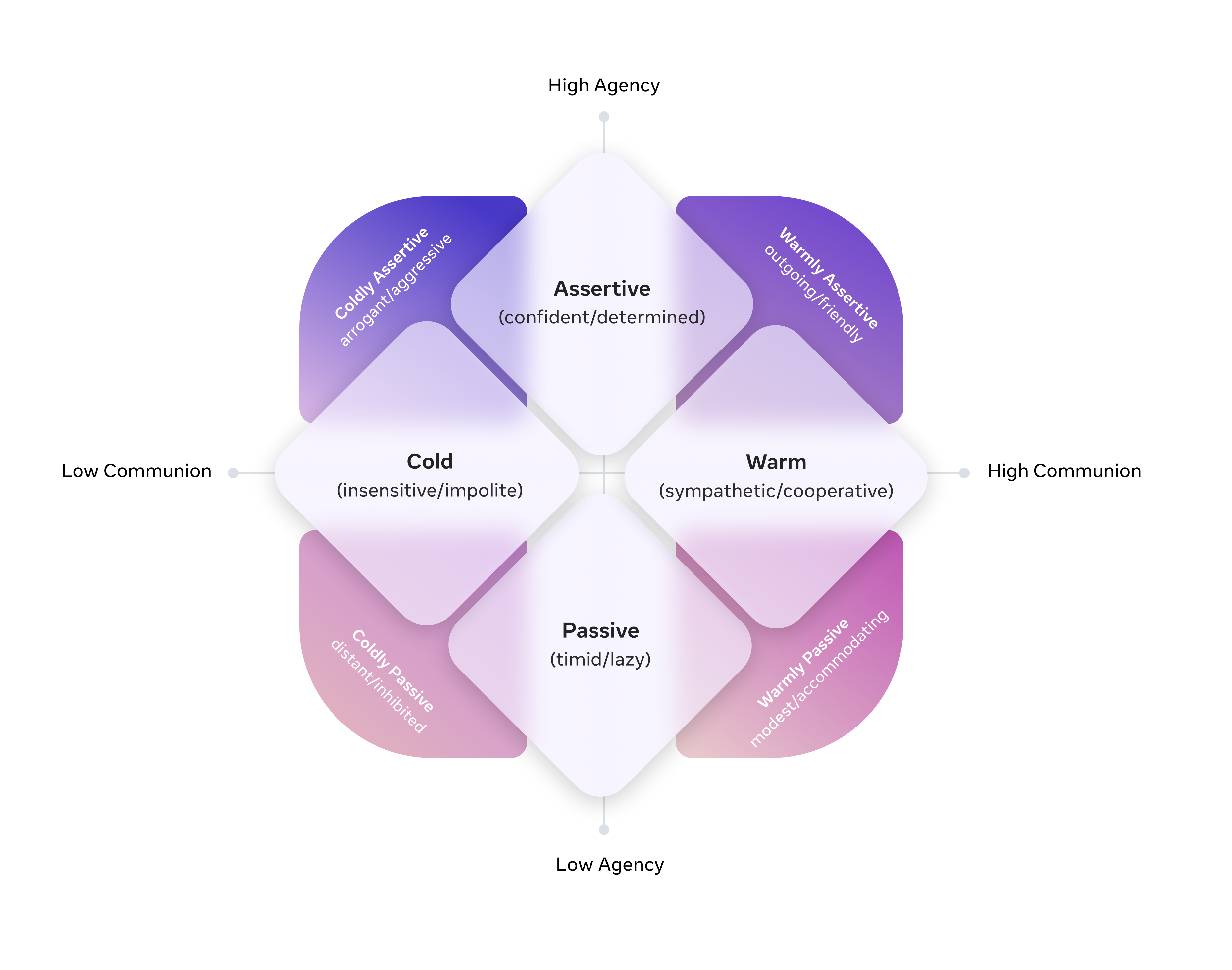}
\end{center}
\caption{The Interpersonal Circumplex (IPC) and its two dimensions of \textit{Agency} and \textit{Communion}.}
\label{fig:ipc}
\end{figure}

\textit{Agency} refers to the development of a distinct sense of self, highlighting goals related to achievement, influence, and independence.  \textit{Communion} refers to the formation of close relationships with others, emphasizing goals related to belonging, intimacy, and nurture. Together, agency and communion provide a nuanced lens through which interpersonal dynamics can be analyzed and understood, offering insights into the motivations and behaviors that characterize human interactions. More specifically, interpersonal phenomena (e.g., traits and goals, strengths and problems, behaviors and perceptions) can be located within a circular structure called the \textit{interpersonal circumplex} (IPC), which is formed around the axes of agency and communion (\autoref{fig:ipc}). 

In the \mosaic dataset, we adopted categories of \textit{interpersonal stances} (i.e., how one approaches a given social interaction based on one's intentions, emotions, and expectations) defined by eight regions of the IPC: high agency and moderate communion (e.g., confident/determined), low agency and moderate communion (e.g., timid/lazy), high communion and moderate agency (e.g., sympathetic/cooperative), low communion and moderate agency (e.g., insensitive/impolite), high agency and high communion (e.g., outgoing/friendly), high agency and low communion (e.g., arrogant/aggressive), low agency and low communion (e.g., distant/inhibited), and low agency and high communion (e.g., modest/accommodating). While not all interpersonal stances are equally common or valued across different cultures and contexts, each is possible and consequential. Using these categories allowed us to more fully capture the spectrum of possible interpersonal interactions. 

CIIT provides testable predictions about the emotional and functional outcomes of interactions between specific interpersonal stances (e.g., encounters are generally smoother when dyadic partners differ on agency but align on communion). While these hypotheses are supported by empirical evidence across various settings \citep[e.g.,][]{hopwood2020}, we are eager to further investigate these predictions using the extensive testbed offered by this new dataset. Similarly, agency and communion have expected relationships with other important psychological phenomena such as personality and emotion  \citep[e.g.,][]{traupman2009,yik2004}.

\subsubsection{Conversation Based Prompt Creation}
\label{section:conversation_prompt_creation}
We attempt to anchor each 2-10 min long dyadic interaction in the IPC via the use of prompts. Prompts are presented separately to each participant in an interaction prior to the start of the interaction. The majority of \textit{meta time} in the dataset (which is not included in this release) is the short intervals between active interactions, during which participants are studying their prompts and asking clarifying questions of the moderators (refer to \Cref{section:interactions_session_scripts} for more detail on \textit{active} and \textit{meta time}).

In the case of \Naturalistic prompts only one of the participants is given a topic anchoring them in the IPC, while the other participant is instructed to behave in whatever way they feel is natural to them. \Improvised (or role-play) prompts, by contrast, provide extensive descriptions for one or both participants. We provide examples of both \Naturalistic (\Cref{tab:prompt_example_naturalistic}) and \Improvised (\Cref{tab:prompt_example_improvised}) prompts below.

\begin{table}[t]
    \centering
    \begin{tabular}{|m{2cm}|m{5cm}|m{5cm}|c|c|}
        \toprule
        \textbf{Situation} & \textbf{Participant A} & \textbf{Participant B} & \textbf{IPC A} & \textbf{IPC B} \\ 
        \midrule
         & Your Partner is going to ask you to describe a situation in which someone else criticized you, but you felt the criticism was unfair (this can be from anytime in your life). What happened and how did it make you feel? & Ask your partner to describe a situation where someone criticized them unfairly. Discuss it with them. & ANCM & - \\ 
         \bottomrule
    \end{tabular}
    \caption{Example Naturalistic prompt pair for an interaction.}
    \label{tab:prompt_example_naturalistic}
\end{table}

\begin{table}[t]
    \centering
    \begin{tabular}{|m{2cm}|m{5cm}|m{5cm}|c|c|}
        \toprule
        \textbf{Situation} & \textbf{Participant A} & \textbf{Participant B} & \textbf{IPC A} & \textbf{IPC B} \\ 
        \midrule
        A concerned citizen speaks with a city council member about proposed zoning & (Unassuming): As the concerned citizen, approach the conversation with a humble and non-confrontational tone, avoiding aggressive language or accusations. Speak in a respectful and polite manner, showing deference to the council member's position and expertise. Convey your concerns and questions in a straightforward and sincere way, without being pushy or demanding. & (Decisive): As the city council member, respond to the citizen's concerns with confidence and authority, clearly articulating your position on the proposed zoning law. Speak in a direct and assertive tone, avoiding ambiguity or hesitation. Show a sense of conviction and certainty in your decision-making process, while still being open to listening to the citizen's perspective and addressing their concerns. & AMCP & APCN \\ 
        \bottomrule
    \end{tabular}
    \caption{Example Improvised prompt pair for an interaction.}
    \label{tab:prompt_example_improvised}
\end{table}

\paragraph{\Naturalistic prompt design.} Naturalistic prompts are designed such that:
\begin{itemize}
    \item They are answerable by the average person
    \item The are general enough to allow for a range of response directions depending on the comfort and preference of the participants.
    \item They provide topic areas that align with a particular IPC octant, but are not prescriptive. 
\end{itemize}

For familiar dyads within the \Naturalistic dataset, prompts are designed such that they follow the \Naturalistic prompts generally, with the addition of occasionally making use of—or reference to—history shared by the interlocutors.

\paragraph{\Improvised prompt design.} Improvised (role-play) prompts are designed such that they:
\begin{itemize}
    \item Require prior experience in acting or in improvisation
    \item Are detailed and prescriptive - describing the roles, stances and emotions that should be represented by the participants.
    \item Are typically preceded by an adjective intended to describe the “overall manner” of the interpersonal stance of for a particular participant. This adjective also provides are mapping back into IPC octants.
\end{itemize}

\paragraph{Human- and model-based prompt creation.} There are three versions of prompts in the \mosaic dataset. The first two versions contain prompts that are hand-written by the core project team. The third and final version of prompts scaled this approach with the use of text-based LLM - prompting a Llama3.1-70b-instruct model to generate prompts geared towards \textit{Naturalist} stranger pairs, \Naturalistic familiar pairs, and \Improvised pairs. 

\subsubsection{Activity Based Interactions}
\label{section:activites_prompts}
In addition to conversation-based prompts anchored in the IPC, we include another category of prompted interactions we call activities. Activities are designed to explore dynamics outside of free-flowing conversation, particularly as they relate to turn-taking, the relation between language and gesture, and the relation between gesture and intent.

\begin{table}[h]
    \centering
    \begin{tabular}{cccccc}
        \toprule
        \textbf{Type} & \textbf{Hours} & \textbf{Interactions} & \textbf{Sessions} & \textbf{Participants} & \textbf{Prompts} \\
        \midrule
        IPC conversation & 3,464.16 & 53,938 & 5,098 & 4,284 & 953 \\
        Language \& Gesture & 379.12 & 6,364 & 3,465 & 3,702 & 296 \\
        Collaborative story-telling & 146.89 & 2,845 & 	2,844 & 3,105 & 1 \\
        Silent Charades & 74.87 & 1,592 & 1,592 & 1,910 & 1 \\
        \bottomrule
    \end{tabular}
    \caption{Data volumes by interaction type.}
    \label{tab:interaction_types}
\end{table}

\Cref{tab:interaction_types} shows data volumes for IPC-based conversation prompts along with three activities - ``Collaborative story-telling'', a ``Language and gesture,'' and a ``Silent charades.'' We provide examples of each of the activity-based prompts in \Cref{section:appendix_activity_based_examples}.

\subsection{Metadata and Annotations}
The \mosaic dataset provides rich contextualization at the level of interactions (prompts and IPC anchoring), individuals (personality), and dyads (relationships). Below, we describe metadata related to the latter two. We reserve a more detailed description of the taxonomy and its relation to prompt creation in \Cref{section:taxonomy_and_ipc}.

\subsubsection{Personality}
\label{section:personality}

The Five Factor Model of Personality \citep{mccrae1999} is one of the most robust and empirically supported theories in psychology \citep{john2008}. It provides a comprehensive framework for understanding how individuals perceive, interpret, and engage with the world, organizing these differences into five broad dimensions known as the Big Five personality traits: agreeableness, conscientiousness, extraversion, neuroticism, and openness. These traits have been shown to predict important real-world outcomes, such as job performance, academic success, and health \citep[e.g.,][]{ozer2006a,roberts2007,soto2019}, often surpassing competing predictors like socioeconomic status and cognitive ability \citep{roberts2007}. 

By shaping affective, cognitive, and social processes, these personality traits profoundly influence dyadic interactions and constitute a critical aspect of the context in which they occur. For example, highly extraverted individuals are often assertive and engaging, whereas highly agreeable individuals tend to favor cooperation and harmony \citep{john1999}. Although less explicitly interpersonal, the other Big Five traits also affect communication styles: openness is associated with creativity and flexibility, conscientiousness with structure and goal-directedness, and neuroticism with emotionality and reactivity \citep{leary2009}.

\begin{figure}
\includegraphics[width=\textwidth]{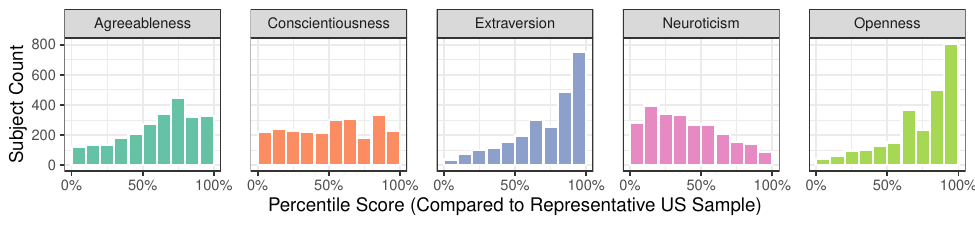}
\caption{Distributions of participants' personality scores from the BFI-2.}
\label{fig:bfi}
\end{figure}

In the \mosaic dataset, we assessed participants' personality traits using the Big Five Inventory--2 (BFI-2), a 60-item questionnaire with robust psychometric properties \citep{soto2017}. Across the 2,260 participants (52\%) who chose to complete the BFI-2, inter-item reliability \citep[i.e., coefficient omega;][]{dunn2014} was ``excellent'' for neuroticism (0.90) and ``good'' for conscientiousness (0.88), extraversion (0.85), openness (0.84), and agreeableness (0.81). \autoref{fig:bfi} depicts each trait's distribution in our sample as compared to the representative US adult samples ($n=3,071$) from \citet{soto2019}. Our participants represented the entire range of all five traits, but higher levels of extraversion and openness were more common.

\subsubsection{Participant Relationship}
\label{section:participant_relationship}

The relationship and the degree of familiarity between the interlocutors accounts for a large degree of variation in the topics, stances, and behaviors that make up an interaction \citep{giles1991accommodation}. \Cref{tab:naturalistic} provides the statistics of dyadic relationships in \Naturalistic data. For recorded collections, such as the \mosaic dataset, having some familiarity with your conversational partner also has the advantage of reducing the need for participants to build rapport or go through an often awkward process of ice-breaking and familiarization.  With these aspects in mind, approximately half of the \Naturalistic dataset dyads were specifically recruited as familiar pairs - participants with some prior knowledge of one another, including friends, family, colleagues and romantic partners. \Cref{tab:relationship_detail} shows the distribution across familiar categories.

\begin{table}[h]
    \centering
    \begin{tabular}{cccccc}
        \toprule
        \textbf{Relation} & \textbf{Hours} & \textbf{Interactions} & \textbf{Sessions} & \textbf{Participants} & \textbf{Prompts} \\
        \midrule
        Familiar & 1,427.61 & 24,205 & 1,752 & 2,231 & 597 \\
        Stranger & 1,317.82 & 23,128 & 1,611 & 1,820 & 	597 \\
        \bottomrule
    \end{tabular}
    \caption{Corpus statistics for the \Naturalistic partition, by dyad relationship.} 
    \label{tab:naturalistic}
\end{table}

\begin{table}[h!]
    \centering
    \begin{tabular}{c|c}
        \hline
        Relationship Detail & Proportion \\
        \hline
        Friends & 0.605 \\
        Coworkers & 0.132 \\
        Family-generic & 0.089 \\
        Familiar-generic & 0.088 \\
        Dating/spouse/romantic partner & 0.056 \\
        Classmates & 0.022 \\
        Siblings & 0.005 \\
        Parent/child & 0.002 \\
        Neighbors & 0.001 \\
        Roommates & 0.001 \\
        \hline
    \end{tabular}
    \caption{Relationship details and proportions, rounded to three decimals}
    \label{tab:relationship_detail}
\end{table}

\subsubsection{Annotations}
\label{section:annotations}
Annotations were performed on a subset of the dataset, to add information about observed visual behaviors, the internal states of the participants, and the possible reasoning behind these visual behaviors. More than 4,500 moment of interest segments were annotated in the dataset. A moment-of-interest was defined as a moment that includes interesting, conspicuous visual behaviors that differ from the normal flow of interaction. To ensure a good representation of the internal state and reasoning of the participant's behaviors, we asked the participants to annotate themselves directly. These first-party (1P) annotations were complemented by a superset of annotations from third-party (3P) annotators. The first-party annotators (the participants themselves) were asked to annotate the internal states (IS) and the internal state rationales (R) for the moment of interest they remembered. In parallel, third-party annotators were also asked to provide information about the internal states (IS) and internal state rationales (R), but were also asked to annotate fine-grained descriptions of the visual behaviors, aka visual elements (V). A detailed description of each annotation type is provided in \Cref{tab:annotation_types}.

The concepts of moment of interest, internal state, internal state rationale, and visual elements are defined as follows:
\begin{itemize}
    \item \textbf{Moment of interest} A short span of time (usually a few seconds) in an interaction during which a participant acts in a way that is visibly or audibly different from what would be considered their baseline behavior. 
    \item \textbf{Internal state} is defined as including emotions, feelings, thoughts, or internal dialogue. The instructions included examples for annotators to help guide their descriptions of their internal states. The word examples were classified for emotions, interpersonal, and cognitive: \subitem \textbf{ Emotions} Following the dimensional representation of emotional state, examples for low and high arousal dimension, as well as positive and negative valence dimension
        \subitem \textbf{Interpersonal} We followed the interpersonal theory previously described in  \Cref{section:taxonomy_and_ipc}, including the eight quadrant along agency and communion dimensions.
        \subitem \textbf{Cognitive} We also included examples of cognitive states that could be observed during interpersonal interactions, including engagement, comprehension and focus.
    \item \textbf{Internal state rationale} is defined as the reason for which a participant's internal state was triggered. The rationale can be the trigger itself, regardless of the intent, or it can be the assumed intent behind the trigger. If we take the same situation as above (i.e., Participant A is talking, Participant B starts yawning, Participant A becomes offended), a description of the internal state rationale could be either of the following:
    \begin{itemize}
        \item \textit{I felt offended} because he|she started yawning and looking bored.
        \item \textit{I felt offended} because I could tell that he|she was no longer paying attention.
    \end{itemize}
    \item \textbf{Visual element} is defined as a description of the gestures, movements, facial expressions, and other behaviors that mark a moment of interest and are not considered baseline behaviors for the participant being viewed.
\end{itemize}

\begin{table}[h]
    \centering
    \begin{tabular}{ |m{1cm}|m{13cm}| }
    \toprule
    \textbf{\small{TYPE}} & \textbf{\small{DESCRIPTION}}\\
    \midrule
    \multicolumn{2}{| c |}{\textbf{First-person (first-party) annotations}} \\
    \midrule
    \small{1P-IS} & \small{Participants annotate their own internal states. Internal states include emotions, feelings, thoughts, or internal dialogue.}\\
    \midrule
    \small{1P-R} & \small{Participants annotate their own behavior rationales or theories of mind. Rationales are the reason(s) for which a participant's internal state was triggered. The rationale can be the trigger itself, regardless of intent, or it can be the assumed intent behind the trigger (or theory of mind).} \\
    \midrule
    \multicolumn{2}{| c |}{\textbf{Third-person (third-party) annotations}}\\
    \midrule
    \small{3P-IS} & \small{Trained annotators annotate the participants' perceived internal states}\\
    \midrule
    \small{3P-R} & \small{Trained annotators annotate the participants' perceived behavior rationales}\\
    \midrule
    \small{3P-V} & \small{Trained annotators annotate visual elements in the participants' non-baseline behaviors} \\
    \bottomrule
    \end{tabular}
    \caption{Description of annotation types.}
    \label{tab:annotation_types}
\end{table}

An example of the annotation workflow can be found in  \Cref{section:annotation-workflow}.

\paragraph{Annotation statistics.}

Table \ref{tab:annotation-stats} shows annotation volumes by type. Note that our process involved providing different types of annotations for a given moment of interest. These volumes reflect the total number of moment of interest segments and durations when treating the annotations independently. In total, we release $n=16,898$ annotations and $n=5,137$ MOI segments covering 4.74 hours of dyadic interaction.

\begin{table}[h]
    \centering
    \begin{tabular}{cccc}
        \toprule
        \textbf{Annotation Type} & \textbf{Total Duration (hrs)} & \textbf{\# of annotations}  & \textbf{Mean \# tokens} \\
        \midrule
        1P-IS & 1.1 & 751 &  5.8 \\
        1P-R & 1.1 & 751 & 10.2 \\
        \midrule
        3P-IS & 4.7 & 5,132 & 5.3 \\
        3P-R & 4.7 & 5,132 & 11.3 \\
        3P-V & 4.7 & 5,132 & 14.6 \\
        \bottomrule
    \end{tabular}
    \caption{Annotation statistics.}
    \label{tab:annotation-stats}
\end{table}


Different third-party annotators attend to different body parts, movement, and emotions. First-party annotations are also intrinsically subjective. Internal states and rationales are colored by unknown prior context, personality attributes, and biological/cognitive states.

\subsection{Dataset Methodology and Technical Details}

\subsubsection{Train/Dev/Test/Private-Test Splits}

For both the \Naturalistic and \Improvised parts of the \mosaic dataset, we release public train, dev, and test sets and report the descriptive statistics in \Cref{tab:naturalistic_improvised}. We hold-out a private test-set for future benchmarking purposes. Given the amount of metadata about participants, dyads, and interactions, there are many dimensions by which the dataset could be partitioned. Given the degree of behavioral variation at the level of individuals, we have partitioned \mosaic dataset at the level of participants --- no individual participants are mixed with the train/dev/test splits within the primary \Naturalistic and \Improvised parts.

\begin{table}[h]
    \centering
    \begin{tabular}{ccccccc}
        \toprule
        \textbf{Part} & \textbf{Split} & \textbf{Hours} & \textbf{Interactions} & \textbf{Sessions} & \textbf{Participants} & \textbf{Prompts} \\
        \midrule
        \Naturalistic & Train & 2625.34 & 45,126 & 3,205 & 3,243 & 672 \\
         & Dev & 35.22 & 673 & 48 & 83 & 260 \\
         & Test & 44.13 & 786 & 58 & 109 & 291 \\
         & Private & 40.75 & 748 & 52 & 52 & 337 \\
        \midrule
        \Improvised & Train & 1268.01 & 16,623 & 1,656 & 897 & 829 \\
         & Dev & 20.98 & 323 & 28 & 41 & 155 \\
         & Test & 15.31 & 220 & 23 & 35 & 150 \\
         & Private & 15.32 & 240 & 28 & 38 & 138 \\
        \bottomrule
    \end{tabular}
    \caption{Descriptive statistics by major parts (\Naturalistic and \Improvised) and partition.} 
    \label{tab:naturalistic_improvised}
\end{table}

Given the dyadic nature of the collection and challenges in recruiting thousands of participants across the US, we allowed participants to participate in multiple sessions. For the \Naturalistic dataset, which does not involve any professional actors, we allowed participants to participate in up to 10 sessions. For the \Improvised dataset, we allowed participants to participate in up to 30 sessions with a max of 10 with the same partner. In the case of repeated sessions in the \Improvised part, we required that the same dyad be always given a new set of prompts to respond to. Across the data set, no two participants should have responded to the same prompt more than once.

A side effect of allowing multiple participation is the underlying network structure induced in the dataset - cliques of mutual participation are present. As such, participant-level partition is functionally \textit{clique}-level partition.

\subsubsection{Technical Set-up}
\label{section:technical-setup}
All recordings took place indoors. Over 95\% percent of the interactions took place in a common-room (both participants were present in the same room). Less than five percent of sessions were recorded in separate-rooms, with life-size monitors placed in front of participants. These separate-room set-ups were used to experiment with speaker-bleed reduction techniques, but the set-up is isolated to this small subset.

Audio is recorded at 48khz / 16bit using worn lapel mics (a small subset used Shotgun mics which were subsequently switched out due to high levels of speaker-bleed).

Video is recorded in UHD 4k with 16:9 vertical/portrait aspect ratio (2,160 x 3,840 pixels) at 30 FPS.

An example of the participant workflow is shown in appendix, in \Cref{section:participant-workflow}.

\subsubsection{Participant Recruitment}

\begin{figure}
\includegraphics[width=1\textwidth]{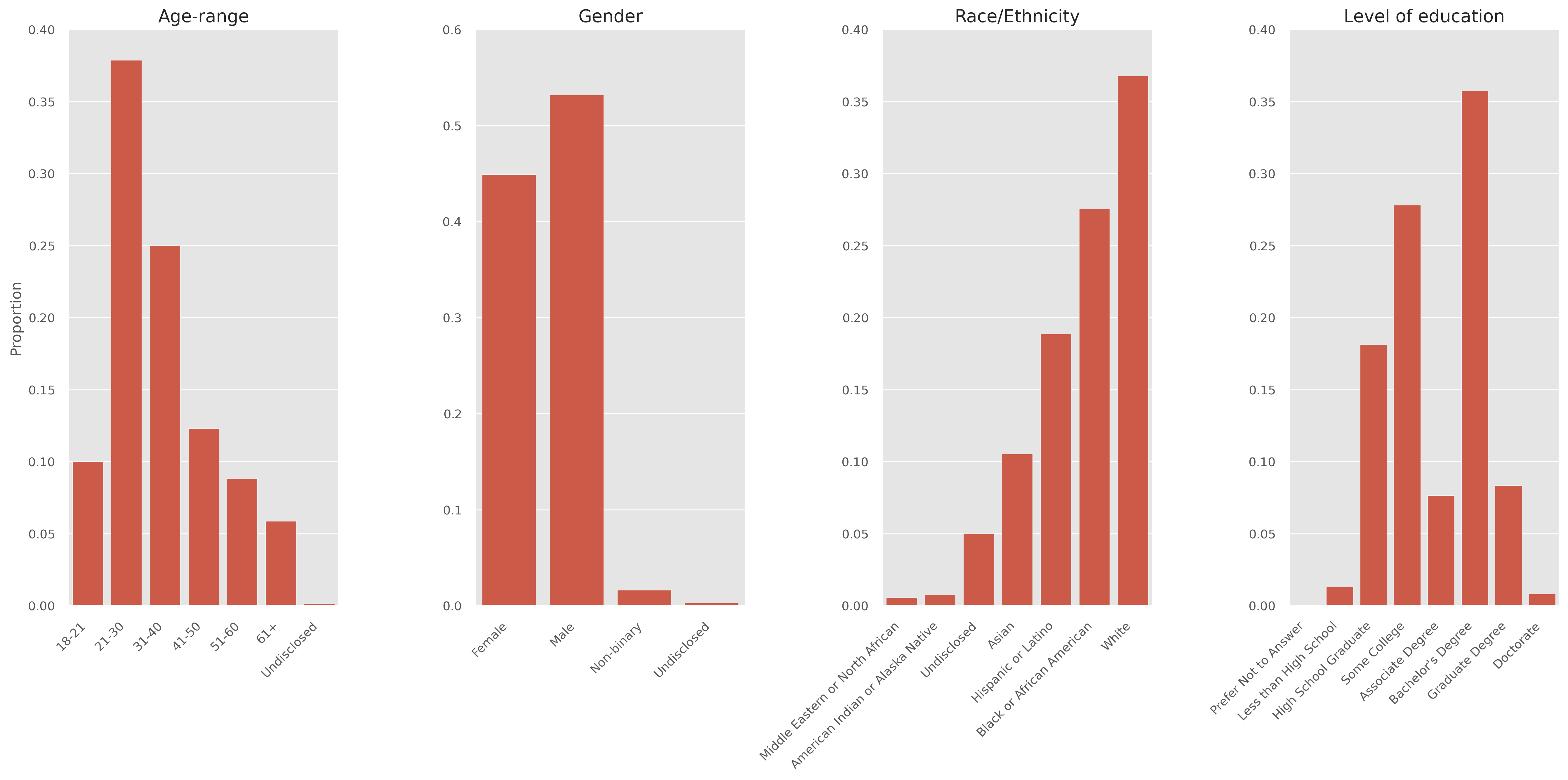}
\caption{Demographic category distribution. All demographic information was provided optionally by participants. Some participants chose not to disclose any information. Some chose to partially disclose.}
\label{fig:mosaic_demo_distr}
\end{figure}

Participants who were native speakers of English over 18 years of age were recruited in 10 cities across six states. All participants signed an informed consent and were paid for their time. We attempted a best effort at recruiting diverse pairs in terms of age, gender identity, and ethno-cultural background - both within pairs (i.e., many pairs with dissimilar individuals) and across pairs (i.e., many different pairs with similar individuals). Similarly, we attempted a best effort at recruiting diverse and equal distributions of relationship types (e.g., romantic partners, family members, friends, coworkers, acquaintances) for familiar dyads (see \Cref{tab:relationship_detail} for the distribution of familiar pair types). Demographic information was not used as part of the acceptance criteria in the collection and its disclosure was optional by the participants. \Cref{fig:mosaic_demo_distr} shows distributional properties across \textit{Age}, \textit{Gender}, \textit{Race/Ethnicity}, and \textit{Education} for the set of participants who disclosed this information.

\subsubsection{Moderators and Their Responsibilities}
\label{subsubsec:moderators}
A moderator is always present in a recording session. They can both see/hear the participants and speak to them, either via video/audio stream to a separate room, or in the same room but not visible in either video and far enough away that their key-strokes are not audible.

\paragraph{Time-stamping.} The moderator's primary job is to record the time stamps of when each prompt/interaction starts and ends (as provided by the time-synchronized recording software):
\begin{itemize}
    \item An interaction \textit{starts} the moment that the initiator begins speaking.
    \item An interaction \textit{ends} when:
    \begin{itemize}
        \item the participants mutually agree to move on to the next prompt
        \item a participant breaks character and asks a question to the moderator about the prompt
        \item the moderator themselves interjects in the conversation
    \end{itemize}
\end{itemize}

\paragraph{Preventing Privacy Disclosures and Sensitive Topics.} The moderator is instructed that it is absolutely essential that no personal details are disclosed during the interaction. This includes full names, addresses, email addresses, phone numbers, etc. (see \Cref{section:participant_preamble} for details). 

Moderators are also instructed that it is essential that participants not engage on potentially sensitive topics such as overt political debate, descriptions of violence, discrimination, harassment, hate speech, or anything that would cause undue discomfort. 

If a participant accidentally discloses personal details, or says something that could be considered a sensitive topic, the vendors are asked to remove this data from the delivery. The moderator is asked to be vigilant in listening for such instances, and in the event that they happen:
\begin{itemize}
    \item Ensure this is cut from the data. This could be done by stopping recording, removing the clip from the previous recording, and restarting, or by noting for the post-production team to remove it.
    \item Remind the participants again not to disclose good data.
\end{itemize}

\subsubsection{Known Limitations and Areas of Future Work}
Due to the scale and level of complexity in collecting the \mosaic dataset there are several aspects that will be the focus of continued work and improvement in future versions.

\paragraph{Errors in manual time-keeping} The core unit of the \mosaic dataset is an interaction. Interactions define \textit{active time} in which participant conversation and behavior can be linked to a pair of prompts (see Section \ref{section:interactions_session_scripts}). We have observed multiple instances of misaligned time-stamps in which moderators did not correctly identify start and end times for interactions or instances in which the start and end-times are correct, but mapped to the wrong interaction meta-data. 

Such misalignment result in interactions that may be too long (the annotated start time is too early or the annotated end time is too late) or too short (start time is too late or end time is too early). In some cases, the prompt text does not align with the spoken material indicating that the ordering of prompts was altered (likely due to off-by-one errors in the script ordering). In total, we believe that these errors, resulting in misalignment of prompt text metadata and interactions, impact approximately 10\% of interactions after our attempts at correction.

A certain degree of error is expected in a collection of this size and complexity, but it is important to take this into account. We have attempted a best effort to automatically identify and rectify these errors.

Time stamping "noise" are also present in the MOI segments used in our annotations (Section \ref{section:annotations}). Although there is a degree of subjectivity in defining an MOI, there are rare cases in which the described behavior represents only a subset of the observed behavior in the segment or cases in which the duration of the MOI does not fully capture the annotated behavior.

\paragraph{Incorrect assignment of participant IDs}
In rare cases, we have observed incorrect duplication of participant identifiers (a single identifier is assigned to two different participants). Conversely, we have also observed rare instances of the same person being mapped to different identifiers. We have made a best effort to rectify these errors when identified.

\paragraph{Unreleased \textit{meta time}}
As released, the \mosaic dataset only contains our \textit{active time} segments. This was in part because the essence of the dataset is derived from \textit{active time} interactions that are rooted in our taxonomy. However, the \textit{meta time} between interactions also represents literally hundreds of hours of fascinating data unto itself. Future releases may explore this subset of the data.

\paragraph{Variation in recording consistency}
This project was undertaken with multiple collection groups across multiple recording sites. The degree of recording quality (amount of speaker-bleed, consistency of participants staying in frame, quality of acting in \textit{Improvised} segments) and likelihood of time-stamping errors, varies by vendor with some recording sites displaying better competencies in aspects of the collection. All vendors met the basic technical requirements specified in Section \ref{section:technical-setup} and followed the steps outlined in Section \ref{subsubsec:moderators}, however there is clear variation in the level of production quality between vendors.

%% file: repr/arxiv.tex
\section{Multimodal Representations}
For each sample in the \mosaic dataset, we extract human-centric visual representations, speech tokens, and transcripts. These features enable a variety of downstream modeling applications for \mosaic data, including training and evaluation of dyadic audiovisual motion models. 
 
\subsection{Parametric Human Models}
\label{sec:parametric_human_models}

\paragraph{Body and hands representations.}
We use the \smpl \citep{smpl, romero2022embodied} model to represent the body and hands of each person. \smpl is a parametric human model and represents each person via a global orientation $\mathbf{\phi} \in \mathbb{R}^{3}$, body and hand pose (51 joint angles) $\mathbf{\theta} \in \mathbb{R}^{51 \times 3}$, and shape $\mathbf{\beta} \in \mathbb{R}^{16}$. The \smpl model uses these parameters to generate the mesh vertices $\mathbf{V} \in \mathbb{R}^{6980 \times 3}$ of a person.

For each video sequence in the \mosaic dataset, we track the person in the video and estimate global orientation and body pose using \hmr 2.0~\citep{goel2023humans}. This body reconstruction cannot capture hand details; the hands are flat. Thus, we also detect the hands of the person using ViTPose~\citep{xu2022vitpose} and, for each hand, we estimate the 3D hand pose using \hamer \citep{hamer}. Since the hands are reconstructed from a hand-centric perspective, they may be inconsistent with the arms from the body reconstruction. We address this issue in a post-processing step that transforms each hand-centric coordinate system into the corresponding one of each wrist. Finally, we use the same body shape ($\mathbf{\beta} = 0$) for all individuals.

\subsection{Imitator Face Representation}
\label{sec:rep_imitator_latent}

The Imitator latent representation aims to provide a low-dimensional encoding of facial expressions and positioning of high-level image features, enabling efficient and accurate modeling of talking-head videos. The approach uses two encoders: an expression encoder and an alignment encoder. These encoders work in tandem to extract relevant features from input images that collectively capture the essential characteristics of a talking head image. We will first describe the architecture and output of each encoder, with further details on their application and usage provided later in the paper.

We employ a pre-trained expression encoder that extracts expression features $\mathbf{f} \in \mathbb{R}^{n\times 128}$, using a typical Resnet34 backbone with a linear head. The encoder expects as input a roll-normalized facial image crop, which can be derived using facial landmarks. In parallel, an alignment encoder processes an upper body crop image to produce 3D translation values for the head and body $\mathbf{t}\in \mathbb{R}^{2\times 3}$, as well as rotation angles $\mathbf{R_{head}} \in \mathbb{R}^3$ for the head alone. The alignment encoder also uses a Resnet34 backbone with a linear head.

Both encoders have been trained end-to-end alongside a decoder as part of an internal model for talking head video generation. During training, the objective of image reconstruction on a large dataset with many identities is used to learn these representations in an unsupervised manner. As a result, these representations are not anchored to any interpretable units, such as translations. At inference time, these encoders are used to extract features from new input images, which can then be used for various downstream tasks such as those presented in this paper.

\subsection{Speech Representation}
\label{sec:speech-representation}

We use speech representation from an internally-built speech tokenizer. This speech tokenizer transforms raw speech into a set of discrete tokens, each representing a fixed time slice. We use this speech tokenized representations for conditioning of the dyadic motion models. When integrating with a speech-enabled LLM model, the same speech token representation is used.

\subsection{Text Transcripts}
We perform peak normalization on the raw audio for each video sample. We then post-process each dyadic pair of audios to remove speaker bleed using Beryl AEC, an in-house echo cancellation and suppression algorithm~\citep{berylaec}. It is a carefully tuned lightweight DSP-based module that gives great single-talk and double-talk performance on a wide variety of room, microphone, speaker, and device combinations. We use \silero~\citep{SileroVAD} to extract Voice Activity Detection (VAD) segments and \whisperx~\citep{bain23_interspeech} to obtain word-aligned text transcripts from post-processed audio. We provide details on the make-up of the transcribed corpus in \Cref{section:corpus_text_analysis}.

%% file: motion_model/arxiv.tex
\section{Dyadic Motion Models}

\label{Audiovisual_dyadic}

Our motion models are based on a Diffusion Transformer architecture trained with a flow-matching objective. They are conditioned on dyadic audio input and optionally user's visual features.

\paragraph{Architecture.} 
We use Diffusion Transformer (DiT)~\citep{peebles2023scalable} as the backbone.
From~\citep{zhuo2024lumina}, we use RMSNorm~\citep{zhang2019root} to improve training stability, as well as self-attention with key-query normalization (KQ-Norm) before the key-query dot product attention computation.

\paragraph{Training objective.}
We train with the flow-matching~\citep{lipman2022flow} objective, which learns the velocity field (or \textit{flow}) of samples moving from a \textit{noise} prior distribution (e.g., Gaussian) to the \textit{target} data distribution, and during inference generates samples starting with Gaussian noise.

Given a sample $\mathbf{x}$ from the target distribution, noise $\mathbf{\epsilon}$, and noisy latent $\mathbf{x}_t$, the model is trained to predict $\mathbf{v}_t=d\mathbf{x}_t/dt$.
We adopt the linear interpolation schedule between noise and data, i.e., $\mathbf{x}_t = t\mathbf{x} + (1-(1-\sigma_{min})t)\mathbf{\epsilon}$ where $\sigma_{min}=10^{-4}$. This formulation indicates a uniform transformation with constant velocity between data and noise. The corresponding time-dependent velocity field is given by $\mathbf{v}_t(\mathbf{x}_t) = \mathbf{x} - (1-\sigma_{min})\mathbf{\epsilon}$. 
During training, the flow matching objective directly regresses the target velocity:
\begin{equation}
    \mathcal{L}_{v} = \int_0^1 \mathbb{E}[\parallel \mathbf{v}_{\theta}(\mathbf{x}_t, t) - (x - (1-\sigma_{min})\mathbf{\epsilon}) \parallel^2]dt,
\label{eq:loss}
\end{equation}
which is named Conditional Flow Matching loss~\citep{lipman2022flow}, sharing similarity with the noise prediction or score prediction losses in diffusion models.

During inference, the time-dependent velocity field $\mathbf{v} : [0,1] \times \mathbb{R}^d \to \mathbb{R}^d$ defines an ordinary differential equation (ODE): $d\mathbf{x} = \mathbf{v}_t(\mathbf{x}_t)dt.$ 
We use $\phi_t(x)$ to represent the solution of the ODE with the initial condition $\phi_0(\mathbf{x}) = \mathbf{\epsilon}$. By solving this ODE from $t=0$ to $t=1$, we can transform noise into data samples using the learned velocity field $\mathbf{v}_{\theta}(\mathbf{x}_t, t)$.

\subsection{Audio-only Dyadic Motion Model}

\paragraph{Face and body motion feature processing.}
To model face, we use the Imitator representation which includes an expression code, head rotation, and translation values. We normalize these values and concatenate them to obtain a sequence of facial motion features $\mathbf{f} \in \mathbb{R}^{N\times 137}$ for $N$ video frames. For the body and hands of each individual we use rotations for the \smpl joint angles in 6D representation~\citep{Zhou20196DRotation}. 
These are smoothed using a Savitzky-Golay filter to reduce the jitter.
As the \mosaic dataset contains mainly upper-body gestures, we ignore the 8 leg joints of \smpl, resulting in a sequence of body motion features $\mathbf{b}\in \mathbb{R}^{N\times 258}$. 

\paragraph{Dyadic audio conditioning.} We condition the motion model on the speech tokens obtained from the dyadic audio of the two speakers.
We embed the speech tokens from the two audios with a shared embedding table and concatenate them to encode the speech conditions. To condition the diffusion model, we project the input motion features and the condition features to the same feature space and add them before feeding to the Transformer network. 
To deal with the mismatch of speech and visual rate (12.5~fps vs 30~fps), we resample the speech conditions to match the length of motion features before the addition.
We find that this conditioning approach results in more aligned motion sequence generations compared to cross-attention conditioning.

\subsection{Audiovisual Dyadic Motion Model}

\paragraph{Audiovisual dyadic conditioning.}
In addition to dyadic speech, we also experiment with conditioning the motion model with user's visual features. The user's visual conditions are also concatenated with the speech conditions and are then fed into the Transformer model. For the face, we use Imitator latent as visual information; for the body, we leverage the user's \smpl pose parameters.

\paragraph{Finetuning.} 

To effectively integrate multimodal visual and enhance high-fidelity generation, we design a two-stage training scheme. In the first stage, we train the model over a full set of training data. In the second stage, we finetune the model with flow-matching loss reweighting~\citep{zhang2025energy}, which allows us to better take advantage of the data from an automatic reward.

\subsection{Face+body Joint and Cascaded Model}
To generate both face and body motion sequences, we adopt two different approaches: (a) a joint-model approach, which uses one single model to generate face and body features jointly; and (b) a cascaded-model approach training the face-motion model and body-motion model separately in a cascaded fashion.

\begin{figure}[t]
    \centering
    \includegraphics[width=0.7\linewidth]{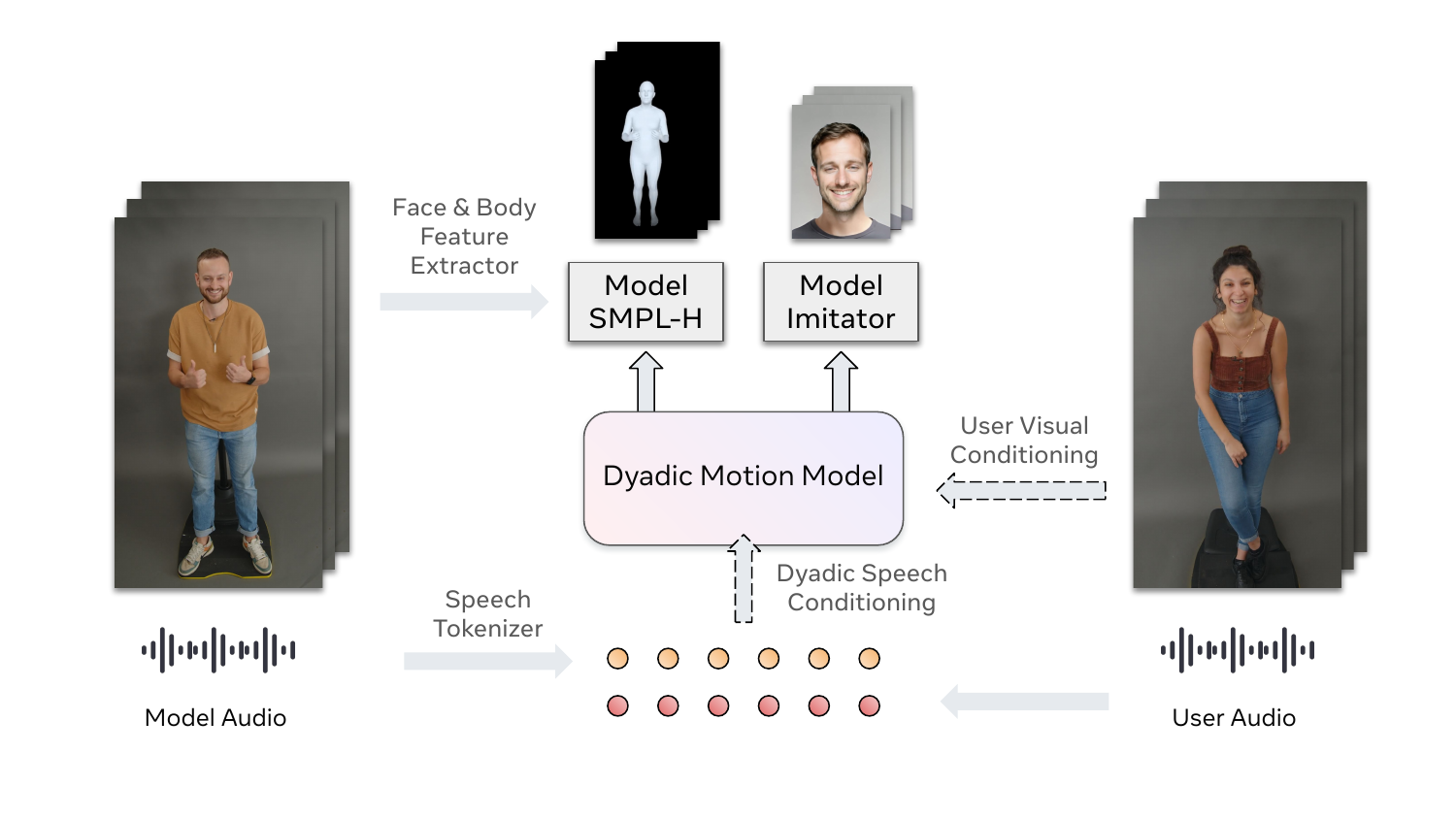}
    \caption{
    %
    Our Dyadic Motion Model is conditioned on speech tokens obtained from dyadic audio and optionally user's visual features, and generates face Imitator features and body SMPL-H features. For Joint Face+Body Model, we concatenate face and body features and generate them jointly.
    }
    \label{fig:joint_diffusion}
\end{figure}

\paragraph{Joint model.}
We use a single face+body motion model to generate face and body motion features jointly as shown in \Cref{fig:joint_diffusion}.
We simply concatenate the face motion features $\mathbf{f} \in \mathbb{R}^{N\times 137}$ and the body motion features $\mathbf{b} \in \mathbb{R}^{N\times 258}$ to obtain joint face+body motion features $\mathbf{j} \in \mathbb{R}^{N\times 395}$. This approach enables the generation of aligned face and body sequences.

\begin{figure}[t]
    \centering
    \includegraphics[width=0.7\linewidth]{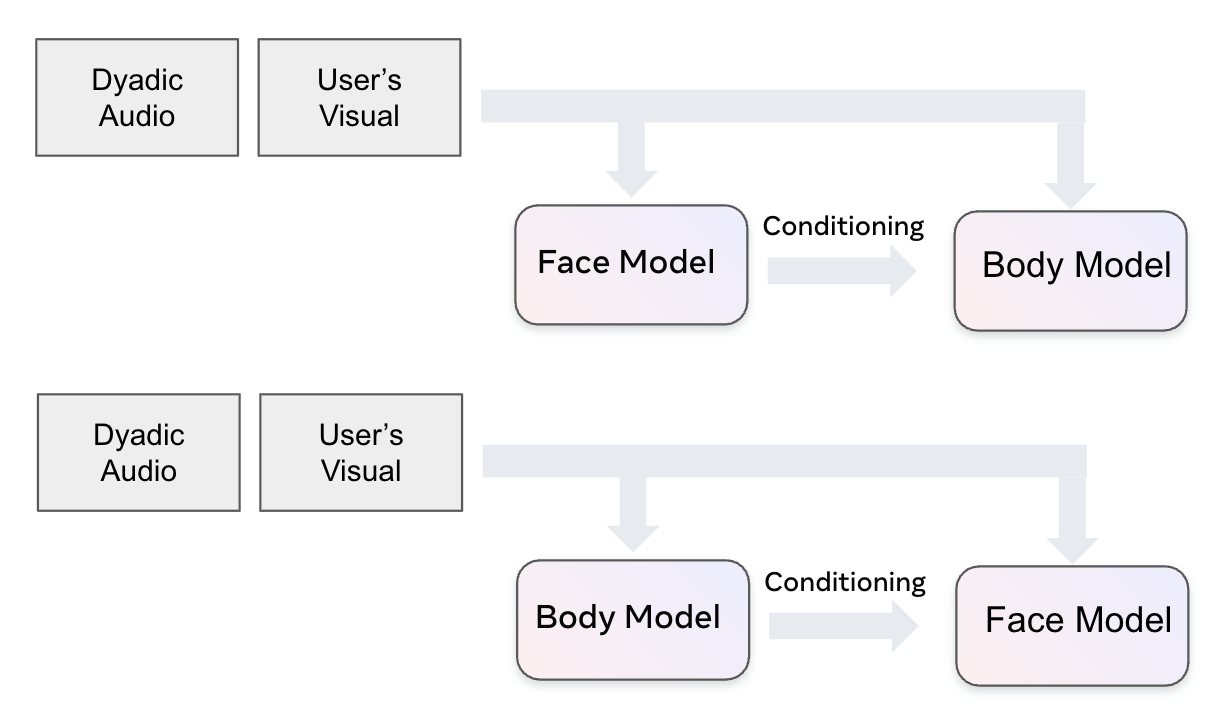}
    \caption{
    Cascaded model (Face2Body or Body2Face generation). 
    }
    \label{fig:cond_pipeline}
\end{figure}

\paragraph{Cascaded model.}
As illustrated in \Cref{fig:cond_pipeline}, to align the body's head pose with face's head rotation, we design the conditional Face2Body model and Body2Face model for cascaded generation. The design tries to avoid the misalignment (especially the head rotation) between face and body outputs.

For Face2Body model, we train several variants with different conditioning, including: (1) full imitator latent; (2) head rotation of imitator latent. In experiments, we find that these models effectively align face with body. 

In inference, we first predict the face imitator latent, following which we predict the body features conditioning on the face features with Face2Body model. As for Body2Face model, the face model is conditioned on body's \smpl representation, to guarantee the alignment between face and body. For automatic evaluation of full-body generation pipeline, the comparison among generations from face + Face2Body, or body + Body2Face pipeline is included. 
In inference, we first predict the body features and then leverage the Body2Face model for predicting the face features.

\subsection{Motion Models with Controllability}
\label{subsec:controllability}

\input{motion_model/controllability.tex}

\subsubsection{Semantic Gesture Controllability}
\label{sec:sem_ges_cond}

Semantic gesture generation plays a crucial role in enhancing the naturalness and expressiveness of virtual agents in speech-driven gesture generation tasks. However, with only speech as a condition, generative models face significant challenges in producing semantic gestures due to their rarity and long-tail distribution.
To tackle this challenge, we propose semantic controllability.

\paragraph{Methodology.} To realize zero-shot arbitrary gesture conditioning in the co-speech diffusion model, we introduce a random \textbf{temporal gesture condition dropping} strategy. This would allow generating any unseen gestures with a smooth transition as long as we have the gesture condition. Let \(\mathcal{G} = \{\mathbf{g}_i\}_{i=1}^T\) denote the ground truth gesture sequence of length \(T\), where \(\mathbf{g}_i\) represents the gesture at time step \(i\). Let \(\mathbf{s} = \{s_i\}_{i=1}^T\) denote the corresponding speech condition sequence, where \(s_i\) is the speech feature at time step $i$.

During training, we apply a random temporal gesture condition drop to simulate partial gesture absence. For each frame index \(i\), the gesture condition \(\mathbf{g}_i\) is retained with probability \(1 - p_{\text{drop}}\) or replaced with a null condition (e.g., a zero vector or mask token) with probability \(p_{\text{drop}}\). The speech condition \(\mathbf{s}\) is always retained to ensure consistent contextual grounding. The modified gesture sequence \(\mathcal{G}_{\text{drop}}\) is defined as:

\[
\mathcal{G}_{\text{drop}} = \{\tilde{\mathbf{g}}_i\}_{i=1}^T, \quad \tilde{\mathbf{g}}_i = 
\begin{cases} 
\mathbf{g}_i & \text{with probability } 1- p_{\text{drop}}, \\
\mathbf{0} & \text{with probability } p_{\text{drop}},
\end{cases}
\]

where \(\mathbf{0}\) denotes the null condition. The diffusion model is trained to predict the original gesture sequence \(\mathcal{G}_{\text{pred}}\) given the dropped gesture sequence \(\mathcal{G}_{\text{drop}}\) and the full speech condition \(\mathbf{s}\) at each diffusion step. This approach encourages the model to generate rhythm-aware gestures when only speech is provided as a condition, and to follow semantic gestures when gesture conditions are available.

In addition to using \smpl as a gesture condition, we also experimented with using VQ-VAE codes of the SMPL parameters as the gesture condition. 
A VQ-VAE is trained to encode \smpl parameters into a discrete codebook \(\mathcal{C}\) of size \(|\mathcal{C}|\), with codebook indices ranging from \(0\) to \(|\mathcal{C}| - 1\). For each SMPL gesture \(\mathbf{g}_i\), the VQ-VAE encoder produces a latent embedding, quantized to the nearest codebook entry, yielding a discrete code \(\mathbf{c}_i \in \mathcal{C}\). The gesture sequence is represented as \(\mathcal{G}_{\text{VQ}} = \{\mathbf{c}_i\}_{i=1}^T\). Similarly, we have the temporarily randomly dropped VQ ID sequences as a gesture condition during training:

\[
\mathcal{G}_{\text{VQ,drop}} = \{\tilde{\mathbf{c}}_i\}_{i=1}^T, \quad \tilde{\mathbf{c}}_i = 
\begin{cases} 
\mathbf{c}_i & \text{with probability } 1- p_{\text{drop}}, \\
\mathbf{|\mathcal{C}|} & \text{with probability } p_{\text{drop}},
\end{cases}
\]

The diffusion model is trained to predict the original gesture sequence \(\mathcal{G}\) (in \smpl parameters) given the dropped VQ code sequence \(\mathcal{G}_{\text{VQ,drop}}\) and the full speech condition \(\mathbf{s}\) at each diffusion step. 

\paragraph{Temporal Condition Integration.}
For discrete temporal conditions, such as gesture VQ IDs and audio token IDs, we map the IDs to randomly initialized embeddings, which are tuned during diffusion training, and encode them to the target dimension. For continuous temporal conditions, such as the \smpl condition, we employ an MLP to get encoded features. These conditions are then concatenated with the noisy diffusion input along the channel dimension, and the concatenated output is linearly projected back to the original dimension of the noisy input.

%% file: motion_model/controllability.tex
\subsubsection{Emotion Controllability}
\label{sec:emo_control}
In order to achieve precise and fine-grained control over the emotion during generation, we incorporate the Arousal-Valence (A-V) Matrix~\citep{russell1980circumplex}. This model provides a structured, two-dimensional framework for representing and steering affective states within the generated output. Compared to categorical emotion labels, which are often difficult to classify reliably and prone to oversimplification, the A-V Matrix offers significant advantages: it naturally captures the continuous spectrum and subtle gradations of human affect. We use an internal tool to extract arousal and valence sequences for each sample: $\mathbf{a} = \{a_0, a_1, ..., a_{n-1}\}$ and $\mathbf{v} = \{v_0, v_1, ..., v_{n-1}\}$, where $n$ is the frame number of the sample.

The arousal and valence sequences $\mathbf{a}$ and $\mathbf{v}$ fall into the numerical range from $-1$ to $1$. A straightforward way is using the continuous arousal and valence values as conditioning signals, i.e.,  the model learns the conditional distribution $p(\mathbf{lat_t} \mid \mathbf{lat_{t-1}}, \mathbf{a}, \mathbf{v})$ for timestep $t$, where $\mathbf{lat_t}$ means the Imitator latent of the $t$-th timestep. In order to improve the robustness of conditioning and fit discrete emotion tokens as used in \Cref{subsubsec:codebook}, we propose \textbf{bucket-based conditioning} in place of real-valued emotion conditions. Specifically, continuous values are discretized using predefined buckets (e.g., partitioning the $[-1,1]$ valence space into $k$ intervals, in practice, uneven buckets can be set empirically). Each $(\mathbf{a}, \mathbf{v})$ pair is mapped to a bucket index $b_a, b_v \in \{1, 2, \dots, k\}$. The model then learns $p(\mathbf{lat_t} \mid \mathbf{lat_{t-1}}, b_a, b_v)$, allowing generation to be conditioned on value \textit{ranges} rather than exact values.

During training, we apply \textit{condition dropout} (rate $\rho$) to randomly mask these signals for the above two conditioning methods, enabling the model to learn the generation of conditioned and unconditioned.

During inference, the model can generate outputs in the following manners: 1) unconditioned generation; 2) conditioned on constant values $(a_c, v_c)$ for stable emotional profiles; 3) conditioned on externally generated sequences $({\mathbf{a}}, {\mathbf{v}})$ (e.g., from audio features or semantic analysis) for dynamic affect.

\subsubsection{Expressivity Controllability}
\paragraph{Control signals.}
The facial expressivity is related to the movement of multiple correlated components (i.e., head pose, eye brows, mouth region, etc.). Hence, we design our architecture to jointly conditioned on these control signals. Specifically, we consider the following control signals:
\begin{itemize}
    \item \textbf{Head rotation}: For each video frame sequence of length $n$, we extract a 3-dim (pitch, yaw, roll) head rotation sequence $\mathbf{R_{head}} = \{\mathbf{r_0}, \mathbf{r_1}, ..., \mathbf{r_{n-1}}\}$, where each $\mathbf{r}_i \in \mathbb{R}^3$ is the instantaneous Tait-Bryan rotation vector:
    \begin{equation}
        r_i = \left( 
                \begin{aligned}
                    & r_{pitch} \\
                    & r_{yaw} \\ 
                    & r_{roll} 
                \end{aligned}
                \right) (radians).
    \end{equation}
    The values of $r_{pitch}$, $r_{yaw}$, $r_{roll}$ represent the radian angle along pitch, yaw, roll axis respectively. In concrete terms, $r_{pitch}$ embodies the rotation of the head in the sagittal plane, with positive values denoting chin downward and negative values denoting chin upward, thus we can regard there is a head nod motion if $\Delta(r_{pitch})$ is larger than a threshold. Similarly, $r_{yaw}$ means horizontal plane rotation, i.e., head turns to left or right. $r_{roll}$ represents coronal plane rotation, describing the head tilt.
    \item \textbf{Eye brows}: To encode the person-independent facial expressions, we use the Facial Action Coding System (FACS) \citep{ekman1997face} which describes a taxonomy of facial action units (FAU) to capture the movements of different muscles or muscle groups. For each video sequence with length $n$, we first use an internal tool to extract the FAU sequence $\mathbf{F_{fau}} = \{\mathbf{f_0}, \mathbf{f_1}, ..., \mathbf{f_{n-1}}\}$, where each $\mathbf{f}_i \in \mathbb{R}^{46}$ is the FAU value vector with 46 facial motion unit values. Next, we select the 3 FAU types only related to eye brows and use their average value vector $\mathbf{m_{eb}}$ as the control signal for eye brows movement.
    \item \textbf{Mouth}: Similar to the eye brows control signal, we select the 20 FAU types related to mouth and use their average value vector $\mathbf{m_{m}}$ as the control signal for the mouth movement.
    \item \textbf{Eye Gaze}: We introduce gaze values to control the angle and direction of eye gaze. For each video sequence of length $n$, we extract the 2-dim pitch and yaw eye gaze sequence $\mathbf{g_{gaze}}=\{\mathbf{g_0}, \mathbf{g_1}, ..., \mathbf{g_{n-1}}\}$, where $g_i \in \mathbb{R}^2$ is the rotation vector of pitch and yaw values for both eyes.
\end{itemize}

\paragraph{Conditioning methods.}
Given the continuity of each control signal, we apply a moving average to it to mitigate the interference of outliers in the original data.
After that, for each aforementioned control signal, we propose several ways to introduce them into the model training process (along with \textit{condition dropout} of $\rho$ for each signal to enable both conditioned and unconditioned generation setting). 
Similar to the conditioning ways mentioned in emotion controllability (Sec~\ref{sec:emo_control}), the easiest way to involve those control signal is using the original sequences ($\mathbf{R_{head}}$, $\mathbf{m_{eb}}$, $\mathbf{m_{m}}$, $\mathbf{g_{gaze}}$) directly.

In order to focus more on movement patterns rather than static poses, we calculate and condition the model on motion dynamism $\mathbf{\dot{s}}$ for each $\mathbf{s} \in \{\mathbf{R_{head}}, \mathbf{m_{eb}}, \mathbf{m_{m}}, \mathbf{g_{gaze}}\}$: 
\begin{equation}
    \begin{aligned}
        & \mathbf{\dot{s}} = \{\dot{s_0}, \dot{s_1}, ..., \dot{s}_{n-1}\}, \\ 
        & \dot{s}_t = \text{abs}(s_t - s_{t-1}).
    \end{aligned}
    \nonumber
\end{equation}

For dynamism ($\mathbf{\dot{s}}$) of each control signal, we run over a random selected subset including $300$-hour videos to obtain the statistics, including the maximum value, the minimum value, and different quartiles. We then build up a set of threshold $\mathbf{\tau} = \{\tau_0, \tau_1, ... \tau_{k-1}\}$, where $k$ is the total number of buckets in total. Thus, it is easy to convert the signal into bucket index. Take the head rotation as an example:
\begin{equation}
    \mathbf{b^j} = \text{bucket}(\mathbf{r_j}) = \sum_{i=1}^{k-1} \mathbb{I} \left( \mathbf{r_j} > \tau_i^j \right), \quad j \in \{\text{pitch, yaw, roll}\}.
    \nonumber
\end{equation}
Then the model takes the bucket index $b_j$ as the condition:
    $$p(\mathbf{lat_t} \mid \mathbf{lat_{t-1}}, {\mathbf{\dot{b^{head}}}}, \mathbf{\dot{b^{eb}}}, \mathbf{\dot{b^{m}}}, \mathbf{\dot{b^{gaze}}}),$$ 
    where $\mathbf{\dot{b^{head}}}, \mathbf{\dot{b^{eb}}}, \mathbf{\dot{b^{m}}}, \mathbf{\dot{b^{gaze}}}$ are the bucket index sequences of $\mathbf{\dot{R_{head}}}$, $\mathbf{\dot{m_{eb}}}$, $\mathbf{\dot{m_{m}}}$, and $\mathbf{\dot{g_{gaze}}}$ respectively. 
Taking advantage of the statistics, the bucket-based conditioning approach is robust to perceptually equivalent classes.

\paragraph{Expressiveness level.}
While the model can achieve fine-grained single-signal control, different combinations of multiple signals can present different levels of expressiveness, allowing the overall modulation of expressiveness. For example, when we want the avatar to perform more nodding motions, we can set a higher pitch-axis head rotation dynamism $\mathbf{\dot{r}_{pitch}}$ during inference. With more head nodding, the avatar tends to be regarded as more expressive. In a similar way, setting large eyebrow motion dynamism mouth motion can also help with improving the avatar expressiveness.

%% file: llama_integration/arxiv.tex
\subsection{Integration with Speech LLM Model}

We integrated our dyadic motion models with a speech LLM model trained on transcribed dialogs.

\begin{figure}[H]
    \centering
  \includegraphics[width=0.6\linewidth]{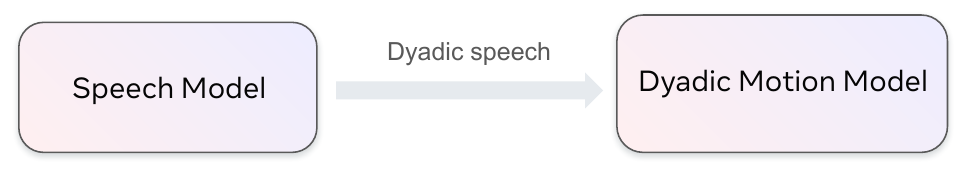}
  \caption{Cascaded integration of dyadic diffusion with speech LLM model.}
  \label{fig:llama_diffusion}
\end{figure}

For this work, we always freeze the speech model to ground our diffusion generation.
During inference, we initiate two speech models as two agents to talk to each other, each treating the other agent as a user.
Two agents together can generate coherent and expressive dialogues following our dyadic speech prompts, showing great in-context learning capability.
As shown in \Cref{fig:llama_diffusion}, we feed the generated dialogue (in the form of speech tokens) to ground our dyadic motion models.

\subsubsection{LLM-Guided Codebook Generations}
\label{subsubsec:codebook}

Diffusion motion models conditioned on speech and visual signals have demonstrated natural face and body generation in dyadic interactions. Additional conditioning further modulates the generation, for example, the emotion control adds more expressivity to facial expressions. This suggests that diffusion models could benefit from extra guidance from the speech model.
Hence we propose \adapter and codebooks as the bridge between speech model and diffusion. Codes are inferred from speech model, and then are used as conditions to drive diffusion models.

\begin{figure}[H]
  \centering
  \includegraphics[width=0.9\linewidth]{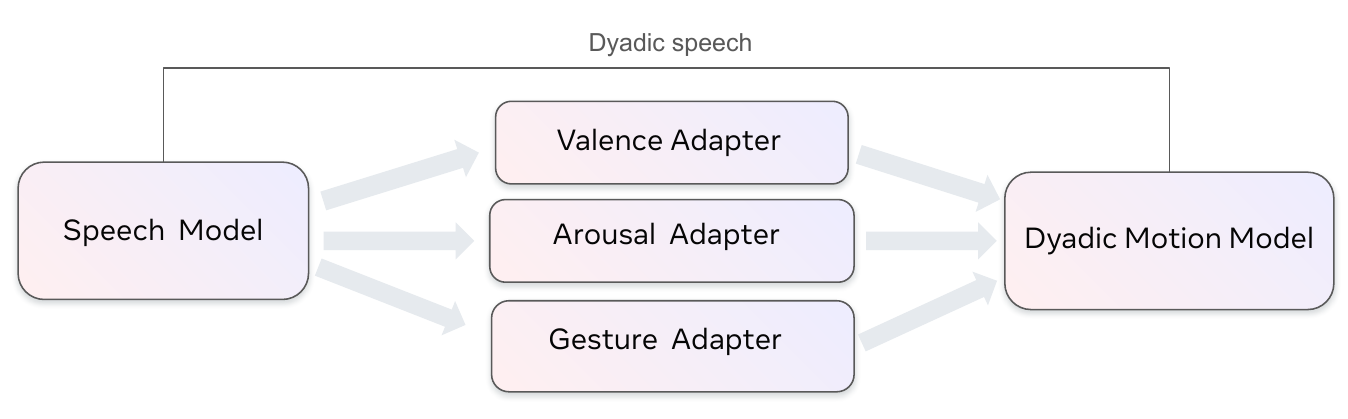}
  \caption{Codebook integration of dyadic diffusion with speech LLM model.}
  \label{fig:llama_codebook_diffusion}
\end{figure}

Now we will present the design for the LLM-driven codebook integration with speech model and diffusion. As shown in \Cref{fig:llama_codebook_diffusion}, an \adapter is built on top of a speech language model, taking its hidden states
as input and predicting emotion and gesture codes to guide the face and body diffusion models. 

\textbf{Adapter}. The \adapter is a multi-layer perceptron with GELU activation. Suppose that the predicted token rate of the adapter is $R$, that is, the number of tokens in one second. We extract the hidden states of the agent speech tokens from the last layer of speech model, and interpolate these representations to match the token rate of $r$. The \adapter transforms interpolated speech representation with MLP layers, and projects it to codebooks via the projection layer.
\begin{align}
\tilde{\mathbf{C}}=\text{Adapter}(\text{Interpolate}(\mathbf{H}, r)),
\end{align}
where $\mathbf{H}$ is speech model hidden states, and $\tilde{\mathbf{C}}$ is prediction on predefined codebooks.

For training, speech model takes dialog data as input, and the adapter extracts the hidden states from its last layer and makes code predictions. Cross-entropy loss is measured between ground truth codes $\mathbf{C}$ and predictions $\tilde{\mathbf{C}}$. The \adapter is tuned to minimize the loss, while the parameters in speech model are all frozen.

We introduce the emotion and gesture codebooks below, and discuss how they guide the face and body diffusion models, respectively.

\paragraph{LLM-Driven Generations with Emotion Codebook.}
The emotion codebook consists of $12$ valence tokens and $12$ arousal tokens. Similar to \Cref{sec:emo_control}, valence and arousal are extracted from the face in each video frame.  Their values range from $0$ to $1$, and are quantized into $12$ equal-sized bins as discrete emotion tokens. Emotion adapters are trained to predict valence and arousal tokens.

The face diffusion model takes valence and arousal as extra conditioning information in addition to dyadic speech. The emotion condition modulates the facial expressions in generation.

\paragraph{LLM-Driven Generations with Semantic Gesture Codebook.}
The \mosaic dataset provides a set of \textit{semantic gesture game} data where actors make semantic and illustrative gestures while talking about words of interest. We collected a set of gestures as the gesture vocabulary, and built a lookup table to map each gesture to its corresponding \smpl sequence. 

To train a gesture \adapter, we labeled the segments with a gesture that its spoken words trigger. If no semantic gesture exists in one segment, we assign a special null gesture label. The adapter essentially learns a multi-class classification task, predicting a gesture on a given spoken segment. 

The body diffusion model takes predicted gesture labels as well as speech to control gestures and body movements as described in \Cref{sec:sem_ges_cond}.

%% file: visualization/arxiv.tex
\section{Visualization}

Our Dyadic Motion Models were designed to always generate two streams of motion codes: one stream for the face (a.k.a. \textit{face codes}) and one stream for the body (a.k.a. \textit{body codes}). A significant advantage of generating these intermediate codes instead of directly generating video pixels is that we can not only generate 2D videos with our Dyadic Motion Models, but also 3D rendered Codec Avatars, that can be visualized in settings like Virtual Reality (VR) headsets or Augmented Reality (AR) glasses. 

\subsection{2D Video Generation}

Our 2D video generation approach follows the image-to-video technique where the generative process is conditioned on face and body pose. Given a single reference image of the subject, the sequence of face codes, and the body code sequence, the algorithm generates a 2D video in which facial appearance, head motion, and articulated body dynamics are coherently aligned with the face and body codes.

\subsubsection{2D Rendering}
Our 2D Rendering model is a diffusion model that receives a single reference image $I \in \mathbb{R}^{3\times H \times W}$ and, while operating entirely in a compressed latent space, synthesizes a temporally coherent video clip \(V\).
The design of the rendering model interleaves three key components—(1) a Temporal Autoencoder (TAE), (2) window\hyp{}efficient Diffusion–Transformer (DiT) blocks~\citep{peebles2023scalable}, and (3) a lightweight latent\hyp{}space concatenation strategy.
Together, these modules enable high\hyp{}fidelity, temporally smooth animations. We initialize our model from a pretrained diffusion model checkpoint and train on the \mosaic dataset.

\paragraph{Temporally AutoEncoder (TAE).}
The pipeline begins by inflating a conventional 2D VAE into \(3\text{D}\).
Spatial convolutions alternate with lightweight \(1\text{D}\) temporal blocks, allowing the TAE encoder to map the reference image to a spatio\hyp{}temporal latent grid $z \in \mathbb{R}^{T \times H' \times W' \times C}$, where each axis is down\hyp{}sampled by a constant factor (e.g.,~8) yet perceptual detail is largely retained. This latent grid serves as the substrate upon which all subsequent denoising and conditioning operations are performed. 

\paragraph{Diffusion–Transformer (DiT).}
To process \(z\), a single 3D convolution patch\hyp{}embeds the grid and flattens it into a token stream enriched with factorized positional encodings.
A DiT then iteratively denoises these tokens; its self\hyp{}attention is restricted to \emph{shifted, non\hyp{}overlapping windows}.
The shift mechanism, inherited from the Swin\hyp{}Transformer~\citep{liu2021swin}, propagates information across windows while keeping the cost nearly linear in sequence length~\citep{peebles2023scalable}.

\paragraph{Image conditioning.}
To animate a still photograph, the encoder treats \(I\) as a single frame video.
Its latent slice \(z_{0}^{\text{img}}\!=\!E(I)\) is replicated along the temporal axis to the target length \(T\).
During diffusion, the current noisy sample \(z_t\) is concatenated channel\hyp{}wise with this fixed appearance code, anchoring identity and illumination while the network synthesizes plausible future motion.

Finally, the TAE decoder maps \(z_{0}\) back to the RGB space, yielding the video $z_{0}$.
If even stronger identity preservation is desired, additional vision\hyp{}specific tokens can be appended to the token stream without retraining the backbone.

\subsubsection{Body and Face Conditioning}
In our cascaded pipeline the body and face are driven by two dedicated code sequences: a \smpl sequence $b_{1:T}$ for full-body articulation and face latent (Imitator latent) $f_{1:T}$ for expression dynamics. 
  
\paragraph{Body conditioning.} \label{sec:model_2d_body_cond}
Following a rich line of pose-based video synthesis work \citep{Villegas2018,Chan2019,Siarohin2019,Yang2020,Petrov2023}, we represent body motion with a \emph{human-skeleton video}. For every frame in the training set we first detect 2D keypoints using Sapiens~\citep{khirodkar2024sapiens} and connect them according to the canonical limb graph defined by OpenPose~\citep{Cao2017}. At each diffusion timestep 
\(t\), the resulting pose tensor is concatenated (along the channel dimension) with the encoded reference image and the noise latent \(z_t\), enabling our model to couple appearance and motion in a single forward pass.

At inference we only dispose of a predicted \smpl mesh sequence rather than 2D keypoints. Because \smpl is defined in an uncalibrated, human-centric coordinate frame, we first estimate the full camera parameters from the reference image --- recovering intrinsics (e.g., focal length) and extrinsics (rotation~\&translation) with a standard PnP solver \citep{Lepetit2009} initialized by structure-from-motion~\citep{Schonberger2016}. 
We render the mesh sequence under this camera to produce an aligned synthetic video and then run Sapiens~\citep{khirodkar2024sapiens} to obtain the 2D keypoints, which are fed into the diffusion model exactly as during training. This procedure ensures that pose guidance remains geometrically consistent even when the only available motion source is a 3D parametric body model.

\begin{figure}[H]
    \centering
    \includegraphics[width=0.8\linewidth]{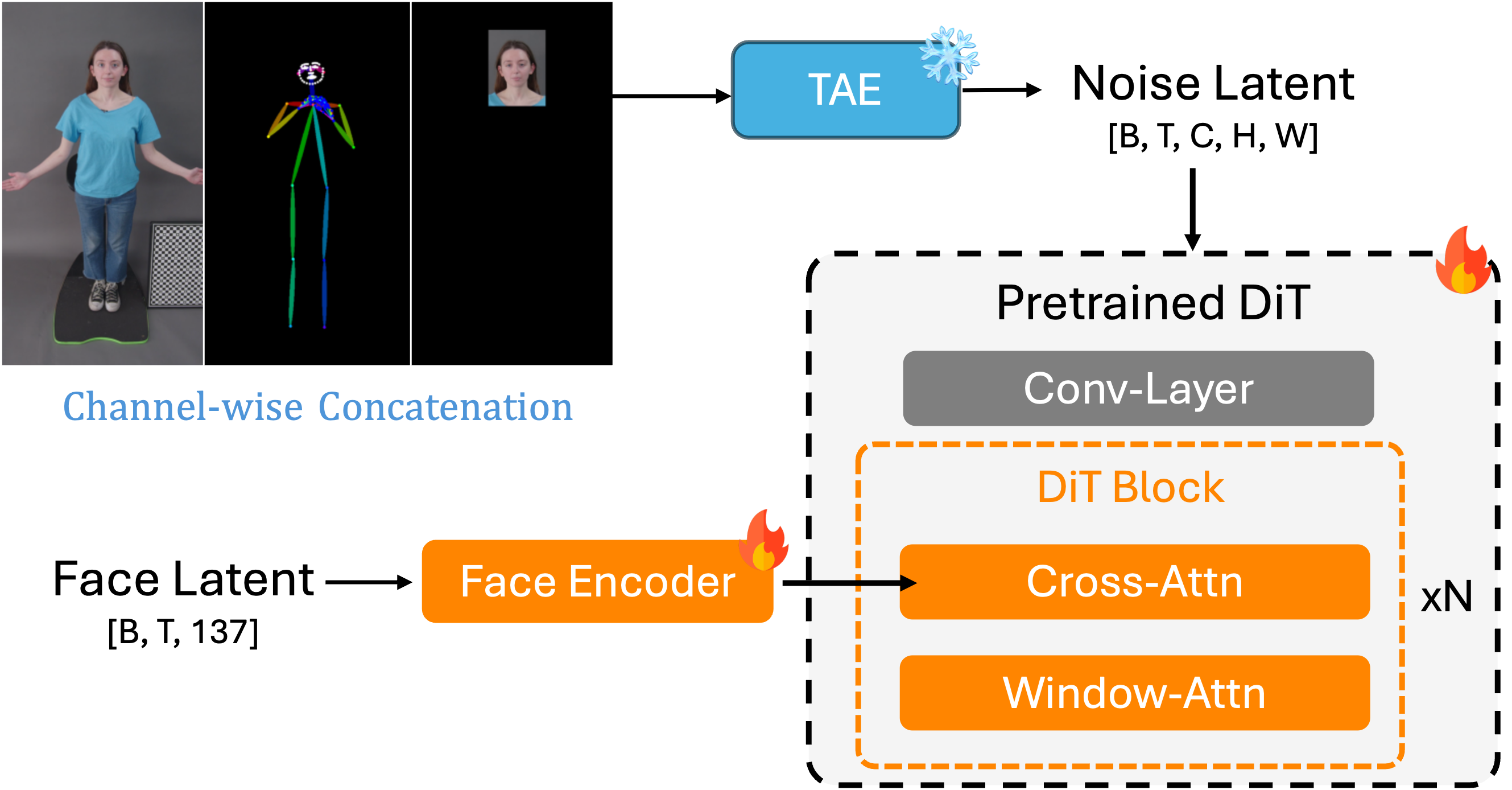}
    \caption{2D  rendering pipeline.}
    \label{fig:p2v_pipeline}
\end{figure}

\paragraph{Face conditioning.} 
There are two vanilla ways to apply face conditioning: (a) the inpainting method, which decodes the face latent into a face video, and the model impaints the body video on top of the face video; and (b) direct injecting the face latent through cross-attention. Both vanilla methods have shortcomings. The inpainting method cannot handle the scenario where the hand overlaps with the face intended by body conditioning, while the direct injecting scheme suffers from quick overfitting.  

To overcome the issues of both vanilla methods, our 2D Avatar renderer adopts a hybrid scheme. We apply channel-wise concatenation to the face video, reference image \(I\), pose-skeleton sequence, and \(z_t\) before the TAE encoder (\Cref{sec:model_2d_body_cond}). Meanwhile, every DiT block takes face latent $f_{1:T}$ (after extra face encoder) in its cross-attention layer. During both training and inference, we drop any face-video frame whose face is occluded by another body part — a condition detected directly from the pose skeleton. The network is therefore forced to reconstruct the missing regions from the expression latent, the pose guidance, and the surrounding unoccluded frames, yielding stable lip-sync and facial expressiveness even under severe self-occlusion.

\subsection{3D Codec Avatar Rendering}
\begin{figure}[H]
    \centering
    \includegraphics[width=1.0\linewidth]{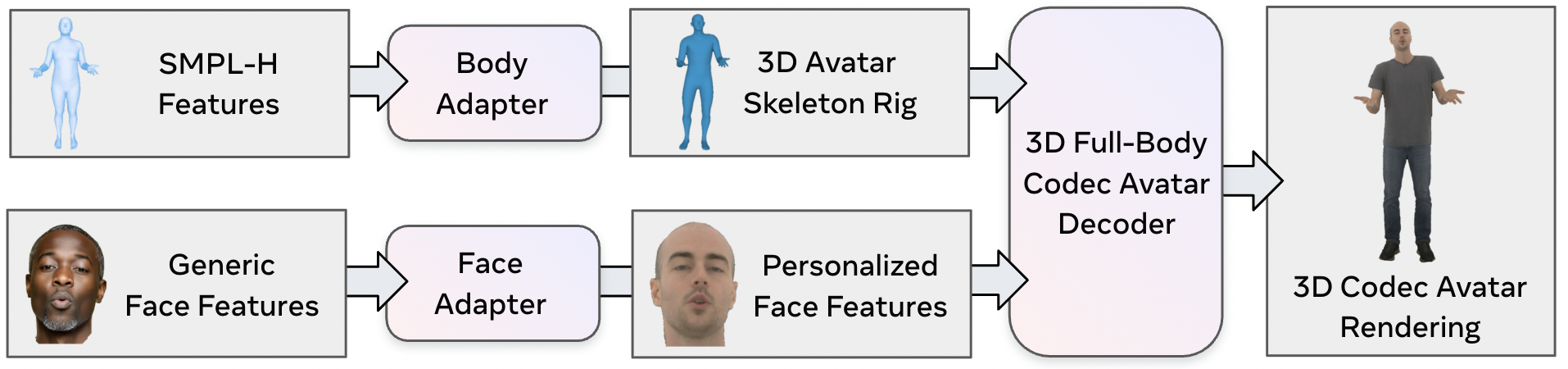}
    \caption{
    3D Codec Avatar Rendering Overview.
    Given generic expression features and \smpl features as input, we employ a face and body adapter to map to the driving signals required by the 3D Full-body Codec Avatar decoder.
    The decoder produces a set of Gaussian primitives that encode the geometry and appearance of the avatar.
    Finally, images are rendered based on the Gaussian primitives using a tile-based Gaussian splatting renderer.
    }
    \label{fig:ca_overview}
\end{figure}
In addition to the 2D rendering discussed in the previous section, we connect the dyadic motion model to 3D Full-body Codec Avatars \citep{bagautdinov2021driving,martinez2024codec} to enable free-viewpoint rendering; see \Cref{fig:ca_overview} for a pipeline overview.
This requires a face and body adapter to map from the generic expression features and \smpl body features that are the output of the dyadic motion model to the person-specific expression features and the full-body avatar skeleton rig that is expected as input by the 3D Full-body Codec Avatar decoders.
We employ 3D Full-body Codec Avatars based on Gaussian splatting \citep{WangARXIV2025, kerbl3Dgaussians} to achieve high-fidelity results.
In the following, we describe the underlying datasets and models that are required to train personalized Full-body Codec Avatars and map to the driving signals for the face and body.

\subsubsection{3D Full-body Codec Avatars}
\begin{figure}[H]
    \centering
    \includegraphics[width=1.0\linewidth]{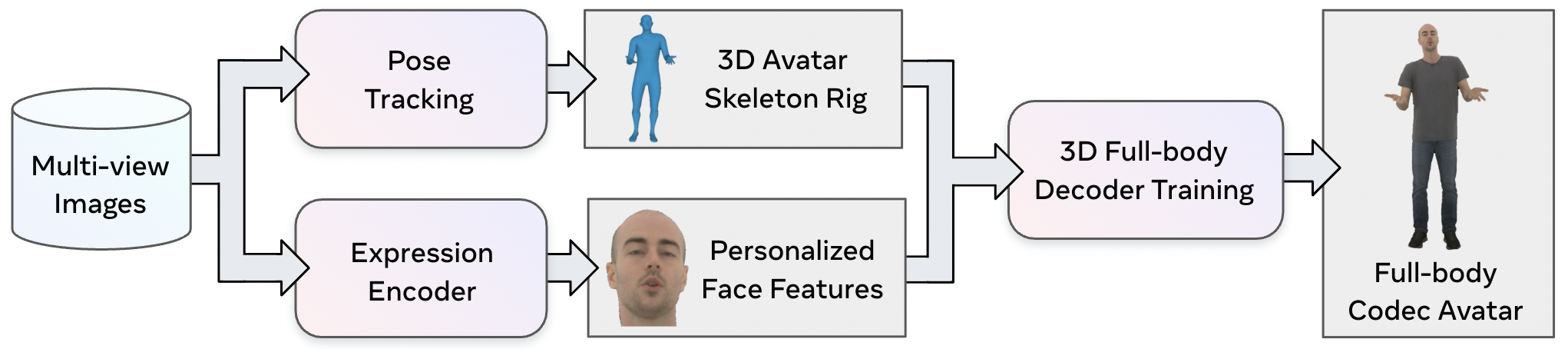}
    \caption{Codec Avatar Decoder Overview. Given a multi-view capture of an actor, we first extract per-frame 3D skeletons based on 3D keypoint detections as well as personalized expression features based on frontal view expression encoding. Given the extracted face and body features as input, the 3D Full-body Codec Avatar decoder is trained end-to-end to synthesize the multi-view images in the dataset by minimizing a photometric re-rendering loss.}
    \label{fig:ca_decoder_overview}
\end{figure}
In this section, we describe the datasets and models that are required to train a 3D embodiment based on 3D Full-body Codec Avatars. \Cref{fig:ca_decoder_overview} provides an end-to-end overview of the entire pipeline from the multi-view dataset over pose tracking and expression encoding to 3D Full-body Codec Avatar decoder training.

\paragraph{Multi-view performance capture.}
We start by collecting a dataset of our actor's shape and pose-dependent appearance based on a multi-camera capture systems with 512 synchronized and calibrated cameras \citep{WangARXIV2025}.
Each of the cameras has 24 mega-pixel resolution and records videos at $30$Hz.
The multi-view capture system has a radius of $2.75$ meters, which is large enough to collect the full range of human motion.
For each actor, we collect around $30$K frames of training data covering a large variety of facial expressions and body poses.
The resulting dataset contains synchronized and calibrated multi-view images $ \{ I^k_f \} $ with $k$ being the camera index and $f$ being the frame index.

\paragraph{Pose tracking and mean shape estimation.}
Given the set of multi-view images $I^k_f$ as input, we first extract 2D keypoints and foreground-background segmentation masks for all images \citep{khirodkar2024sapiens}.
Next, we triangulate the per-camera 2D keypoint detections to obtain a set of 3D per-frame keypoints.
We use inverse kinematics (IK) to fit the avatar skeleton rig pose parameters $s_f$ to the per-frame 3D keypoints.
In addition to the avatar skeleton, we also estimate the actor's canonical mean shape mesh.
We employ multi-view 3D reconstruction to obtain per-frame 3D point clouds for a subset of frames that contain peak poses.
Next, we solve for the canonical template mean shape mesh that best fits all point clouds when deformed using linear blend skinning (LBS) based on the known tracked avatar skeleton pose parameters $s_f$.

\paragraph{Person-specific expression code estimation.}
To model facial expressions, such as the shape of the mouth, position of the eyes, and the motion of the eye brows, we employ a set of per-frame expression features.
The expression features $e_f \in \mathbb{R}^{128}$ are computed by first selecting one frontal view image for each frame of the multi-view capture.
The selected frontal view images are then processed by the facial expression encoder that has been introduced in \Cref{sec:rep_imitator_latent} to extract $e_f$ for each frame.

\paragraph{Codec Avatar decoder training.}
\begin{figure}[H]
    \centering
    \includegraphics[width=1.0\linewidth]{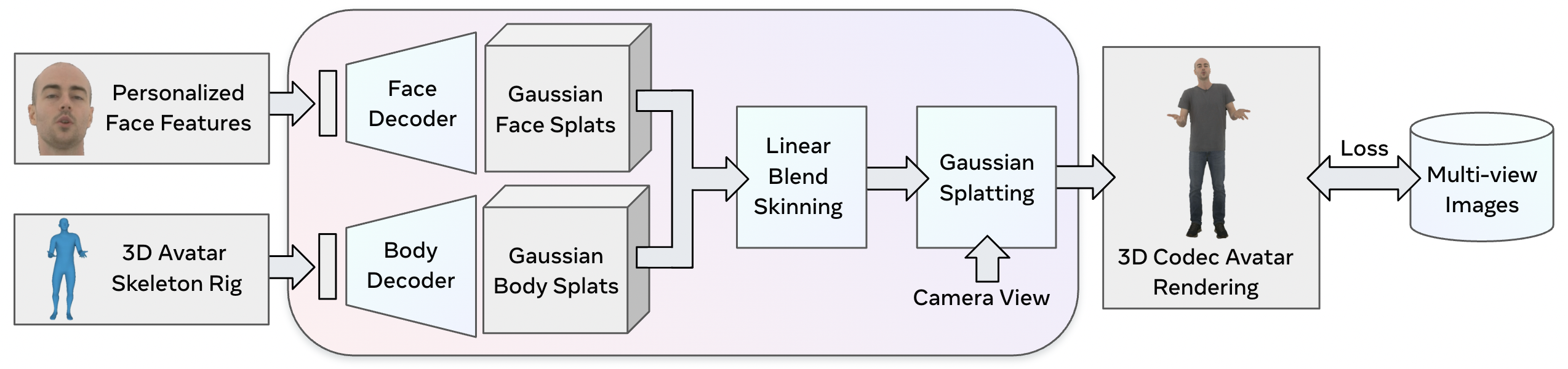}
    \caption{
    Codec Avatar Decoder Details.
    3D Full-body Codec Avatar decoders are trained end-to-end using a photometric loss with respect to the captured multi-view image dataset of an actor.
    The decoder is a de-convolutional neural network with a face and body decoder branch that each output a set of Gaussian primitives that are combined and then posed via linear blend skinning (LBS), before finally being rendered to the screen via Gaussian splatting.
    }
    \label{fig:ca_decoder_details}
\end{figure}
Given the multi-view images $I_f^k$, avatar skeleton rig pose parameters $s_f$, and expression codes $e_f$ as input, we train a 3D Full-body Codec Avatar decoder.
The 3D Full-body Codec Avatar decoder is a de-convolutional neural network with a face and body decoder branch (see \Cref{fig:ca_decoder_details}).
The face decoder takes as input the per-frame expression codes $e_f$ and maps them to a tensor $G_{face}(e_f) \in \mathbb{R}^{N_w \times N_h \times N_c}$ that stores the parameters associated with the set of decoded Gaussian primitives.
Here, $N_w^{f}=512$ and $N_h^{f}=512$ defines the number of Gaussians along the width and height of the body template's uv-map leading to $262$k Gaussian face primitives.
The number of parameters per Gaussian primitive is $N_c=38$ to parameterize its position ($3$), orientation ($4$), scale ($3$), opacity ($1$), and appearance ($27$).
The geometry and appearance of our Full-body Codec Avatars is modeled as the composition of a set $G = \{g_i\}_{i=0}^{N-1}$ of $N$ Gaussian 3D primitives $g_i$.
The Gaussian primitives $g_i=\{t_i, R_i, s_i, o_i, c_i^k \}$ model both the face and body geometry as well as their view-dependent appearance.
Here, $t_i \in \mathbb{R}^3$ is the position of the primitive in unposed model space, $R_i \in \mathcal{S}\mathcal{O}(3)$ its orientation, $s_i \in \mathbb{R}^3$ the per-axis scale factor, and $o_i \in \mathbb{R} $ its opacity.
Together, these parameters describe the 3D Full-body Codec Avatar's face and body shape in canonical space.
The $c_i^k \in \mathbb{R}^3$ are coefficients of 2\textsuperscript{nd}-order spherical harmonics and parameterize each Gaussian's view-dependent appearance.
The body is decoded by a decoder that outputs the tensor $ G_{body}(s_f) \in \mathbb{R}^{N_w^{body} \times N_h^{body} \times N_c}$.
Here, $N_w^{body}=1024$ and $N_h^{body}=1024$ lead to $1$M Gaussian primitives.
Finally, the face and body Gaussian primitives are combined.
The final step is to transform the Gaussian primitives from canonical space to deformed space using linear blend skinning (LBS) based on the known tracked avatar skeleton pose parameters $s_f$ before being rendered into the camera view using Gaussian splatting \citep{kerbl3Dgaussians}.
Given the decoded Gaussian primitives $g_i$ in world space and the user's viewing direction $v$ as input, we employ volume rendering to generate the final images.
We first evaluate the color of each Gaussian as $c_i = \sum_{k=0}^{9}{c_i^k \mathcal{S}\mathcal{H}_k(v)}$.
We then determine all Gaussians that overlap each of the image tiles and sort them in front-to-back order based on their depth with respect to the camera.
For each pixel $p$ within the tiles, we compute its color $c(p) = \sum_{i \in S(p)} { c_i o_i T(i) }$ based on front-to-back alpha compositing.
Here, $S(p)$ is the set of indices that correspond to the front-to-back sorted Gaussians overlapping pixel $p$ and $ T(i) = \prod_{j=0}^{i}{(1-o_i)} $ is the transmittance function that takes the opacity of all closer Gaussians into account.
3D Full-body Codec Avatar decoders are fully differentiable with respect to their inputs and thus can be trained end-to-end using a photometric $\ell_2$-loss with respect to the captured multi-view image datasets of the actors.
We use Adam \citep{2015-kingma} and train for $300$k iterations using $4$ GPUs with a batch of $4$ randomly sampled images.

\subsubsection{Codec Avatar Face Adapter}

\begin{figure}
    \centering
    \includegraphics[width=1.0\linewidth]{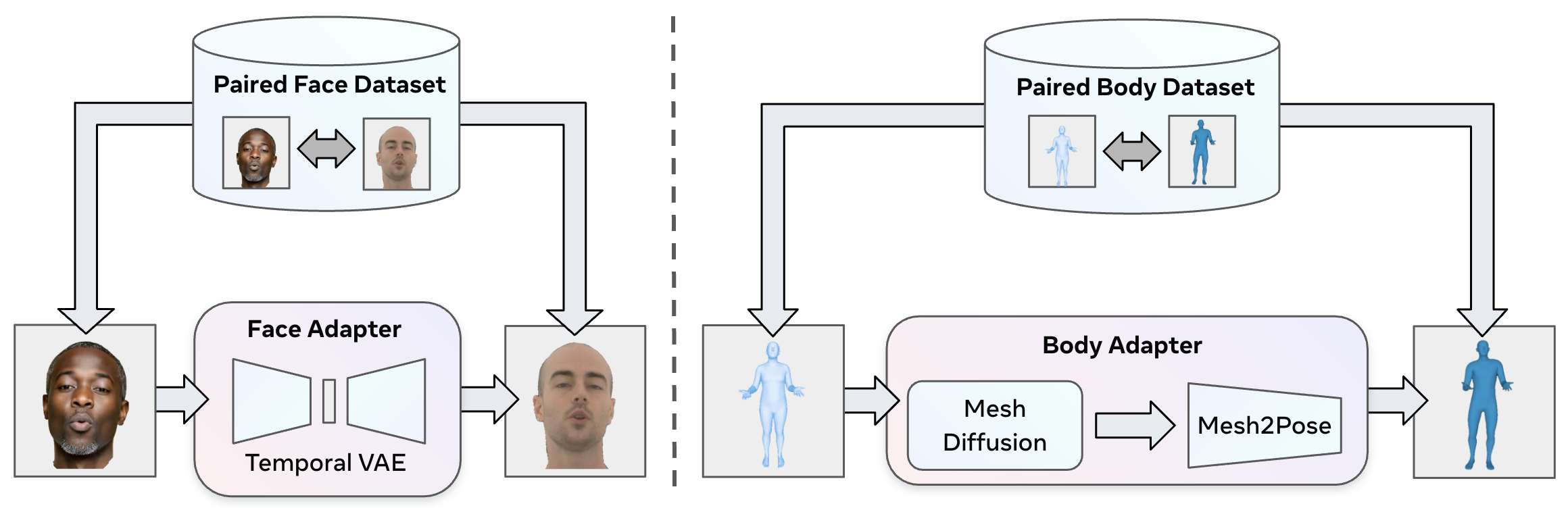}
    \caption{Codec Avatar Adapters. \textbf{Face Adapter (Left).} We create a paired dataset of generic and person-specific face expression codes, and then train a temporal VAE to map from generic to person-specific expressions. The output can be consumed by the 3D Full-body Codec Avatar decoder to render the 3D face embodiment. \textbf{Body Adapter (Right).} We create a paired dataset of \smpl and the corresponding Codec Avatar skeleton rig pose parameters, and then train a temporal diffusion model to map from one representation to the other. The output can be consumed by the 3D Full-body Codec Avatar decoder to render the 3D body embodiment.}
    \label{fig:ca_adapters}
\end{figure}
Our Codec Avatar decoders are person-specific full-body digital humans that have been trained with personalized inputs.
The face expression features produced by the dyadic motion model, however, are generic, non-personalized expression codes.
These features do not perfectly disentangle expression and identity, causing moderate identity information to persist in the latent expression space.
As a result, the generic expression space has multiple distinct features to represent the same facial expressions based on different regions of the latent space due to appearance and identity bleed.
For the Codec Avatar decoder, which is trained only on a small amount of person-specific data, this poses a challenge due to the domain gap between the two modalities.
In training, the decoder has never seen expression examples from an expression space region outside of the single identity on which it has been trained.
Thus, using the generic expression features directly to drive the 3D Full-body Codec Avatar's facial expressions would result in out-of-domain examples and uncanny facial expressions and rendering artifacts.

To reliably create authentic facial expressions, all expression inputs should ideally fall into the expression subspace of the single identity on which the decoder has been trained.
We bridge this gap by training a model to map from generic expression features to person-specific codes, which are within the person-specific decoder domain by design (see \Cref{fig:ca_adapters}, left).

\paragraph{Dataset creation.}
We first create a dataset of expression pairs of arbitrary people and person-specific expressions of our 3D Full-body Codec Avatar.
To this end, we use $800$K images from a 2D face dataset, compute their (non-personalized) expression features, and re-render the expressions with the Codec Avatar's frontal face appearance using the 2D rendering pipeline.
From these renders, we extract expression codes which now fall only into the subspace of expressions of the specific decoder identity.
The result is a paired dataset of generic, non-person-specific expressions features with the corresponding person-specific expression codes that are expected as input by the decoder.

\textbf{Face Adapter Model.}
We train a model that maps from generic expression features to person-specific expression codes (\Cref{fig:ca_adapters}, left).
Since the decoder is trained only on a small amount of data, it is crucial that the face adapter should produce temporally coherent personalized expression codes and should suppress expression outliers that the decoder cannot handle.
We therefore model the adapter as a temporal variational autoencoder (VAE) that maps from generic expression features to person-specific latent codes using 1D-convolutions.
We empirically find this architecture to produce higher quality expressions than frame-wise face adapters.

\subsubsection{Codec Avatar Body Adapter}
The dyadic motion model outputs a \smpl body representation that is not compatible with 3D Full-body Codec Avatars.
First, \smpl itself is a low resolution rig, whereas Codec Avatars are built upon a higher-resolution avatar rig.
Second, the specific \smpl output produced by the dyadic motion model only generates upper body motion but ignores lower body and leg motion, and only produces fixed body size and shape.
Additionally, since the \smpl features are extracted from monocular video, depth ambiguity and occlusions lead to inaccurate 3D features. These do not affect 2D rendering from the same camera viewpoint but become visible in 3D rendering, where the camera viewpoint can change.

All these aspects together result in an information deficiency of the \smpl body representation, more specifically: lower rig resolution, missing lower body motion, ambiguities from monocular tracking, and required motion adjustments to fit the size of the Codec Avatar.
We bridge this gap based on a generative body motion adapter that infuses the missing information (\Cref{fig:ca_adapters}, right).
We start by curating a paired \smpl to 3D Full-body Codec Avatar dataset, and then train a generative diffusion model to map from the information deficient \smpl representations to the Codec Avatar skeleton rig.

\paragraph{Dataset creation.}
We use a dataset of densely tracked 3D full body meshes from a multi-view camera dome \citep{bagautdinov2021driving}, similar to the full-body data of \citep{martinez2024codec}.
The dataset includes 128 participants performing various motions for about $45$ minutes each.
We extract frontal view images and then extract \smpl features similar to the features obtained from the \mosaic dataset and then additionally track the 3D Codec Avatar skeleton rig based on the multi-camera data.
The pairing of the \smpl features to the 3D Codec Avatar rig parameters is used as training data for the generative body adapter model.

\paragraph{Body adapter model.}
The body adapter consumes, as input, \smpl joint angles for the upper body and hands, and produces, as output, full-body joint angles for the high-resolution Codec Avatar skeleton rig.
In this process, lower body motion needs to be synthesized, high-resolution rig-details need to be inpainted, and movements need to be re-targeted to a rig with differences in bone lengths.

Since all of this requires infusing information not present in the \smpl input, we formulate the problem as a temporal generative model.
A lightweight temporal diffusion transformer with four layers first generates sparse Codec Avatar meshes from the \smpl parameters given additionally target bone lengths as input. Then, an MLP regresses joint angles for the Codec Avatar skeleton rig from the output mesh.
The model employs a limited self-attention window of $9$ frames per layer, resulting in a total receptive field of $1.2$ seconds.
We train the diffusion model with a simple $ \ell_2 $-loss on vertices in Codec Avatar mesh space, and train the mesh-to-pose regressor with an $ \ell_2 $-loss in joint angle space.

\begin{figure}[t]
    \centering
    \includegraphics[width=1.0\linewidth]{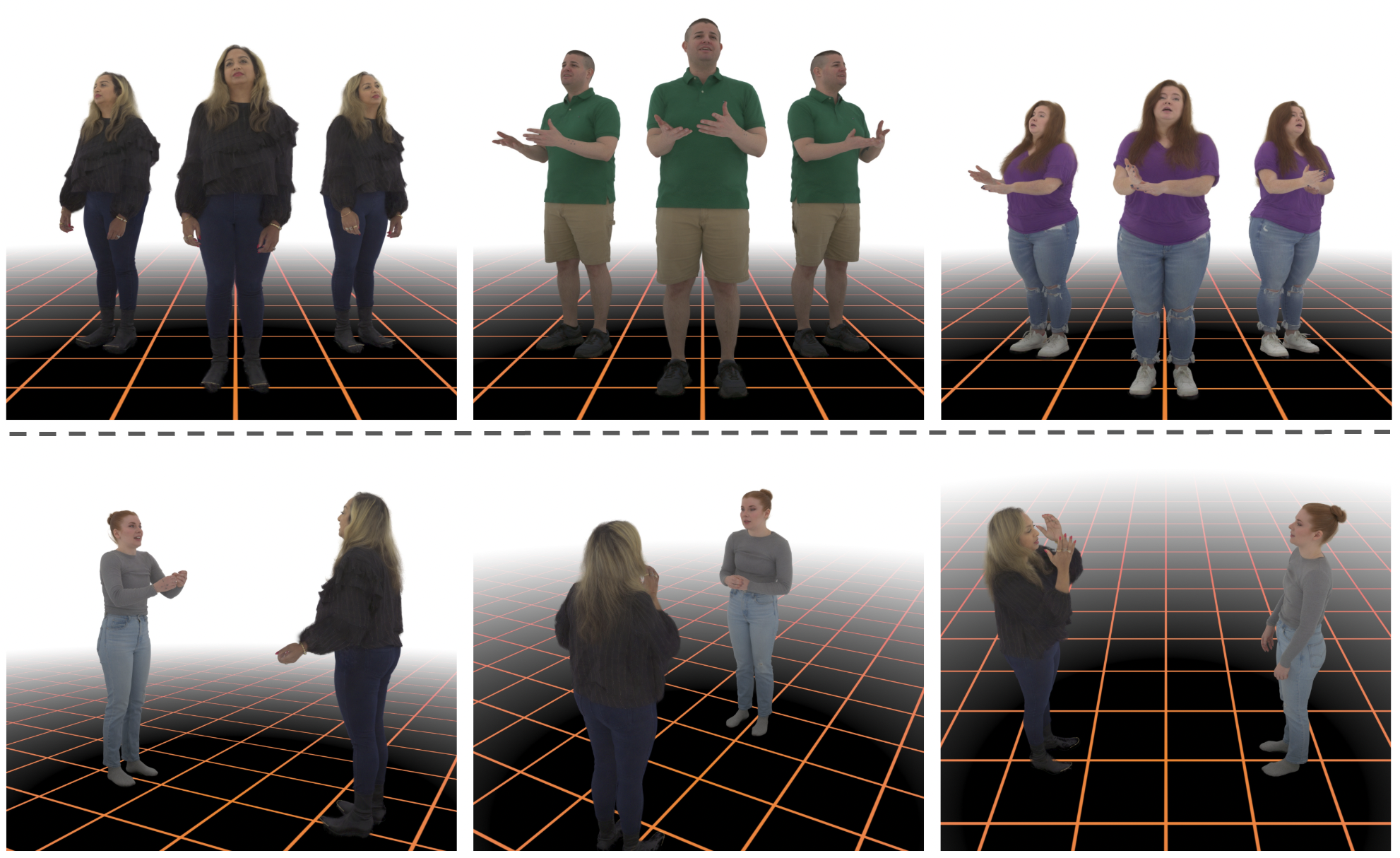}
    \caption{
    3D Full-body Codec Avatar Decoders.
    \textbf{Top:} Our full-body decoders are of high fidelity and their face/body can be animated based on the motion features produced by the dyadic motion model.
    Since the decoders are fully in 3D, they can be rendered from arbitrary camera viewpoints.
    \textbf{Bottom:}
    Since our decoders are fully in 3D, we can place two 3D Full-body Codec Avatar decoders into a shared world space coordinate system to simulate a dyadic conversation.
    In addition, the dyadic conversations can be rendered from arbitrary camera viewpoints.
    }
    \label{fig:ca_results_decoders}
\end{figure}
\subsubsection{Codec Avatar Results}
We train several 3D Full-body Codec Avatar decoders ($3$ male and $4$ female) from the collected multi-view performance capture data.
Each decoder is trained for $300$K iterations using $4$ NVIDIA H100 GPUs, which takes approximately 1 day.
A subset of the trained decoders is shown in \Cref{fig:ca_results_decoders}.
Our 3D Full-body Codec Avatar decoders are fully in 3D and completely model the actors, including their heads, hair, hands, arms, clothing, and the rest of their bodies.
The decoders are high fidelity and fully drivable; i.e., their face and body can be animated using the proposed face/body adapters that convert from the motion synthesized by the dyadic motion model to the driving representation expected by the decoder.

The 3D nature of these decoders allows us to render them from arbitrary camera viewpoints as can be seen in \Cref{fig:ca_results_decoders} (top).
The diversity of the decoders in terms of face/body shape, hair style, and clothing is to be noted.
We also demonstrate that two 3D Full-body Codec Avatar decoders can be placed into a shared world space coordinate system to render multiple actors in the same space.
Additionally, by placing two decoders into a shared world space coordinate system and making them face each other, we can simulate a dyadic conversation between two avatars.
Since the decoders are fully in 3D, the dyadic conversations can be rendered from arbitrary camera viewpoints (see images based on a circular camera trajectory around the two avatars shown in \Cref{fig:ca_results_decoders}, bottom).
These renderings highlight the spatial relationship between the two actors and create the sense of a real conversation in a shared space.
Both decoders are driven by the outputs of the face/body adapters and feature expressive face and upper body motion based on the employed dyadic motion model that produces realistic hand/arm gestures.

\subsubsection{Limitations}
\begin{figure}[t]
    \centering
    \includegraphics[width=1.0\linewidth]{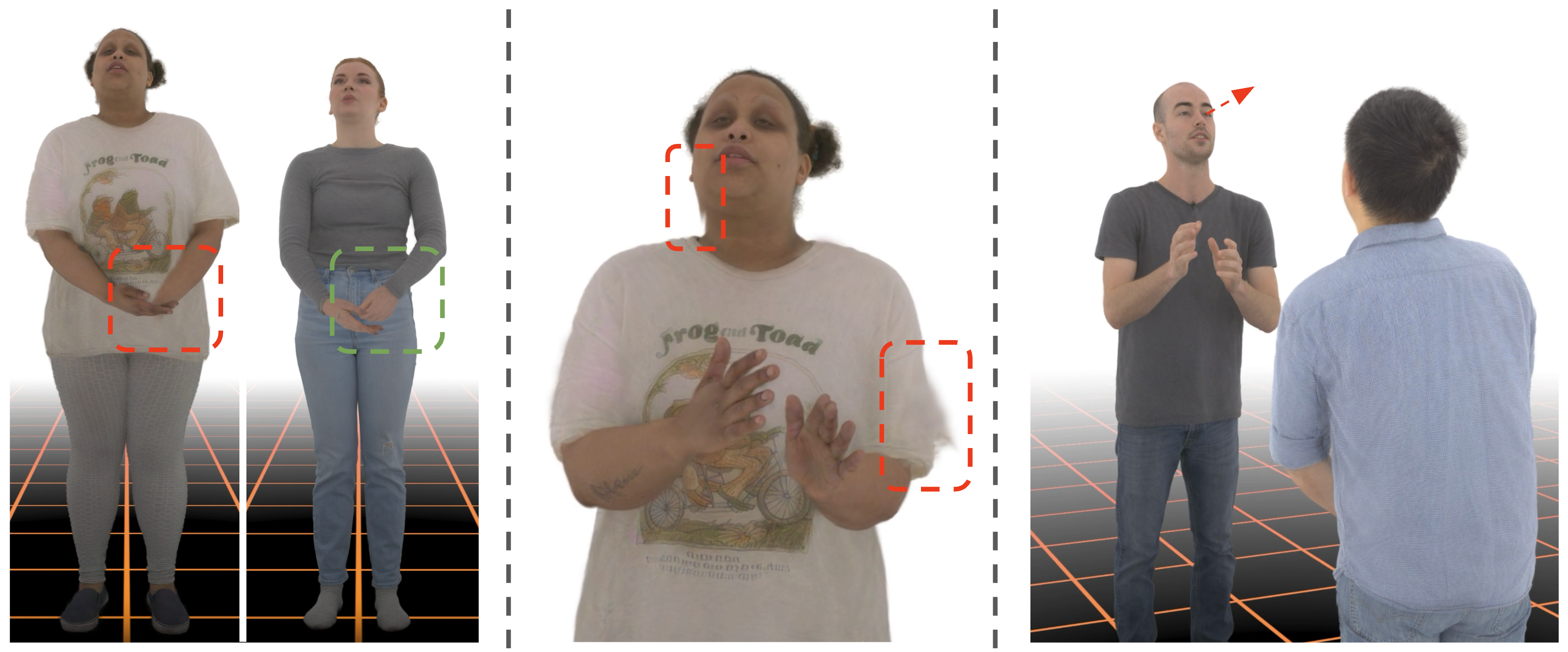}
    \caption{
    Limitations.
    \textbf{Left:} The dyadic motion model is unaware of person-specific body shapes. Therefore, self-penetrations can happen for some decoders while other decoders work perfectly fine.
    \textbf{Center:} Capacity limitations of the model lead to artifacts in clothing and appearance.
    \textbf{Right:} In the dyadic 3D rendering, eye contact is not explicitly established between the actors.
    }
    \label{fig:ca_limitations}
\end{figure}
While we show promising neural rendering results for drivable 3D Full-body Codec Avatars, some limitations still remain to be addressed in future work, some of which are unique to the 3D rendering setting.

First, neither the dyadic motion model nor the body adapter are currently aware of the actor's body shape beyond their bone length.
This can lead to self-interpenetration; e.g., hand-hand interpenetration or the arms moving partially inside the torso if the regressed body motion is incompatible with the actor's body shape (\Cref{fig:ca_limitations}, left).
This can be resolved by training body shape-aware motion models.

Second, our decoders are per-frame models and employ linear blend skinning (LBS) to explicitly model the articulated motion of humans by deforming a template mesh with tightly attached Gaussian primitives.
While our decoder also produces per-Gaussian position correctives that allow to deviate from the LBS motion, our models still struggle with large non-articulated motions, such as the dynamic motion often observed for long hair and loose clothing (\Cref{fig:ca_limitations}, center).
Further research on temporal decoders as well as multi-layer representations that model body and hair/clothing motion independently is needed to improve these results.

Finally, the training data for the dyadic motion model are two independent monocular videos that are temporally in sync, but not spatially calibrated with respect to each other.
Thus, the motion model has no awareness of the spatial relationship between the two interlocutors.
As a result, when two 3D Full-body Codec Avatars are placed facing each other in 3D space, they will not necessarily establish eye contact (\Cref{fig:ca_limitations}, right).
This can be resolved in the future by training the dyadic motion model based on 3D motion data that captures the spatial relationship between both interlocutors in a shared world space coordinate system.

%% file: eval/arxiv.tex
\section{Evaluation Methodology and Experiments}
Evaluating dyadic behavior in embodied agents remains an unsolved problem. However, there is increasing interest in this area, as demonstrated by the recent series of GENEA Challenges and leaderboards, as well as ongoing efforts to develop automatic metrics \citep{Youngwook2022,Kucherenko2023,nagy2024genealeaderboardextended}. In the following section, we present a series of calibration experiments intended to build further traction on the question of how to measure dyadic-interaction quality. 

We start by presenting an approach to human subjective (user and annotator) studies that incorporates face- and body-dyadic protocols. Next, we describe a series of ablation studies using automatic metrics commonly examined by the community. Finally, we explore the relationship between a pair of face-centric automatic metrics and the corresponding human subjective results.

\subsection{Human Studies}
We introduce two user-evaluation protocols: a \textit{face-dyadic} protocol, which emphasizes facial expressions and head movements in photorealistic renderings, and a \textit{body-dyadic} protocol, which focuses on visual behavior involving overall pose, hands, arms, shoulders, and head, but without facial rendering.

In both studies, we use a pairwise approach where participants view two dyads side-by-side. Each pair consists of an \textit{Anchor} and a \textit{Candidate}. The \textit{Anchor} videos provide the full context of an interaction, while the \textit{Candidate} videos, which include either rendered ground-truth (GT) or model-based generations, are the focus of human ratings. Participants are asked to provide a preference rating for each pair of stimuli, choosing from five possible values: \{-2, -1, 0, 1, 2\}, corresponding to \{``Much prefer A'', ``Slightly prefer A'', ``Tie'', ``Slightly prefer B'', ``Much prefer B''\}.

Our work differs from previous dyadic studies in several essential ways. First, our \textit{Anchor} stimuli include actual RGB video footage of one of the participants from the real dyadic conversation. Second, instead of presenting short, 10-second segments, as done in the speaking-focused studies reported in \cite{Kucherenko2023}, we present 20-second segments that include multiple speaking turns, similar to the ``Interlocutor'' setup in \cite{Kucherenko2023}. Finally, we significantly expand the number of evaluative dimensions to cover a wide range of quality attributes for both speaking and listening behaviors.

\subsubsection{Protocol and Evaluative Dimensions}
Both the face- and body-dyadic protocols are based on 10 core evaluative dimensions. The first set of dimensions focuses on overall preferences, the second on listener-behavior preferences, and the third on speaking-behavior preferences. Before each of the three sections, participants are provided with a video player, allowing them to view the stimuli while answering questions.

These evaluation dimensions include lifelikeness, clarity of intent, turn-taking, listening and speaking. We provide the text of each protocol in its entirety in \Cref{section:protocols}.

\subsubsection{Face Dyadic Study}

\paragraph{Data.} We selected n=61 segments, each 20 seconds long, from an earlier version of the \mosaic test set. These segments were chosen to include at least 2 or more speaking turns, with the number of turns ranging from 2 to 5 within the 20-second period. A 2-second buffer is allowed at the beginning and end of each sample to ensure that speaking behavior remains within phrase boundaries. Our focus is on the visual behavior of our models during both speaking and listening, which is why each sample requires the agent to have at least one speaking turn and one listening turn. 

\begin{figure}
    \centering
    \begin{subfigure}[b]{0.49\linewidth}
        \includegraphics[width=\textwidth]{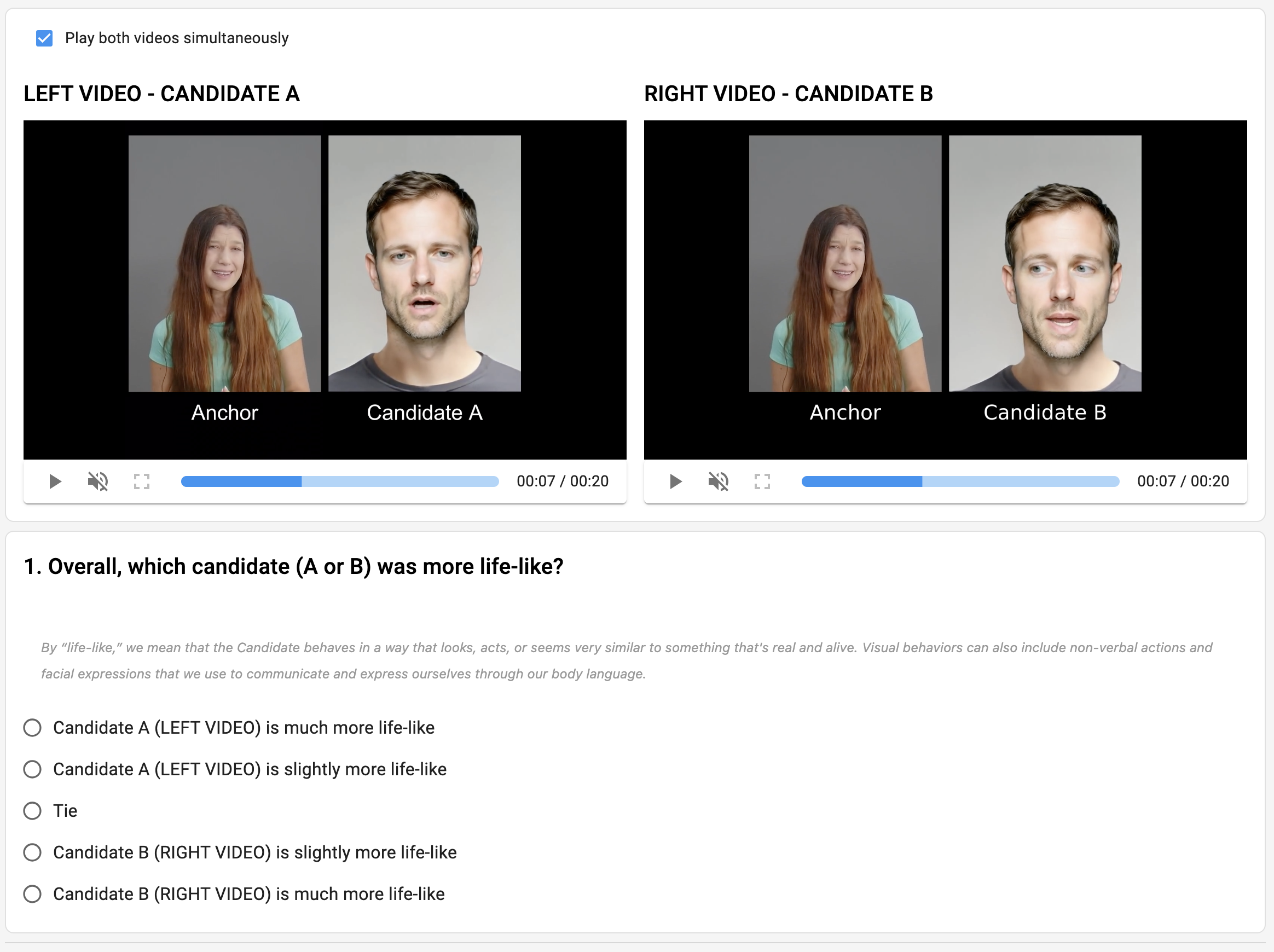}
        \caption{facial study}
        \label{fig:eval_ui_facial_study}
    \end{subfigure}
    \begin{subfigure}[b]{0.49\linewidth}
        \includegraphics[width=\textwidth]{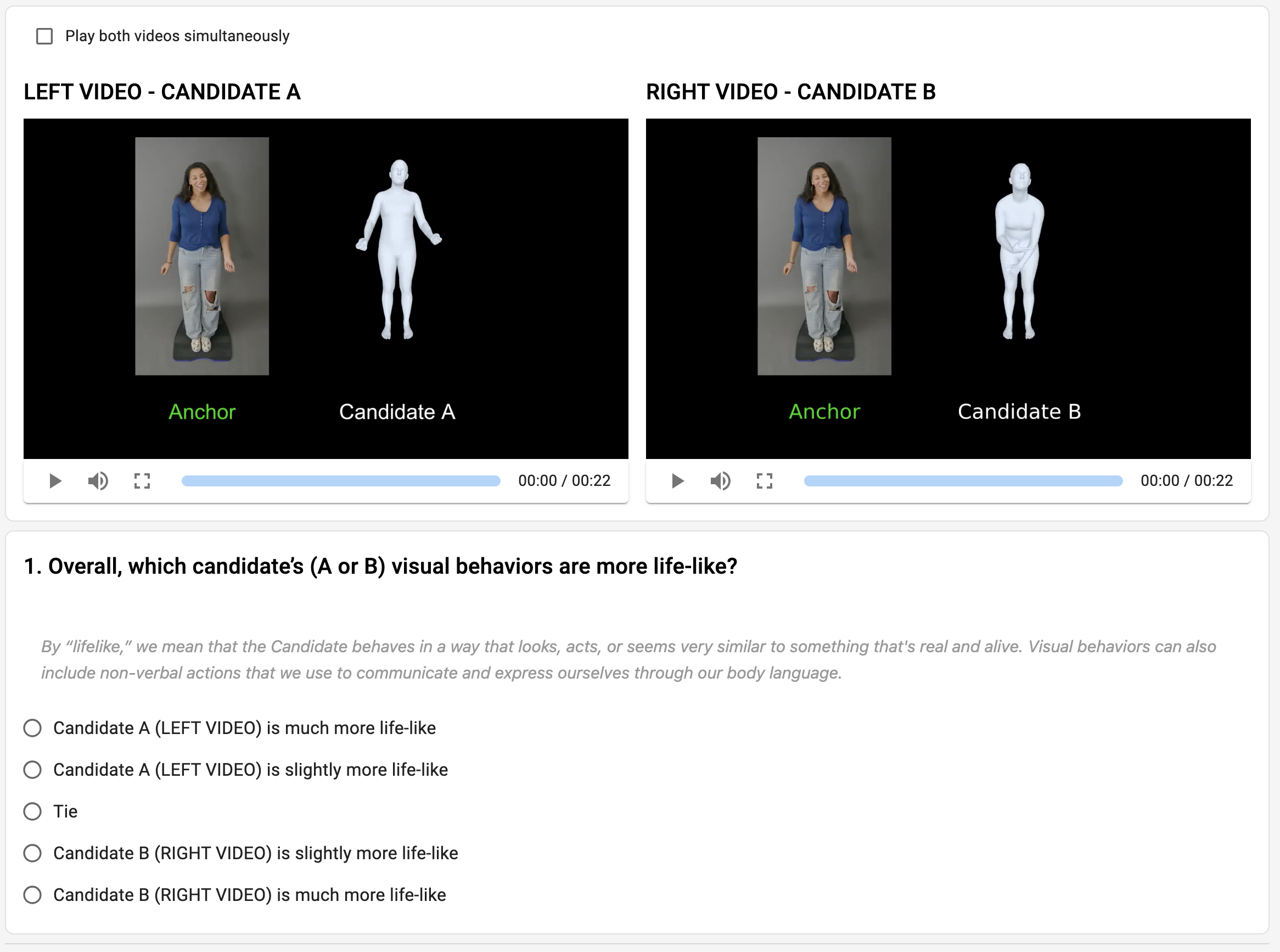}
        \caption{body study}
        \label{fig:eval_ui_body_study}
    \end{subfigure}
    \caption{Examples of actual response interfaces for body and facial studies}
    \label{fig:eval_ui_studies}
\end{figure}

\paragraph{Stimuli and ratings.} Each sample consisted of a pair of dyadic interaction videos presented side-by-side. Both videos share the same audio track, which is actual speech data from the \mosaic test-set samples. Visually, each video includes one ground-truth RGB Anchor (cropped to show information from the shoulders up) and one model- or ground truth-rendered \textit{Candidate} (see \Cref{fig:eval_ui_facial_study}). To reduce cognitive load, additional labels (Anchor and Candidate) were added to clearly identify the ground-truth interlocutor from the generated stimuli. Additionally, we use ``A/B'' labels for the \textit{Candidate} stimuli to provide clear visual cues to emphasize left and right positions, minimizing potential ambiguity.

A video synchronizing button was added, enabling simultaneous playback of videos and simplifying side-by-side comparisons if desired, however participants were also able to play the dyad videos independently.

We used the same face rendering pipeline for each \textit{Candidate} video with voice-matched gender of the renderings.

In this setup, each test item allows for a comparison within one pair of models. In total we have $n=610$ test-item pairs (61 samples times 5 choose 2 pairs). Each test-item receives five ratings from different study participants.

\paragraph{Models.} We compare four preliminary models to an actual ground-truth interlocutor renderings (see \Cref{tab:face-dyadic-models}).

\begin{table}[H]
    \centering
    \begin{tabular}{ccc}
        \toprule
         \textbf{Name} & \textbf{Model} & \\
         \midrule
         GT & Rendering of ground-truth human interlocutor & \\
         A & Dyadic Joint Face+Body with Windowed Self-attention  & \\
         B & Dyadic Joint Face+Body with Self-attention & \\
         C &Dyadic Self-attention without Expression normalization &  \\
         D &AV Dyadic Self-attention without Expression normalization &  \\ \bottomrule
    \end{tabular}
    \caption{Preliminary models used in the Face-dyadic human subjective calibration study.} 
    \label{tab:face-dyadic-models}
\end{table}

\paragraph{Participants.}

A total of n=81 participants were recruited through a third-party annotation service with the following requirements: participants needed to be native speakers of English  residing in the USA, Canada, the UK, or New Zealand. Throughput varied by participant, with the average participant providing n=37.7 ratings (min=1, max=86). Studies began with an introduction to the study including example stimuli to orient participants to the task. Participants were paid upon  completion of the study, relative to the number of rated samples.

\begin{figure}[H]
    \includegraphics[width=\textwidth]{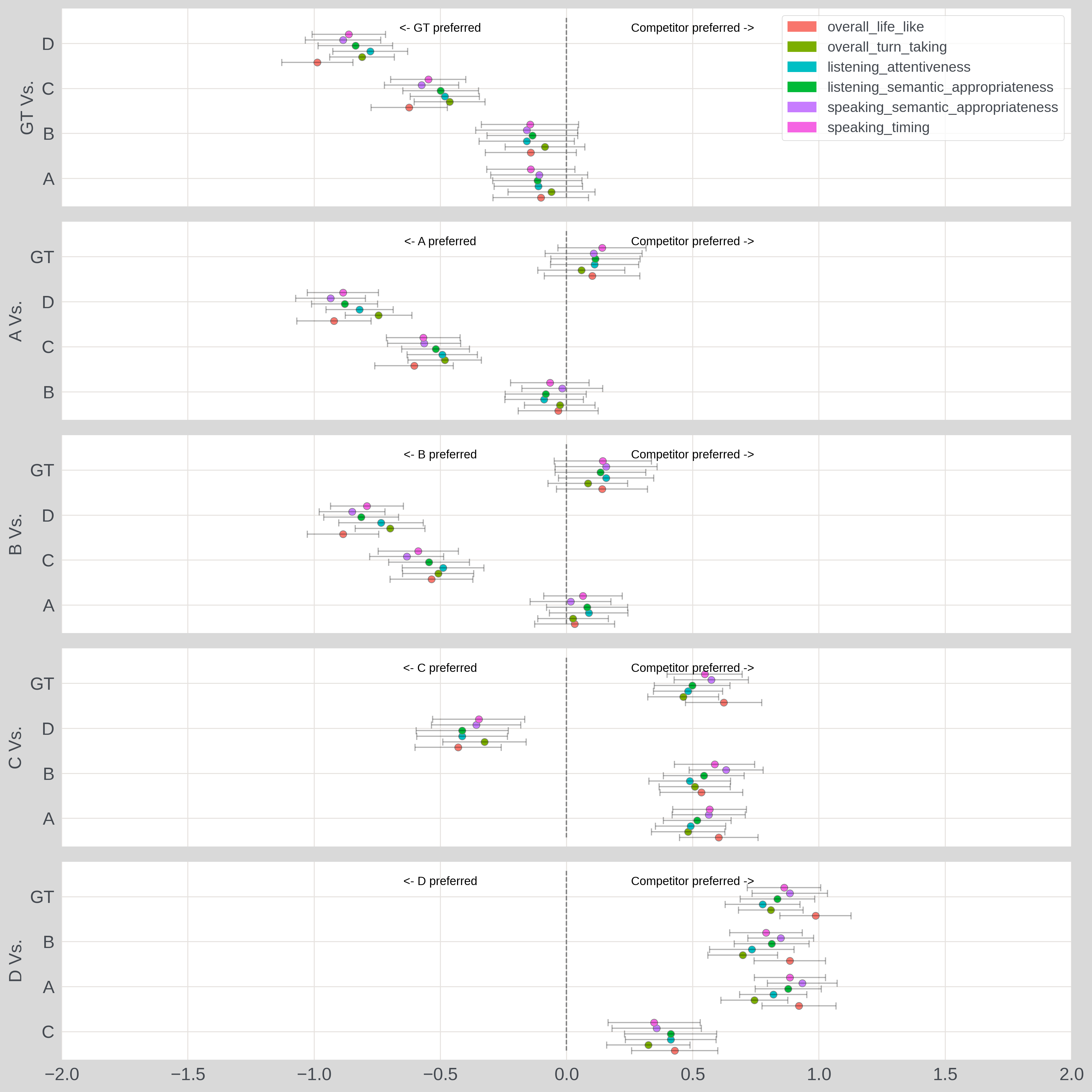}
    \caption{Face-dyadic study results. Horizontal axis displays average rating by match-up ranging from -2 (Much prefer left) to 2 (Much prefer right). Each facet shows a different match-up relative to a single model. For example the top facet shows average ratings compared to GT renderings. The vertical axis within each facet shows the competitor model for a given match-up. Error bars represent analytically computed 95\% CIs.} 
    \label{fig:human-eval-results}
\end{figure}

\paragraph{Analysis and Discussion.}
\label{section:face-dyadic-analysis}
We compute item-level scores by taking the mean preference rating across the five ratings at the item level. We then compute the average rating for each model match-up across items. \Cref{fig:human-eval-results} shows estimated mean preference scores by match-up with each facet corresponding to the average preference rating for a given model compared to the competitor set. (For readability, we only present six of the 10 evaluative dimensions in \Cref{fig:human-eval-results}, and note that results are directionally consistent across all dimensions.)

Results indicate a clear ordering across the six dimensions, in which all models are dispreferred to GT. Model A outperforms all other generations, coming close to GT performance. Model B outperforms Models C and D, but lags Model A and GT. Finally Model C outperforms Model D, the worst performing model overall. 

The finding that at least one of our models (Model A) is close to GT renderings warrants additional commentary. First, this pattern is validated by internal qualitative and quantitative user research (UXR) studies and manual inspection by the team - in many cases it is extremely difficult to distinguish between model- and GT-renderings. However, it should also be noted, similar to prior work \citep{Kucherenko2023}, that in some cases we observe artifacts that are unique to the GT rendering process and are not present in model generations. These artifacts may include what is perceived by a human evaluator as jitter, stretching, or other discontinuities between body parts (such as misalignment between the head and shoulders). Such artifacts represent confounds stemming from the underlying representation extraction and rendering process, and may be responsible, in part, for the closeness of ratings to GT. Future work will attempt to better characterize such artifacts either through manual inspection or the use of data-analytic techniques.

An additional reason why we see such close scores with human GT may stem from the level of activity of various samples. That is, we observed in prior studies that low-activity (i.e., boring) segments of actual GT human behavior is typically dispreferred when compared to more active synthetic behavior. This \textit{activeness preference} requires deeper analysis, and future work may employ various measures of activity to investigate it directly.

\subsubsection{Body Dyadic Study}
\paragraph{Data.}
A total of n=71 20-second segments were selected from an earlier version of the \mosaic test-set\footnote{These samples differ from the set used in the face-dyadic study due to differences in item-rendering workflows; however, they are both sampled from the same test-set.}.

\paragraph{Stimuli and ratings.}
Participants were again presented with two pairs of videos side-by-side. However, rather than showing the \textit{Anchor} stimuli cropped from shoulders up, they were shown the actual full-body RGB video. For the \textit{Candidate} stimuli, full-body, faceless, mesh-based renderings (SMPL-H) were presented (see \Cref{fig:eval_ui_body_study} for an example of what participants saw).

The absence of facial characteristics or lip-movements in the body-dyadic study can make it difficult to follow conversation dynamics (i.e., who is speaking). To address this, we provided an additional cue to the source of speech by highlighting the video label (\textit{Anchor} or \textit{Candidate}) with green when VAD-detected activity is present. This feature enabled participants to easily follow the conversation dynamics, thereby reducing cognitive load.
The left-facet of \Cref{fig:eval_ui_body_study} provides an example in which the \textit{Anchor} video is speaking.

\paragraph{Models.} We compare three preliminary models to a ground-truth interlocutor as well as to a baseline system consisting of negative-sampled (shuffled) ground-truth videos (see \Cref{tab:body-dyadic-models}).

\begin{table}
    \centering
    \begin{tabular}{ccc}
        \toprule
         \textbf{Name} & \textbf{Model} & \\
         \midrule
         GT & Rendering of ground-truth human interlocutor & \\
         A & Dyadic Joint Face+Body with Windowed Self-attention  & \\
         B & Dyadic Joint Face+Body with Self-attention & \\
         E & AV Dyadic Self-attention &  \\
         GT-N &Negative sample (shuffled) rendering of ground-truth human interlocutor&  \\ 
         \bottomrule 
    \end{tabular}
    \caption{Preliminary models used in the Body-dyadic human subjective calibration study.}
    \label{tab:body-dyadic-models}
\end{table}

\paragraph{Participants.} A set of n=105 participants were recruited through a third-party annotation service, with the following requirements: participants needed to be native speakers of English residing in the USA, Canada, the UK, or New Zealand. Throughput varied by participant, with the average participant providing n=33.8 ratings (min=1, max=91). Studies began with an introduction to the study, including example stimuli to orient participants to the task similar to what is shown in \Cref{fig:eval_ui_body_study}. Participants were paid upon completion of the study, relative to the number of rated samples.

\begin{figure}[H]
    \includegraphics[width=\textwidth]{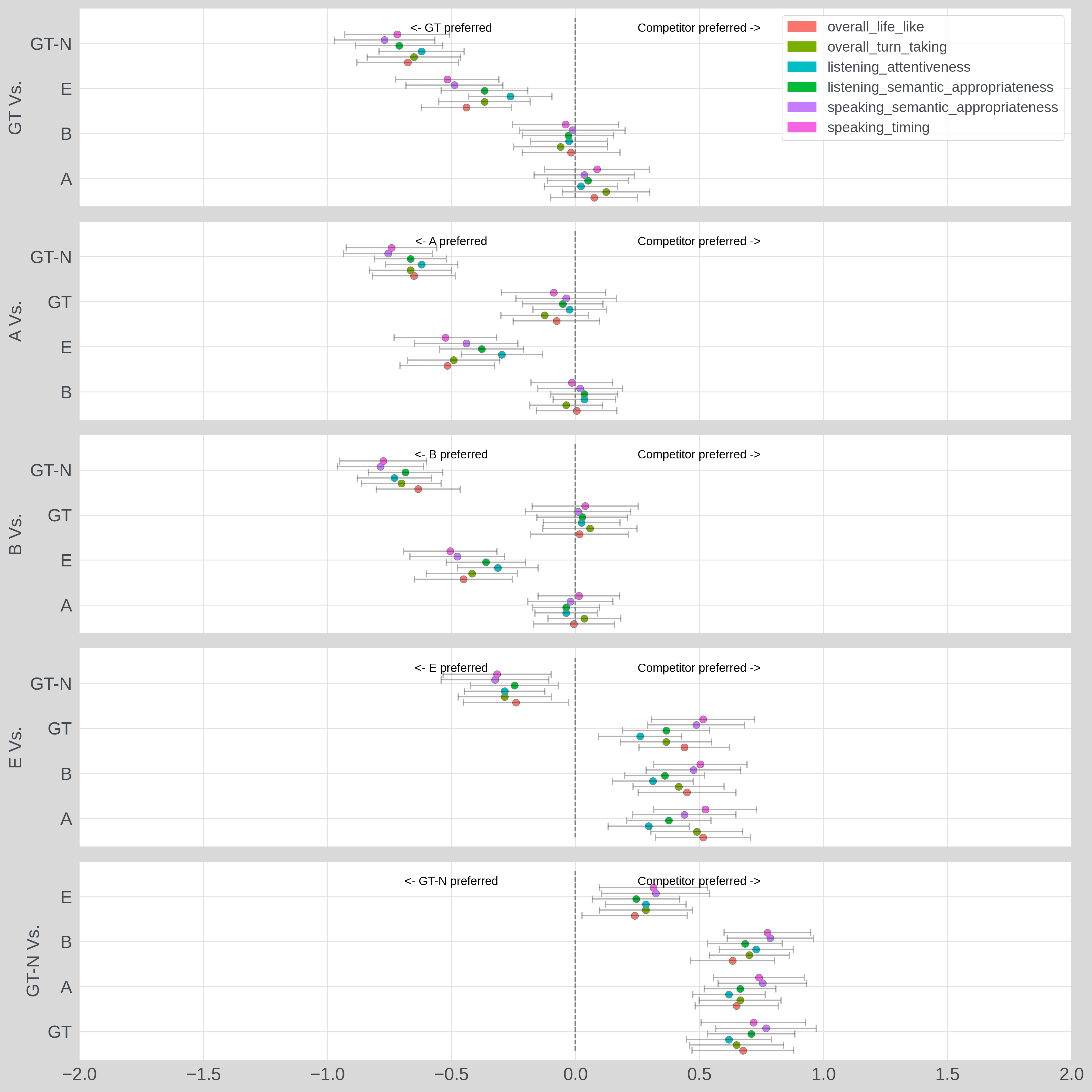}
    \caption{Body-dyadic study results. Horizontal axis displays average rating by match-up ranging from -2 (Much prefer left) to 2 (Much prefer right). Each facet shows a different match-up relative to a single model. For example the top facet shows average ratings compared to GT renderings. The vertical axis within each facet shows the competitor model for a given match-up. Error bars represent analytically computed 95\% CIs.} 
    \label{fig:body-eval-results}
\end{figure}

\paragraph{Analysis and Discussion.}
We compute item- and match-level scores as in \Cref{section:face-dyadic-analysis} and present a reduced set of evaluative dimensions in \Cref{fig:body-eval-results}. 

Results indicate that two of the models (A and B) are virtually indistinguishable from GT mesh renderings, with even a directional preference for Model A over GT. Model E only outperforms the naive baseline (GT-N) which includes randomly shuffled ground-truth behavior.

Similar to the commentary provided in \Cref{section:face-dyadic-analysis}, these findings require additional attention, as both the presence of rendering artifacts and the possibility of an \textit{activeness preference} effect may be partially responsible for the proximity to GT performance. Prior internal studies again indicate that models are extremely difficult to distinguish with GT, however we also observe  the occasional presence of body-representation artifacts from the GT rendering process, which can be subtle and pervasive. Issues such as \textit{collisions} (in which limbs such as the hands appear to merge or disappear upon contact with the body or one-another), \textit{jitter}, and \textit{skating} (in which the rendering appears to be unanchored to the ground) represent potential confounds in the current analysis. These observations underscore the importance of developing tools to detect and quantify these artifacts as essential future work.

\subsection{Experiments with Automatic Metrics}
\label{section:objective_evals}

\subsubsection{Dyadic Motion Model Analyses}

This work focuses on interactive head and body generation in a dyadic setting. In this section, we report the results of evaluations on generation quality with commonly used automatic metrics. Extensive empirical results compare various architectures and training recipes of diffusion models.

\paragraph{Test set.} A set of $200$ dyadic samples were selected from the test split of \mosaic. Each sample is a video segment of dyadic interactions, which consists of $2-5$ turns between two speakers for around $20$ seconds.

\paragraph{Models.} 
Independent face and body generations may result in inconsistent movements; for example, head and body might or moving in opposite directions. Therefore, we explored combining face and body modeling for more synchronized movements with either a cascaded approach or a joint manner. In the cascaded approach of \emph{Face2Body}, we first trained a face diffusion with dyadic audios, and conditioned the body diffusion on the face Imitator features. Similarly, we obtained the cascade of \emph{Body2Face}, where face diffusion is conditioned on the body's \smpl features. Lastly, we had \emph{Joint Face+Body}, where a single model was trained for both face and body generation. With the joint model, we conducted ablation studies of different conditions in the diffusion model: (1) Monadic: diffusion is conditioned on single-channel audio; (2) Dyadic: diffusion condition is dialogs; (3) AV: the conditions include both audios and visual features.

Diffusion models consist of 12 Tranformer layers with hidden dimension of $1024$ and feedforward dimension of $4096$. They are trained with flow-matching objective with condition dropout of $0.2$.
During inference, we apply a CFG weight of $1.5$, and use $100$ steps ODE~\citep{song2021denoising} for all experiments.

\paragraph{Automatic metrics.} To quantitatively evaluate the visual quality of face generation, we use Fr{\'e}chet Feature Distance (FFD), Lip-Sync score (Sync-C and Sync-D)  and Fr{\'e}chet Inception Distance (FID). FFD is the Fr{\'e}chet distance \citep{dowson1982frechet} of Imitator features between predictions and ground truth. Sync-C and Sync-D \citep{prajwal2020lip} assess the lip synchronization with speech. FID \citep{heusel2017gans} measures the face image quality based on features encoded by the Inception network.

As for the body, we apply Fr{\'e}chet Gesture Distance (FGD) as used in previous works \citep{genea2023}. It quantifies the discrepancy in distribution between generated outputs and real data across all samples. Diversity \citep{liu2022learning} is another metric for gesture generation, which measures the range of variations present in the generated gestures.

\begin{table}[H]
\centering
\resizebox{\textwidth}{!}{
\begin{tabular}{lccccccc}
\toprule
& & \multicolumn{4}{c}{\textbf{Face}} & \multicolumn{2}{c}{\textbf{Body}} \\
\cmidrule(lr){3-6} \cmidrule(lr){7-8}
\textbf{System} & \textbf{Condition} & \textbf{FFD} ($\downarrow$) & \textbf{Sync-C} ($\uparrow$) & \textbf{Sync-D} ($\downarrow$) & \textbf{FID} ($\downarrow$) & \textbf{FGD} ($\downarrow$) & \textbf{Diversity} ($\uparrow$) \\
\midrule
Dyadic Face2Body & A1+A2 & $0.11 \pm 0.01$ & $1.72 \pm 0.01$ & $\mathbf{7.06 \pm 0.01}$ & $1.79  \pm 0.04$ & $0.93 \pm 0.14$ & $3.83 \pm 0.10$ \\
AV Dyadic Face2Body & A1+A2+V2 & $\mathbf{0.10 \pm 0.01}$  & $1.68 \pm 0.01$ & $7.10 \pm 0.01$ & $\mathbf{1.53 \pm 0.02}$ &  $1.09 \pm 0.18$ & $ 3.82 \pm 0.08$ \\
Dyadic Body2Face & A1+A2 & $0.20 \pm 0.02$ & $2.34 \pm 0.01$ & $7.44 \pm 0.01$ & $1.81 \pm 0.04$ & $1.61 \pm 0.19$ & $\mathbf{4.02 \pm 0.05}$ \\
\midrule
\multicolumn{7}{l}{\quad \textit{Joint Systems}} \\
Monadic Face+Body & A1 & $0.29 \pm 0.01$ & $\mathbf{2.54 \pm 0.01}$ & $7.36 \pm 0.01$	& $2.27 \pm 0.12$	& $1.22 \pm 0.24$ & $3.42 \pm 0.08$  \\
Dyadic Face+Body & A1+A2 & $0.26 \pm 0.02$ & $2.48 \pm 0.01$ & $7.46 \pm 0.01$ & $1.89 \pm 0.06$ & $1.73 \pm 0.26$ & $3.91 \pm 0.03$ \\
AV Dyadic Face+Body & A1+A2+V2 & $0.26 \pm 0.01$ & $2.24 \pm 0.01$ & $7.84 \pm 0.01$ & $1.70 \pm 0.04$ & $\mathbf{0.89 \pm 0.08}$ & $3.71 \pm 0.04$ \\
\bottomrule
\end{tabular}}
\caption{Evaluation results of diffusion models on face and body generation. We run generation on the test set $5$ times and report the mean ($\pm$ standard deviation) values for each metric. (A1: model speech, A2: user speech, V2: user visual.)}
\label{tab:diffusion_results}
\end{table}

\paragraph{Results.} 
\Cref{tab:diffusion_results} reports automatic metrics of various diffusion models which use standard self-attention to attend to input conditions. For face generation, \emph{Dyadic Face2Body} achieves the best Sync-D, \emph{AV Dyadic Face2Body} gets the best FFD and FID, and \emph{Monadic Face+Body} has the best Sync-C. As for body generation, \emph{Joint AV Dyadic Face+Body} obtain the lowest FGD and \emph{Dyadic Body2Face} has the highest gesture diversity.
\begin{itemize}
    \item \textbf{Cascaded versus joint model}. Among \emph{Dyadic Face2Body}, \emph{Dyadic Body2Face} and \emph{Dyadic Joint Face+Body}, \emph{Face2Body} has the best FFD and FGD, while \emph{Body2Face} and \emph{Joint Face+Body} have comparably good performances on lip synchronization and gesture diversity.
    \item \textbf{Monadic versus dyadic input}. The comparison between \emph{Monadic Face+Body} and \emph{Dyadic Face+Body} suggests that the monadic condition helps to have slightly better lip synchronization and lower FGD, but the dyadic condition achieves better face quality with lower FFD and FID, and it also leads to more diverse gestures.
    \item \textbf{Visual input condition}. Considering \emph{Dyadic Face2Body} and \emph{AV Dyadic Face2Body}, we see that adding visual condition helps to decrease FID. Similarly for \emph{Joint Dyadic Face+Body} models, the addition of visual condition brings down face FID and body FGD.
\end{itemize}

\begin{table}
\centering
\resizebox{\textwidth}{!}{
\begin{tabular}{ccccccc}
\toprule
                       & \multicolumn{4}{c}{\textbf{Face}}    & \multicolumn{2}{c}{\textbf{Body}} \\
                       \cmidrule(lr){2-5} \cmidrule(lr){6-7}
                       & \textbf{FFD} ($\downarrow$) & \textbf{Sync-C} ($\uparrow$) & \textbf{Sync-D} ($\downarrow$) & \textbf{FID} ($\downarrow$) & \textbf{FGD} ($\downarrow$) & \textbf{Diversity} ($\uparrow$) \\ 
                       \midrule 
Standard self-attention    & $0.26 \pm 0.02$ & $\mathbf{2.48 \pm 0.01}$ & $\mathbf{7.46 \pm 0.01}$ & $\mathbf{1.89 \pm 0.06}$ & $1.73 \pm 0.26$ & $\mathbf{3.91 \pm 0.03}$ \\
Standard cross-attention   &  $0.14 \pm 0.01$ & $0.31 \pm 0.00$ & $10.25 \pm 0.02$ & $2.38 \pm 0.09$ &  $1.59 \pm 0.37$ & $3.75 \pm 0.07$ \\
Windowed self-attention  &  $0.17 \pm 0.01$ & $1.95 \pm 0.01$ &  $8.17 \pm 0.02$ & $3.26 \pm 0.06$ &  $2.52 \pm 0.50$ & $3.90 \pm 0.06$ \\
Windowed cross-attention &  $\mathbf{0.13 \pm 0.01}$ & $0.87 \pm 0.01$ & $8.92 \pm 0.02$ & $2.12 \pm 0.07$ & $\mathbf{1.25 \pm 0.37} $ & $3.82 \pm 0.08$  \\
\bottomrule
\end{tabular}}
\caption{Comparison of the architectures of conditioning in a join diffusion model. We sample $5$ generations for each test sample and report the mean ($\pm$ standard deviation) for each metric.}
\label{tab:arch}
\end{table}

\paragraph{Ablation of joint model architectures.} With Transformer as the backbone of the joint model conditioned on dyadic audios, we further experiment with different architectures for diffusion conditioning: (1) conditioning on dyadic audios via standard self-attention as used by models in \Cref{tab:diffusion_results}; (2) conditioning via standard cross-attention; (3) conditioning via windowed self-attention, which limits self-attention to a window of local context, and the intuition is that motion in one frame is more relevant to neighboring frames; (4) conditioning via windowed cross-attention. \Cref{tab:arch} demonstrates both face and body metrics. Windowed cross-attention yields the lowest FGD in body generation. Both standard and windowed cross-attention have comparable FFD in face generation, and outperform self-attention conditioning. However, they result in bad lip synchronization, as reflected by low Sync-C scores. Standard self-attention outperforms other architectures in Sync-C, Sync-D, FID as well as gesture diversity.

\begin{table}[H]
\centering
\begin{tabular}{ccccccc}
\toprule
                       & \multicolumn{4}{c}{\textbf{Face}}    & \multicolumn{2}{c}{\textbf{Body}} \\
                       \cmidrule(lr){2-5} \cmidrule(lr){6-7}
                       & \textbf{FFD} ($\downarrow$) & \textbf{Sync-C} ($\uparrow$) & \textbf{Sync-D} ($\downarrow$) & \textbf{FID} ($\downarrow$) & \textbf{FGD} ($\downarrow$) & \textbf{Diversity} ($\uparrow$) \\ 
                       \midrule

Full Imitator Latent &  \multirow{2}{*}{$0.10 \pm 0.01$} & \multirow{2}{*}{$1.68 \pm 0.01$} &  \multirow{2}{*}{$7.10 \pm 0.01$} &  \multirow{2}{*}{$1.53 \pm 0.02$}  & $1.96 \pm 0.20$  & $ 3.25 \pm 0.10$  \\
Head Rotation  &  & & &    & $\mathbf{1.09 \pm 0.18}$   & $\mathbf{3.82 \pm 0.08}$ 	  \\
\bottomrule
\end{tabular}
\caption{AV Dyadic \emph{Face2Body} conditioning ablation. We sample $5$ generations for each test sample and report the mean ($\pm$ standard deviation) for each metric.} 
\label{tab:condition_pipeline}
\end{table}

\paragraph{Ablation of cascaded models.} For the cascaded model \emph{Face2Body}, we further study what useful information face diffusion could pass to body diffusion model as tabulated in \Cref{tab:condition_pipeline}, including: (1) full imitator latent; (2) head rotation of imitator latent; The body diffusion model taking head rotation as condition reduces FGD as well as enhances gesture diversity. It indicates that face's rotation condition is most effective in aligning body with head pose, while conditioning on additional face information will hurt the quality of body generation.

\subsubsection{Analysis of Gesture Controllability}
\label{sec:sem_ges_exp}
\paragraph{Evaluation Metrics.} Regarding the effectiveness of gesture conditioning, we care about two things most: condition following and smoothness at the gesture condition boundary. 

\begin{itemize}
\item \textbf{Condition Following}. 
To evaluate the model's ability to follow gesture conditions, we compute the L2 reconstruction error between the gesture conditioning part of the generated gesture sequence and the ground truth gesture sequence. The semantic gesture condition is a sequence \(\mathcal{G}_{\text{sem}} = \{\mathbf{g}_j ^ {sem}\}_{j=0}^{T_{\text{sem}}}\)  outside the training set. We set the semantic gestures to start at the $t_{start}$ index in the generated gesture sequence.

The model generates a gesture sequence \(\mathcal{G}_{\text{gen}} = \{\hat{\mathbf{g}}_i\}_{i=1}^{T_s}\) conditioned on \(\mathcal{G}_{\text{sem}}\) for time steps \(i \in [t_{\text{start}}, t_{\text{end}}]\) and the speech condition \(\mathbf{s}\) for all \(i \in [1, T_s]\), where $t_{\text{end}} = t_{\text{start}} + T_{\text{sem}} $ . For time steps outside \([t_{\text{start}}, t_{\text{end}}]\) and $1\leq  t_{\text{start}} < t_{\text{end}} \leq  T_s$, the model is conditioned on \(\mathbf{s}\) sorely. The L2 reconstruction error is computed over the gesture conditioned part to assess the model's ability to follow the semantic gesture condition:

\[
\mathcal{L}_{\text{recon}} = \frac{1}{T_{sem}} \sum_{i=t_{start}}^{t_{end}} \|\mathbf{g}_i - \hat{\mathbf{g}}_i\|_2^2,
\]

where \(\|\cdot\|_2^2\) denotes the squared Euclidean norm. This error evaluates the fidelity of the generated sequence \(\mathcal{G}_{\text{gen}}\) to the ground truth \(\mathcal{G}\), particularly emphasizing adherence to \(\mathcal{G}_{\text{sem}}\) within \([t_{\text{start}}, t_{\text{end}}]\).

\item \textbf{Boundary Smoothness}. 
It is crucial that the motion transition is temporally smooth from gesture condition OFF to ON and ON to OFF.  
Let $\mathbf{P}(t) \in \mathbb{R}^{N \times 3}$ represent the 3D keypoint positions at time $t$, where $N$ is the number of keypoints and each keypoint has coordinates $(x, y, z)$. For a temporal sequence of length $T$, we have $\{\mathbf{P}(t_i)\}_{i=1}^{T}$ where $t_i = i \cdot \Delta t$ and $\Delta t = 1/\text{fps}$ is the temporal sampling interval.

The smoothness metric is derived from the jerk, defined as the third-order temporal derivative of position. For each keypoint $j \in \{1, 2, \ldots, N\}$ and spatial dimension $d \in \{x, y, z\}$, we compute:

$$\mathbf{j}_{j,d}(t_i) = \frac{d^3\mathbf{P}_{j,d}(t)}{dt^3}\bigg|_{t=t_i}$$

The jerk magnitude for each keypoint $j$ at time $t_i$ is computed as the Euclidean norm across spatial dimensions:

$$|\mathbf{J}_j(t_i)| = \sqrt{\mathbf{j}_{j,x}(t_i)^2 + \mathbf{j}_{j,y}(t_i)^2 + \mathbf{j}_{j,z}(t_i)^2}$$

The overall jerk metric is defined as the temporal and spatial average of jerk magnitudes:

$$\bar{J} = \frac{1}{T \cdot N} \sum_{i=1}^{T} \sum_{j=1}^{N} |\mathbf{J}_j(t_i)|$$

The final smoothness score $S$ is computed using an exponential decay function to map jerk values to a normalized smoothness measure:

\begin{equation}
    \label{eq:boundary_smoothness}
    S = \exp\left(-\frac{\bar{J}}{\sigma}\right)
\end{equation}

where $\sigma$ is a scaling parameter that controls the sensitivity of the smoothness score to jerk variations. In our implementation, $\sigma = 100$ provides empirically reasonable behavior across typical motion capture datasets. The final boundary smoothness is determined by averaging the smoothness across gesture condition transitions from ON to OFF and OFF to ON. In our experiment, the smoothness is computed using a 30-frame window, spanning 15 frames before and after each boundary timestamp.

\end{itemize}

\begin{table}
\centering
\resizebox{0.9\textwidth}{!}{
\begin{tabular}{ccccc}
\toprule
\textbf{Condition} & \textbf{Temporal Drop} & \textbf{SMPL L2 Error} & \textbf{Keypoint L2 Error} & \textbf{Boundary Smoothness} $\uparrow$ \\
\midrule
VQ IDs & 0.4 & 0.35 & 0.16 & \textbf{0.66} \\
\arrayrulecolor{gray}\midrule
\multirow{4}{*}{SMPL-H} & 0.8 & 0.05 & 0.02 & \textbf{0.61} \\
 & 0.6 & 0.06 & 0.03 & 0.54 \\
 & 0.4 & 0.04 & 0.02 & 0.54 \\
 & 0.2 & 0.03 & 0.02 & 0.48 \\
\bottomrule
\end{tabular}}
\caption{Semantic gesture control comparison and ablation. The gesture VQ ID-conditioned diffusion falls behind on the condition following with a large gap, while having better boundary smoothness compared with \smpl conditioned diffusion. Ablation on different temporal gesture conditions dropping rate indicates that a higher temporal \smpl condition dropping rate led to smoother boundary transitions. The condition following is evaluated by \smpl and keypoints reconstruction error, and the boundary smoothness is evaluated via \Cref{eq:boundary_smoothness}. }
\label{tab:ges_control_ablation}
\end{table}

\paragraph{Evaluation Results.} \Cref{tab:ges_control_ablation} shows the comparison of semantic gesture control with different conditions and dropout. 
\begin{itemize}
\item \textbf{Condition following.} 
The gesture VQ-ID conditioned diffusion falls behind on the condition following with a large gap, while having better boundary smoothness compared with \smpl conditioned diffusion. This is expected since VQ has a much higher compression rate during the quantization process. 
The condition following metric values for \smpl conditioned diffusion are in a similar scale and do not have a large difference, both visually and quantitatively.

\item \textbf{Boundary smoothness.} 
From the ablation experiment on varying temporal gesture condition dropping rates for \smpl conditioned diffusion, we surprisingly find that higher temporal \smpl condition dropping rates can lead to smoother boundary transitions. This observation is particularly evident in the visual results --- with a lower temporal \smpl dropping rate 0.4, the gesture transition at the boundary looks very sudden; while when the dropping rate comes to 0.8, the gesture transition at the boundary looks much more natural and smoother. Therefore, a high gesture condition temporal dropping rate is crucial for making \smpl conditioned outputs appear natural.

The VQ-conditioned diffusion exhibits the highest boundary smoothness. This may be attributed to the high abstraction level of VQ tokens, which provides the diffusion process with more flexibility to generate smooth motion that aligns with the overall distribution at the gesture condition boundary.

\end{itemize}

\subsubsection{Analyses of LLM-Guided Codebook Generations}
We present results of for LLM-guided codebook generation as discussed in \Cref{subsubsec:codebook}. \Cref{tab:emotion_data} summarizes the total number of tokens for valence, arousal and gesture in the dataset.

\begin{table}[htbp!]
    \centering
    \small
    \begin{tabular}{cccc}
        \toprule
         & {\bf Train} & {\bf Valid} & {\bf Test} \\
        \midrule
        Valence & 4.4M & 0.1M & 0.7M \\
        Arousal & 4.4M & 0.1M & 0.7M \\
        Gesture (w/o ``null'' token) & 22.6k & 2.2k & 2.3k \\
        \bottomrule
    \end{tabular}
    \caption{The number of emotion and gesture tokens in training, valid and test data.}\label{tab:emotion_data}
\end{table}

\paragraph{Emotion \adapter.} We set the token rate of emotion adapters as $1$ token per second, assuming emotion consistency within a one-second window. As the ground truth emotion is extracted in the frame level, we average the emotion values over all frames within one second as the training target of \adapter.

The \adapter prediction is based on acoustic and semantic information from speakers, while emotion labels are extracted from visual signals. Therefore, there might be a gap in fine-grained crossmodal emotion prediction. For example, speakers could exhibit varying intensities in facial expressions even when talking about the same thing. We group $12$ fine-grained emotion tokens into $3$ coarser grained groups. Accuracy of $3$-class prediction is the evaluation metric we report for emotion prediction. The accuracy of emotion adapters is $0.51$ for valence and $0.52$ for arousal. 

\paragraph{Gesture \adapter.} Semantic gestures fall within the long tail of the distribution of human motion. Most of the time, speakers do not make semantic gestures, and therefore the ``null'' token is the dominant gesture label associated with $98\%$ speech segments. Furthermore, the distribution varies a lot for different gestures, and precision, recall and F1 score are good metrics for imbalanced data. We compute the score for each gesture except ``null'' gesture, and report macro-averaged scores by applying the arithmetic mean to per-gesture scores. 

\begin{table}[t]
    \centering
    \small
    \begin{tabular}{cccc}
        \toprule
        {\bf Token rate} & {\bf Precision} & {\bf Recall} & {\bf F1 score} \\
        \midrule
        1 token/sec & 0.47 & 0.30 & 0.37 \\
        2 token/sec & \textbf{0.51} & \textbf{0.47} & \textbf{0.49} \\
        \bottomrule
    \end{tabular}
    \caption{Macro-averaged precision, recall and F1 score of gesture prediction.}\label{tab:gesture_adapter}
\end{table}

Empirically we tried two token rates for gesture adapter, and trained adapters with $1$ and $2$ gesture tokens per second.
\Cref{tab:gesture_adapter} reports results on gesture prediction. The adapter with token rate of $2$ gives better performance than that with token rate of $1$. This suggests that the window of one second is a bit large, cover multiple spoken words. Reducing the window size helps \adapter capture the word of interest.

We analyze examples where gesture predictions disagree with the ground truth, and find that \adapter assigns gestures to synonyms and phrases in a similar context as the word triggering semantic gestures. For example, the gesture control vocabulary contains an illustrative gesture for the word ``cold''. The \adapter also labels ``freezing'' with the same gesture.  Also it assigns the utterance ``don't do that'' with the gesture for ``stop''. These are examples lowering the precision. In other examples, when the word was spoken fast, \adapter missed prediction and got lower recall.

\subsection{Relationship Between Automatic Metrics and Evaluation Dimensions from Human Studies}

We explore the relationship between subjective human results and two automatic measures of generation quality (FFD and Sync-C) on preference data from our face-dyadic study (see \Cref{section:objective_evals} for implementation details of these metrics).

Proxy-metrics are computed for each model at the item-level. We convert these to score deltas (for compatibility with our pairwise preference data) by calculating the differences between scores for each model on a given test-item pair. 

\begin{figure}[h]
    \centering
    \includegraphics[width=0.7\textwidth]{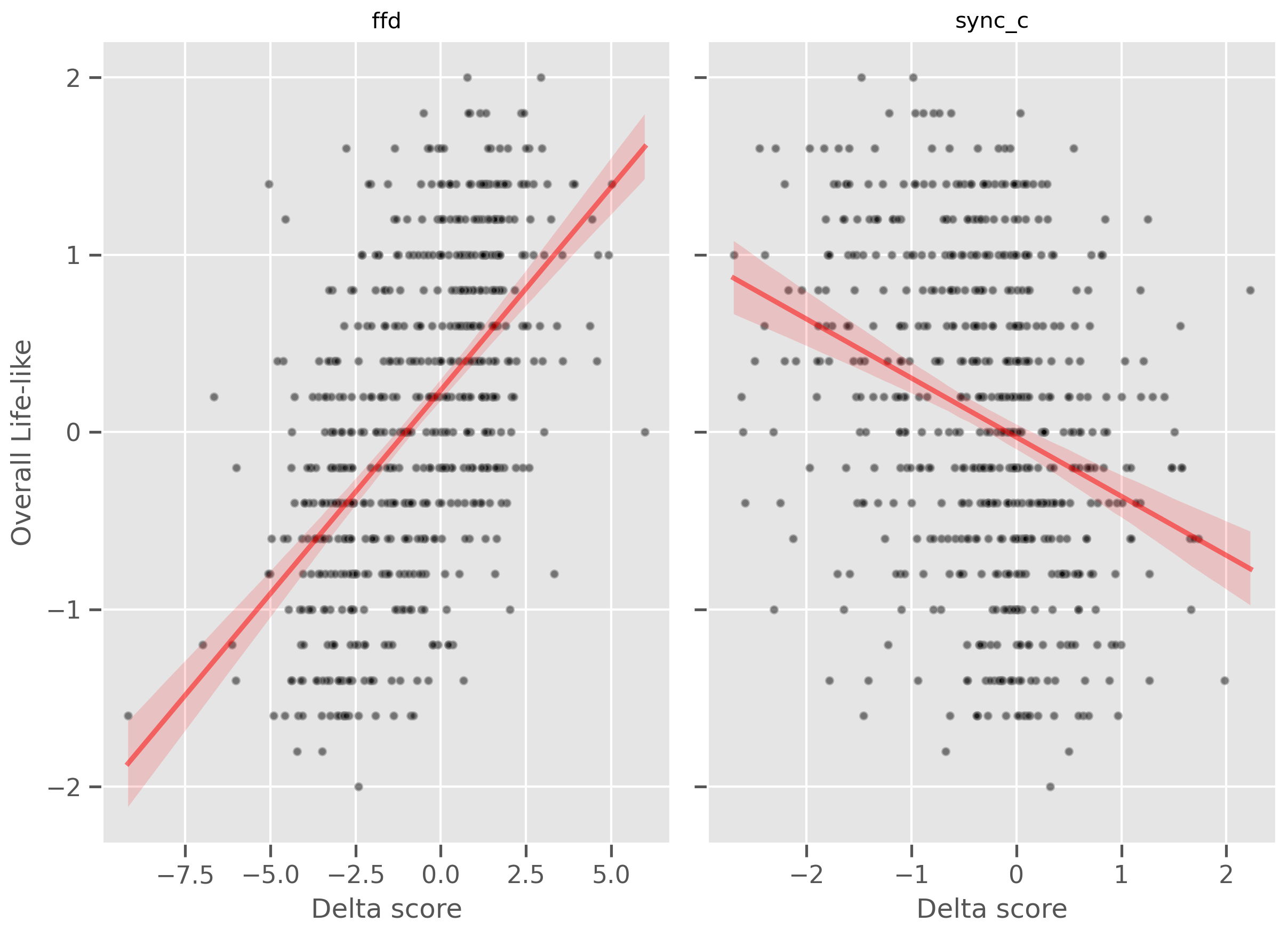}
    \caption{Comparison of Face-dyadic study subjective preferences and proxy-metrics. Vertical axis shows item-level preference for each model-pair for our Overall Lifelike dimension. Horizontal axis shows the item-level proxy-metric score deltas. Left facet shows FFD metric, and right facet shows Sync-C. 
    We quantify the relationship computing Kendall's Tao, displaying the test-statistic and p-value.
    } 
    \label{fig:face-proxy-metrics-correlation}
\end{figure}

We compare the average preference score from our subjective study with the corresponding score delta. \Cref{fig:face-proxy-metrics-correlation} plots the relationship between score deltas and human preferences for FFD and Sync-C.

A significant relationship was observed for both metrics with the ``Overall Like-like'' dimension. FFD delta scores demonstrated a significant positive correlation with human preferences. In contrast, Sync-C exhibited statistically significant negative relationships. Results are displayed in \Cref{tab:proxy-metric-correlations}.

\begin{table}
    \centering
    \begin{tabular}{cccc}
        \toprule
         \textbf{Proxy Metric}  & \textbf{Pearson's \textit{r}}  &  \textbf{Kendall's $\tau$} & \textbf{Spearman's $\rho$} \\
         \midrule
         \texttt{FFD} & 0.562 & 0.573* & 0.406* \\
         \texttt{Sync-C} & -0.307  & -0.328*  & -0.223* \\
         \bottomrule
    \end{tabular}
    \caption{Correlation of human subjective ratings for ``Overall lifelikeness'' with three automatic metrics. We present three measure of association via Pearson's $r$, Spearman's $\rho$, and  Kendall's $\tau$. ``*'' denote statistical significance at $\alpha=0.05$.}
    \label{tab:proxy-metric-correlations}
\end{table}

%% file: rai/arxiv.tex
\section{Responsible AI}

\subsection{Dataset Privacy and Ethics}
Privacy and ethical standards were critical to the \mosaic data collection effort. Systems for ensuring privacy were introduced at multiple stages of the collection --- to the participants via informed consent and constant observation by trained moderators, to the site administrators and moderators responsible for on-the-ground recordings, and finally as a part of an extensive post-collection quality assurance (QA) and filtering procedure by which every video (in total over 14,000 hours of raw footage) was analyzed for the presence of sensitive or personal material.

\paragraph{Participant education.} Participants provided informed consent for their voice and image to be collected and used. They were instructed multiple times prior to recording to avoid topics that include personally identifiable or private information (PII) and were provided guidance from moderators upon doing so. While participants were asked to respond to prompts, some of which were personal in nature, they were never compelled to do so and were free to take the conversation where they felt most comfortable (see \Cref{section:participant_preamble}).

\paragraph{Site administrators and moderators.} Site administrators and especially moderators, were educated on the nature of PII or sensitive data and the importance of indicating that it should be avoided at all times (see \Cref{section:sensitive_data_statement}). They were directed to identify and remove any content from raw deliveries, if such content did not meet our quality and safety standards.

\subsection{Quality Assurance (QA) Processes}
A quality assurance (QA) process was implemented to identify occurrences of sensitive material (PII), offensive material, and recording or content quality issues. This process included three sources of QA signals: (1) human-based video review and QA annotation, (2) text-LLM transcripts analyses, and (3) VLM-based video analysis. Based on combined signals from these three sources, flagged content was removed from the dataset.

\subsubsection{Human QA}
While human review of all 14,000 hours of raw collected footage was beyond reach, the project took as a goal that every video would have at least some portion viewed by a human. To facilitate a comprehensive analysis of video content, a stratified sampling approach was used, wherein 30-second clips were randomly extracted from each video. These clips were reviewed by human evaluators who assessed the content in accordance with a predefined guideline. To streamline the review process and ensure data accuracy, an automated data pipeline was developed, complemented by a user-friendly interface and a dashboard designed to provide real-time insights and analytical visualizations.

The human review process focused on three primary areas: (1) the mention of sensitive material (PII) or material that was offensive in nature, (2) the presence of recording artifacts, which can compromise video quality and authenticity; (3) content alignment issues, indicating discrepancies between the intended and actual content.
We provide a complete list of review areas in  \Cref{section:qa-flags}.

\subsubsection{Model-based QA}
Although human-based review provides valuable insight into quality issues, it is limited by constraints in speed and cost. To scale QA checks to every minute of raw footage, we implemented a model-based QA approach that leverages both Large Language Models (LLMs) on transcripts and Video Language Models (VLMs) on videos.

Model-based QA performance was evaluated using ground-truth human QA annotations with a primary focus on recall of sensitive (PII) information and offensive material. This approach acknowledges that false negatives (i.e., undetected instances of offensive or sensitive content) pose a greater risk to the integrity of the dataset than false positives (i.e., incorrectly flagged content). We provide a more complete list of QA-related flags in \Cref{section:qa-flags}.

Following the model-based checks, internal volunteers sampled a selection of flagged clips for further human review and validation. This additional layer of scrutiny provided a more comprehensive assessment of the automated checks' performance and helped to identify any potential false positives or false negatives that may have been missed during the initial evaluation. The goal of the LLM-based and VLM-based approaches was to scalably identify segments with potential issues - analyzing every frame and word of recorded data.

\paragraph{LLM-based transcript analysis.}
We designed a  prompting strategy to achieve high recall in detecting sensitive and offensive content. This was benchmarked against human-labeled results. This prompt configuration was then applied to the transcripts of the entire dataset to identify segments with potential issues.

\paragraph{VLM-based raw video footage analysis.}
A Video Language Model (VLM) was used to evaluate video content for quality assurance across the categories described in \Cref{section:qa-flags}. Videos from the same interaction were stacked for both the participants, as input to the VLM. A context criteria describing the different QA categories in detail was provided to the VLM as a system message and the VLM was then prompted with a structured input format, requiring it to provide output evaluations in a key-value pair format for each of the QA categories.

\paragraph{Filtering strategy.}
Model performance was evaluated using the human reviewed annotations with F1-scores for text-LLM (0.84) and VLM (0.91).  A conservative filtering strategy was employed - if a given interaction received any sensitive material flag for any of the sub-categories, from any of the three systems (Human, text-LLM, VLM) that interaction video was removed from the dataset. In total, this process removed several hundreds of hours of interaction data. \Cref{fig:qa_flow} provides an illustration of the full QA filtering process.

\subsection{Watermarking}

Another proactive measure to ensure transparency, accountability, and trustworthiness is to watermark AI-generated content. This involves adding noises to original signals (video and audio) that are imperceptible to humans but can be detected by specialized algorithms. With the rapid proliferation of generative models, watermarking becomes an essential component and is now required under multiple laws, including the US Executive Order and AI, European Union's AI Act, Labeling Rules of Cyberspace Administration of China.

We maintain our commitment to responsible AI and adopt watermarking for all of our model outputs. For audio signals, we employ AudioSeal~\citep{san-roman24a}, a state-of-the-art audio watermarking solution with a robust and efficient detection algorithm. For video content, we use VideoSeal, an open and effective video watermarking method that has been shown to perform well on high-resolution videos without compromising detection robustness~\citep{fernandez2024video}. Both AudioSeal and VideoSeal use localized and extractable watermarkings; i.e., the models (generators) were trained to embed secret messages into individual frames of the original contents, and the detectors also extract the messages on the frame level. Each secret message is a binary string of $n$ bits and can be defined by users. We use a fixed set of secret messages to differentiate contents from different sources: Human audios, LLM generated audios, and dyadic motion model video output. The cascaded model architecture allows for an easy composition of different audio and video watermark generators in the decoder in a post-hoc manner. 

We train new AudioSeal and VideoSeal models and use them for the \mosaic dataset samples dyadic model outputs. To improve the scalability for audios, the model is trained in a causal mode, so the watermark was generated autoregressively with access to only the previous frames. The context size is $1920$ frames (for $24$ kHz audios), and we can watermark a very long speech in a streaming manner without significant added latency (our experiments on the A100 GPU suggested a watermarking time of approximately 5 ms per speech token). For videos, VideoSeal introduces the concept of \emph{temporal watermark propagation}: The input video is segmented into $k$ frames, and a watermark is generated for the first frame, which is then propagated (copied or interpolated) to all subsequent frames. In practice, we find $k=4$ and a simple watermark copy provides a good trade-off between quality and robustness. Additionally, the frame is downscaled to $256\times 256$ and the watermark is later upscaled to match the input resolution, this helps increase the training efficiency. 

There are some hyperparameters that we need to choose when fine-tuning the watermark models. The first is the scaling factor $\lambda$, which decides the strength to add watermarks to the original contents. Too high $\lambda$ values make artifacts in the watermarked contents more perceptible and visible, while lower $\lambda$ values result in detectors that are less robust to watermarking attacks. In our experiments, we find $\lambda=0.25$ to produce the optimal trade-off. The other parameter is the number of bits $n$ in the watermark, for which we empirically evaluate per modality and set $n=16$ for audio and $n=128$ for video. To improve the robustness, after the contents are watermarked, we can continue to fine-tune the detector further on extended attacks and content distortion. Our experiments show that this improves the detection accuracy by up to $+37\%$ for popular audio compression attacks such as MP3 and AAC. For videos, fine-tuning on video-specific compression formats such as H.264 and H.265 also improves the performance by more than $20\%$.

%% file: related_work/arxiv.tex
\section{Related Work}

\paragraph{Existing conversational or audiovisual data.} There are a multiple spoken conversation corpora, such as Fisher \citep{cieri2004fisher} and Switchboard \citep{godfrey1992switchboard}. They have been widely used in speech research and dialogue modeling. Besides speech and text modalities, some corpora also provide video capture of human interactions. These audiovisual data resources include the AMI Meeting Corpus \citep{carletta2005ami}, IEMOCAP \citep{busso2008iemocap}, and the CANDOR corpus \citep{candor}. Finally, the \mosaic dataset, introduced in this work, is the largest known video collection of in-person, dyadic, conversation-based interactions.

\paragraph{Human motion model.} Human-centric generative modeling focuses on generating human-like and expressive motions with autoregressive model \citep{motiongpt} or diffusion \citep{peebles2023scalable, human-motion-diffusion, chen2024diffsheg}. One key task of human motion generation is audio-driven generation. Different from the strong semantic alignment between input condition and motion, audio-driven task emphasizes the synchrony between speech and motion as well as the motion appropriateness \citep{Kucherenko2023}. AniPortrait \citep{aniportrait}, HALLO \citep{hallo} and VASA-1 \citep{vasa} train diffusion models for talking face generation, capturing facial expressions and lip movements. Geneface \citep{ye2023geneface++} leverage the efficient motion-to-video renderer to achieves fast training and even real-time inference. INFP \citep{zhu2025infp} proposed models for face generation in the dyadic setting. Besides face, recent work also studies full body generation based on speech. GENEA organizes a shared task on hand and body gesture generation in both monadic and dyadic setting \citep{genea2023}. \citet{conversation-motion} makes use of both autoregressive and diffusion model to learn body and face motion respectively, generating photorealistic embodiment in conversational settings. ConvoFusion \citep{mughal2024convofusion} develops diffusion for conversational co-speech gesture synthesis. 
There is also a trend to generate face expression and body gestures together~\citep{chen2024diffsheg, talkshow}. TalkSHOW~\citep{talkshow} proposes to generate the expression and gesture separately, where the gesture model is generative and the face model is deterministic. DiffSHEG~\citep{chen2024diffsheg} is the first to propose a unified framework for modeling the joint distribution of expressions and gestures, which introduces a diffusion-based architecture with a uni-directional conditioning flow from expression to gesture, ensuring coherence and harmony between the two modalities.

\paragraph{Controllability in diffusion.} Controllability is a crucial aspect of diffusion models, as it enables users to manipulate the generation according to their preferences. Face diffusion models such as VASA-1 \citep{vasa} takes control signals such as eye gaze, head distance and emotion offsets. EMO~\citep{tian2024emo} involves a Face Locator and Speed Layer to weakly control the approximate region of the target face and the rough velocity level of the movement. Besides, EmojiDiff~\citep{jiang2024emojidiff} aims to achieve expression control with highly maintained identity.
Body diffusion models such as MDM \citep{human-motion-diffusion} and MotionDiffuse \citep{motiondiffuse} focused on text control signals, which generates motions based on textual descriptions. C2G2~\citep{ji2023c2g2} enables users to generate and edit the gestures in any time intervals, making sure the generation process is controllable. ConvoFusion \citep{mughal2024convofusion} also takes speaker style and word-excitation to guide gesture synthesis.
On the other hand, semantic gesture control in co-speech gesture generation is challenging due to its long-tailed distribution in the dataset. \citep{zhang2024semantic} proposes to predict the semantic gestures based on the text transcript, and then merge the retrieved semantic gestures with the GPT-generated co-speech gestures.

\paragraph{LLM agent.} 
With rapid advances in LLM \citep{gpt4,gemini2.5,llama4}
, there is a surge of interest in LLM-powered agent systems where LLM functions as the agent brain with general knowledge. LLM agent is an intelligent system to deal with tasks and interact with users. 
Post-training has been widely studied to extend the capabilities of pretrained LLMs in specific domains and downstream tasks. Post-training paradigms include supervised fine-tuning \citep{ouyang2022training}, prefix tuning \citep{li2021prefix}, prompt tuning \citep{lester2021power}, instruction tuning \citep{shengyu2023instruction}, and reinforcement learning \citep{trung2024reft}. In particular, the development of parameter-efficient fine-turning (PEFT) has been spurred, given its computation efficiency compared with full-model tuning. LoRA \citep{lora} and adapters \citep{hu2023llm} are cost-effective PEFT approaches which could also mitigate catastrophic forgetting.

\paragraph{2D Rendering of Photorealistic Avatars.} 
Video generation has evolved significantly over the years. Early text-to-video models based on GANs—such as MoCoGAN-HD~\citep{tulyakov2018mocogan} and StyleGAN-V~\citep{skorokhodov2022styleganv}—as well as VAE variants like VideoVAE~\citep{he2018videovae} and CV-VAE~\citep{zhao2024cvvae}, frequently struggled with mode collapse, temporal flickering, and low spatial fidelity. Recently, diffusion-based denoising models~\citep{ho2020denoising,song2021denoising,nichol2021improved,song2020score,dhariwal2021adm} have supplanted these approaches thanks to their stable, scalable score-matching objectives that extend to datasets containing hundreds of millions of clips. These models gradually refine a noisy sequence—first establishing global structure, then restoring fine-grained details—thus naturally promoting temporal consistency. Moreover, classifier-free guidance enables a single backbone to accommodate diverse conditioning signals such as text, images, depth maps, or pose sequences without architectural changes. Together, these advances yield superior visual sharpness, smoother motion, and more robust identity preservation compared to previous paradigms.

\paragraph{3D Rendering of Photorealistic Avatars.}
Photorealistic avatars have evolved from pure academic research to maturated technology that found its way into products, such as Apple's Personas or Epic Games' Metahumans.
In this work, we build upon Codec Avatars, which have been seminal as a photorealistic representation of humans.
Originally, Codec Avatars were introduced as face-only avatars using deep appearance models~\citep{lombardi2018deep} that represent expression changes on textured face meshes through a view-conditioned variational autoencoder.
In~\cite{bagautdinov2021driving}, a similar concept was applied to full-body avatars.
Breakthroughs in neural rendering such as neural volumes~\citep{Lombardi2019neural} and mixture of volumetric primitives~\citep{lombardi2021mixture} led to significant improvements in the visual quality of Codec Avatars, overcoming major limitations of textured meshes in terms of modeling hair.
The latest generation of Codec Avatars is based on Gaussian splatting~\citep{kerbl3Dgaussians} for both face~\citep{saito2024relightable} and full-body avatars~\citep{WangARXIV2025}.
While 3D representations of photorealistic humans are starting to escape the uncanny valley, driving these representations from sensory inputs remains challenging.
Existing approaches focus on camera-driven avatars~\citep{wei2019vr,bai2024universal}; however, such visual observations are not available when attempting to drive an avatar with speech alone or from the output of an LLM.
For audio-driven face avatars, early works focus on un-textured geometry~\citep{richard2021meshtalk,VOCA2019} or person-specific texture-based models~\citep{richard2021audiogaze}.
More recently, audio-driven face-only Gaussian avatars started showing their promise~\citep{aneja2024gaussianspeech}.
Beyond speech-driven avatars, ~\cite{chatziagapi2025av} demonstrate joint generation of speech and photorealistic 3D face motion directly from LLM-generated text.
Most closely related to our work,~\cite{conversation-motion} drive full-body avatars in dyadic conversations using audio from both speakers as input, however, their approach is limited to a few person-specific avatars and is trained on less than two hours of data per person, putting significant limitations on diversity and comprehensiveness of the synthesized body gestures.

%% file: conclusion.tex
\section{Conclusion}
\label{sec:conclusion}


This paper introduced the \mosaic dataset, a large-scale collection of over 4,000 hours of face-to-face interaction footage from over 4,000 participants in diverse contexts. We also presented a family of research models, Dyadic Motion Models, which can not only generate motion gestures and facial expressions that align with human speech, but also take into consideration the visual behaviors of the interlocutor. We presented a variant with speech from LLM model and integrations with 2D and 3D rendering methods, bringing us closer to interactive virtual agents. Finally, we described controllable variants of our motion models that can adapt emotional responses and expressivity levels, as well as generating gestures that are more semantically relevant. These dyadic motion models are demonstrating the potential for more intuitive and responsive human-AI interactions.






%% file: ack.tex

\section*{Acknowledgements}

We want to extend our gratitude to those who made this work possible below. 
To Arianne Burrell, Ben Samples, Josh Terry, Kenny Lehmann, Julia Vargas, Alyssa Newcomb, Michael Robert Brown, Shun Shiga, IV Tench, Karla Martucci, Michelle Restrepo, Nathan Hass, Junho Kim, Paula Chowles, Kate Bourdeau, Allie Castro, and Britt Montalvo for their tireless efforts in promoting our work.
To Nisha Deo and Ashley Gabriel for their expertise in internal and external communications.
To Idan Afek for strategic guidance and support in forging new partnerships.
To Ernest Hammond and Corey Wallace for providing invaluable counsel and ensuring that our work is compliant with all relevant regulations.
To Rachel Kim and Ty Toledano for their dedication to protecting user data and ensuring that our work meets high standards of privacy and security.
To Maeve Ryan for navigating complex policy landscapes and advocating for our users' interests.
To Yael Yungster and Kei Koyama for bringing our vision to life through their creative and innovative designs.
To Lindsey Miller for insights into user behavior and preferences, which have informed our design decisions and improved the overall user experience.
To Peng-Jen Chen, Min-Jae Hwang, Bokai Yu, Oleg Repin, and Alex Mourachko for engaging in stimulating discussions and sharing their expertise.
To Rob Fergus, Joelle Pineau, and Stephane Kasriel for their leadership, guidance, and unwavering support throughout this project.

%% file: appendix/arxiv.tex
\appendix
\section{\mosaic}
\label{section:mosaic_overview}
\subsection{Properties}
\subsubsection{Corpus Text Analysis}
\label{section:corpus_text_analysis}
Given that the recorded interactions that comprise \mosaic were organized at various locations and did not occur spontaneously, one must consider as a potential risk that the language content may not be representative of naturally occurring conversations. To assess this risk, we analyzed the language of \mosaic along grammatical, lexical, and structural lines.

\subsubsection{Readability, Lexical Diversity, Known Features of Conversation}

\paragraph{Readability.} Although readability scores apply primarily by definition to written material, some tests (such as Flesch's reading ease) also indicate expected score ranges for conversational English \citep{flesch1979write}.

To determine whether the Flesch reading-ease score (FRES) of \mosaic language in general is within the expected range, and determine whether the use of actors affects FRES, 3 corpora are first gathered from \mosaic transcripts, representing language encountered in interactions between non-actor dyads, dual-actor dyads, and single-actor dyads. Next, 2 other corpora are gathered for comparison, representing miscellaneous literature language and US presidential inaugural addresses, respectively. 

The literature corpus is expected to return an average FRES of 60 or below, while spontaneous, naturally occurring speech transcripts are expected to score at least above 82 and even closer to 92. In the absence of prior reference for inaugural addresses, we can only observe that the prevalence of unusually long sentences will cause a fairly low score to be returned.

To compute mean and median FRES, 30 passages of approximately 2,200 words in length are sampled. We use the \texttt{textstat}\footnote{https://pypi.org/project/textstat/} python package to compute FRES on each sample, then compute the mean and median scores for the sample sets. 

We find that \mosaic corpora fall within the expected range for conversational English (82–92), as shown in \Cref{tab:fres}. 

\begin{table}[h]
    \centering
    \begin{tabular}{cccc}
        \toprule
\textbf{Domain} & \textbf{Mean FRES} & \textbf{Median FRES} & \textbf{Expected FRES}\\
\midrule
Inaugural addresses & 43.5 & 42.7 & < 60\\
Literature & 62.9 & 65.2 & $\sim$60\\
\midrule
\mosaic Non-actor & 88.0 & 86.4 & 82–92\\
\mosaic Dual-actor & 91.6 & 91.3 & 82–92\\
\mosaic Single-actor & 87.1 & 86.1 & 82–92\\
\bottomrule
    \end{tabular}
    \caption{Fresch reading ease scores (FRES) for \mosaic corpora.}
    \label{tab:fres}
\end{table}

\paragraph{Lexical diversity.} To complement the previous readability analysis, lexical diversity computation methods such as the Measure of Textual Lexical Diversity (MTLD) can be used \citep{mccarthy2005assessment}. The MTLD method focuses more specifically on lexical variability and is less sensitive to small variations in sample length \citep{mccarthy2010mtld}. When different discourse types are elicited, the measuring of lexical variability returns differences in score ranges, as shown in \cite{fergadiotis2011productive}. Additionally, \cite{biberlongman} shows that in large representative corpora, such as LSWE, conversational English receives by far the lowest lexical variability score due to the high degree of repetition in function words and inserts (e.g., \textit{yeah}, \textit{right}, \textit{um}).

MTLD scores are computed on the 3 \mosaic corpora, as well as the literature and inaugural address (i.e., scripted speech) corpora for control, using the \texttt{lexicalrichness}\footnote{https://pypi.org/project/lexicalrichness/} python package with a type-token ratio (TTR) factor of 0.72.

We find that the \mosaic copora all fall within the MTLD score range we would expect (at least 1.75 times lower than composed literary language). Scores for all 5 corpora are reported in \Cref{tab:mtld}.

\begin{table}[h]
    \centering
    \begin{tabular}{cc}
        \toprule
DOMAIN & MTLD score\\
\midrule
Literature & 85.9\\
Inaugural addresses & 72.3\\
\midrule
\mosaic Non-actor & 48.1\\
\mosaic Dual-actor & 39.8\\
\mosaic Single-actor & 42.1\\
\bottomrule
    \end{tabular}
    \caption{MTLD scores for \mosaic corpora and controls.}
    \label{tab:mtld}
\end{table}
\subsubsection{Collection Facility Geography}
\label{appendix:collection_geo}

\begin{itemize}
    \item Chino Hills, CA
    \item Costa Mesa, CA
    \item Irvine, CA
    \item Los Angeles, CA
    \item Boise, ID
    \item Pittsburgh, PA
    \item Las Vegas, NV
    \item Boston, MA
    \item Waltham, MA
    \item New York, NY
\end{itemize}

\subsubsection{Known Limitations of Text Transcripts}

We acknowledge several limitations of our current text transcript methodology.

WhisperX, like most speech recognition systems, is tuned to provide concise, readable transcripts that ignore hesitations, false-starts, and other disfluencies which have important visual correlates and are interesting phenomena in their own right. We experimented with other systems that aim to robustly transcribe such behaviors ~\citep{wagner2024crisperwhisperaccuratetimestampsverbatim} but found the overall quality to be lower.

We also find that the wav2vec alignment step of WhisperX can produce inaccurate timestamps on short utterances. Approximately 98\% of sessions and 87\% of individual interactions contain at least one word whose timestamp-derived length is greater than 3 standard deviations from the mean.    

While the current time-aligned transcripts enable preliminary corpus analyses on text (see previous sections in this \Cref{section:corpus_text_analysis}) and turn-taking behavior in the spirit of ~\cite{HELDNER2010555}, future work will focus on producing more robust transcripts and accurate timestamps for \mosaic that enable new contributions in these fields.  

\subsection{Quality and Safety}
\begin{figure}
\includegraphics[width=1\textwidth]{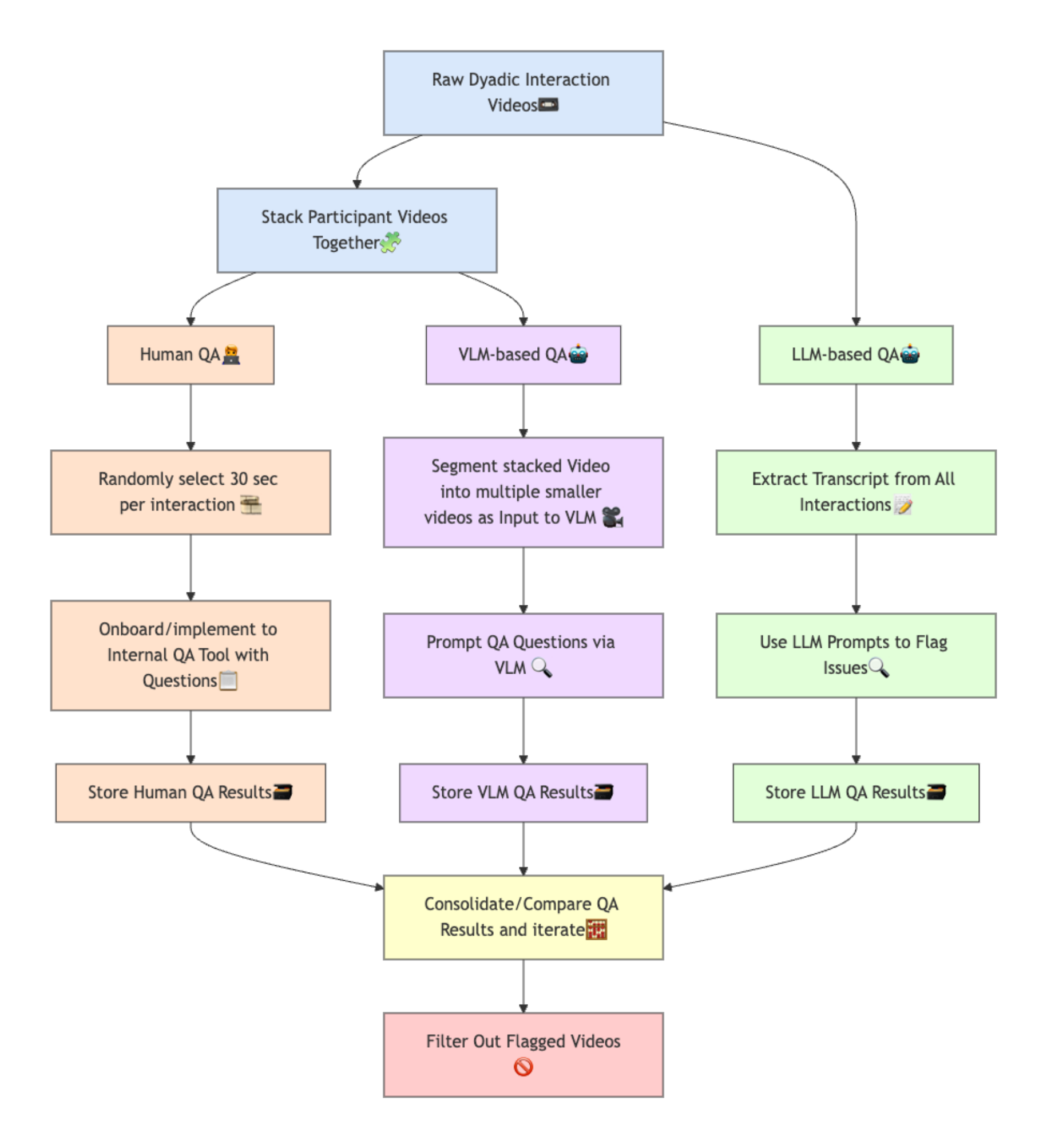}
\caption{Illustration of the scalable Safety and QA process applied to \mosaic. Using this process every minute of footage received a QA signal.}
\label{fig:qa_flow}
\end{figure}

\subsubsection{Participant Pre-amble}
\label{section:participant_preamble}
Moderators were instructed to provide this information at the start of every session recording:

Welcome and thank you for participating in this project. The goal of this project is to study how humans use visual cues in communication. We will just ask you to have a series of short conversations based on prompts we have provided. The most important thing is for you to behave naturally. Try not to let the cameras or the setting intimidate you — just read the prompt and engage with it, try to have fun and try to express your true feelings. 

Please stand where indicated. We would like you to stand for as long as you feel comfortable, but if you get tired you may use the stools. Please just do not touch or interact with the stools using your hands — we'd like your hands to remain free during the conversation. [Indicating to the scripts, face down] These are the scripts which contain the prompts you will discuss. After we are set-up and recording, I will ask you to start with prompt 1. Each time we move to a new prompt, you will pick up the scripts and read the prompt, silently to yourself. Once you have read it and understand it, place the scripts back down. Again, we just want you to not have anything in your hands during the interaction. Then [Initiator] will begin the prompt. [Initiator] will be the person that starts every conversation. 

We expect each conversation to last from a few moments to 5 minutes. I may interrupt the conversation and ask you to move on to the next prompt - please don't read into this, I am just doing it in the interest of time and getting through all the prompts.

If you feel like you don't have anything more to say on a particular prompt, just let me know and we can move on to another one.

While you should feel comfortable to express yourself fully and how you see fit, we do ask that you refrain from discussing any personal details such as phone numbers, addresses, emails, et cetera, from you or anyone else. This is just to protect everyone's privacy.

\subsubsection{Sensitive Data Statement Given to Vendors and Moderators}
\label{section:sensitive_data_statement}
The following text was provided to all collection facilities and communicated to participants via moderators:

During the data collection process, it is important for participants to be mindful of their privacy and avoid disclosing personal or sensitive information. This includes:
\begin{itemize}
    \item Cultural Background: Ethnicity 
    \item Financial Identifiers, Financial Status (e.g., loans) 
    \item Membership of a Trade Union
    \item Political affiliation, Political issues, Political opinion
    \item Health Data, Medical Condition
    \item Biometrics: Body specifications, Fingerprints, Iris scan, skin color
    \item Faith-based affiliation, Faith-based holiday, Faith-based spiritual \& moral belief 
    \item Sexual Orientation
    \item Personal \& government identification number or image (e.g., picture of passport, passport number, driver’s license number, SSN, Tax ID)
    \item Criminal Record
    \item Citizenship
    \item Precise Location/ Location Services
    \item Transgender, Intersex, Non-binary Gender
\end{itemize}

Avoiding sharing any information that could potentially identify them, such as their name, address, or other identifying details. By being mindful of what they share, participants can help protect their privacy and ensure that their personal information remains confidential. It is also important for the moderator to inform participants about avoiding these topics if it's accidentally disclosed during an interaction.  It is also important for the vendor to inform participants about the potential risks and benefits of participating in the study, as well as their rights and responsibilities, as a part of obtaining informed consent.

\subsubsection{Example Participant Recording Workflow}
\label{section:participant-workflow}
\begin{enumerate}
    \item Participants are oriented to the project and presented a consent form. Informed consent (and the signing of the consent form) are mandatory for inclusion in the project.
    \item Moderator greets participants according to the Participant Introduction (\Cref{section:participant_preamble}).
    \item Moderator confirms the participants aren't wearing any sensitive clothing. 
    \item Moderator guides the participants to their places. Moderator encourages the participants to stand, but offers the option to sit if they are tired. Participants are reminded to avoid touching the chair/stool/standing desk with their hands. 
    \item Moderator places the scripts on the script stands and walks participants through the interaction flow, which is:
    \begin{enumerate}
        \item Moderator will ask the participants to begin with a prompt.
        \item Participants will pick up the script off the script stand and read the indicated prompt silently to themselves.
        \item When they have understood the prompt, they put the scripts back down on the script stands.
        \item One of the participants (as indicated by the prompt) begins the interaction.
    \end{enumerate}
    \item Moderator starts recording, ensuring video and audio feed for both talents are recording and time aligned
    \item Moderator begins the session by asking participants to start with the first prompt.
    \item Moderator notes the time-stamp of the interaction start. Moderator then observes the interaction and takes notes where applicable.
    \item Moderator interrupts the interaction and records the timestamp of the end of the interaction.
    \item Moderator records the rating of the interaction.
    \item Steps 6 through 9 are repeated for the remaining prompts in the script, until the end of the session. 
\end{enumerate}

\subsection{High-Level QA Categories}
\label{section:qa-flags}
Below we provide the 7 high-level categories that were used in both Human- and model-based QA checks:
\begin{itemize}
    \item Sensitive material (PII) with separate sub-categories.
    \item Offensive material.
    \item Participant visibility.
    \item Audio comprehension.
    \item Presence of recording artifacts.
    \item Audio-video sync check.
    \item Participant engagement check.
\end{itemize}

\subsubsection{VLM Raw Footage QA Prompts}
\textbf{Context criteria provided to the system:}

\textit{You are a human reviewer who is assessing the quality of a video recording based on different criteria. A stacked video of two participants on the left and right sides is provided. Your task is to answer the question based on the video.
The criteria are as follows:}

\begin{enumerate}
    \item Participant Visibility:
    \begin{enumerate}
        \item Body Visibility: Both participants must be visible from at least waist-up in their respective video frames, with all required body parts (torso, head, shoulders, and hands) fully visible throughout the entire video.
        \item Complete Body Part Visibility: Each participant's face, shoulders, and hands must be completely visible within their frame throughout the entire video. If any of these body parts are not fully visible at any point, answer 'No'.
        \item Frame Presence: Both participants must be present in their respective frames throughout the entire video. If a participant is not present in their frame or if any part of their body goes out of their frame, answer 'No'.
        \item Detailed Inspection: Carefully examine both participants' hands, shoulders, and heads to ensure that they are visible throughout the video. You can inspect the video frame by frame to verify that the participants and their required body parts are visible in all frames.
    \end{enumerate}
    Answer Criteria: Answer 'Yes' only if both participants meet the above criteria for the entire duration of the video. If you are uncertain about the participants' visibility, answer 'Unsure'.
    \item Audio Comprehension:
    \begin{enumerate}
        \item Audio Presence: Determine whether there is any audible speech from either participant throughout the video.
        \item Audio Clarity: If there is audio, assess whether it is clear and intelligible for both participants. Consider factors such as:
            \begin{enumerate}
                \item Volume: Is the audio at a reasonable level, or is it too loud or too soft?
                \item Background noise: Are there any distracting sounds that interfere with the participants' speech?
                \item Audio quality: Is the audio free from distortion, hiss, or other defects?             
            \end{enumerate}
        \item Participant speech: Identify which participant(s) are speaking in the video and determine if their speech is understandable.
    \end{enumerate}
    Answer Criteria:
    \begin{enumerate}
        \item If there is no audio or the audio is completely unintelligible, answer 'No'.
        \item If one or both participants' speech is partially or fully intelligible, answer 'Yes'.
        \item If you are uncertain about the audio comprehension, answer 'Unsure'.
    \end{enumerate}
    \item Recording Artifact Presence:
    \begin{enumerate}
        \item Audio Artifacts: Check for any audio-related issues, such as:
        \begin{enumerate}
            \item Buzzing or humming sounds
            \item Echo or reverberation
            \item Beeping or other high-pitched noises
            \item Distortion or clipping
        \end{enumerate}
        \item Video Artifacts: Check for any video-related issues, such as:
        \begin{enumerate}
            \item Frozen or stuck images
            \item Blurry or pixelated images
            \item Overexposure or underexposure
            \item Other visual distortions
        \end{enumerate}
    \end{enumerate}
    Answer Criteria:
    \begin{enumerate}
        \item If there are no noticeable recording artifacts, answer 'No'.
        \item If one or more recording artifacts are present, answer 'Yes'.
        \item If you are uncertain about the presence of recording artifacts, answer 'Unsure'.
    \end{enumerate}
    \item Audio-Video Sync Check:
    \begin{enumerate}
        \item Lip Sync: Check if the lip movements of both participants align with the audio. Are their lips moving in sync with the words being spoken?
        \item Body Language Sync: Check if the body language of both participants aligns with the audio. Do their gestures, facial expressions, and posture match the tone and content of the conversation?
        \item Audio Video Matching: Check if the audio and video data seem to match naturally. Does the audio appear to be coming from the correct participant at the correct time? Are there any noticeable delays or desynchronization between the audio and video?
    \end{enumerate}
    Answer Criteria:
    \begin{enumerate}
        \item If the audio and video data are well-synchronized and create a natural conversation flow, answer 'Yes'.
        \item If there are noticeable issues with lip sync, body language sync, or audio-video matching, answer 'No'.
        \item If you are uncertain about the synchronization of audio and video data, answer 'Unsure'.
    \end{enumerate}
    \item Participant's Video Sync Check:
    \begin{enumerate}
        \item Video Synchronization: Check if the participant videos are synchronized without any artificial lag or distortion, and they're talking about a related topic throughout the video.
        \item Lag Detection: Look for any noticeable delays or lags between the two participant videos. Are they moving in sync with each other? Specifically, assess if their speech, body language, and facial expressions align with conversational flow.
        \item Speech and Reaction Alignment: Ensure that the conversation timeline is synchronized.
        \item Distortion Detection: Check for any distortions or irregularities in the video feed that could indicate a synchronization issue.
    \end{enumerate}
    Answer Criteria:
    \begin{enumerate}
        \item If the participant videos are well-synchronized and there are no noticeable lags or distortions, answer 'Yes'.
        \item If you observe any synchronization issues, such as lag or distortion, answer 'No'.
        \item If you are uncertain about the synchronization of the participant videos, answer 'Unsure'.
    \end{enumerate}
    \item Offensive Material Check:
    \begin{enumerate}
        \item Profanity: Check if there is any use of profane language or swear words in the video.
        \item Explicit Gestures or References: Look for any explicit gestures, such as middle fingers or other obscene hand signals, or references to explicit content.
        \item Iconography: Check if there are any icons, logos, or graphics on clothing, hats, or other items that may be considered offensive or problematic.
        \item Other Problematic Material: Consider if there is any other material in the video that may be considered problematic or off-putting, such as hate speech, discriminatory language, or violent imagery.
    \end{enumerate}
    Answer Criteria:
    \begin{enumerate}
        \item If you did not observe any offensive material in the video, answer 'No'.
        \item If you observed any offensive material in the video, answer 'Yes'.
        \item If you are uncertain about the presence of offensive material in the video, answer 'Unsure'.
    \end{enumerate}
    \item Sensitive Material Mention:
    \begin{enumerate}
        \item Faith-based affiliation, Faith-based holiday, Faith-based spiritual \& moral belief: Check if there is any mention of faith-based affiliations, holidays, or spiritual and moral beliefs that may be considered sensitive.
        \item Sexual Orientation: Look for any mention of sexual orientation that may be considered sensitive.
        \item Personal \& government identification number or image: Consider if there is any mention of personal or government identification numbers or images, such as passport numbers or driver's license numbers, that may be considered sensitive.
        \item Criminal Record: Check if there is any mention of criminal records that may be considered sensitive.
        \item Citizenship: Look for any mention of citizenship that may be considered sensitive.
        \item Precise Location/ Location Services: Consider if there is any mention of precise location or location services that may be considered sensitive.
        \item Transgender, Intersex, Non-binary Gender: Check if there is any mention of transgender, intersex, or non-binary gender identities that may be considered sensitive.
    \end{enumerate}
    Answer Criteria:
    \begin{enumerate}
        \item If you did not observe any sensitive material in the video, answer 'No'.
        \item If you observed any sensitive material in the video, answer 'Yes'.
        \item If you are uncertain about the presence of sensitive material in the video, answer 'Unsure'.
    \end{enumerate}
    \item Participants Engagement Check:
        \begin{enumerate}
            \item No Distractions: Look for any signs of distraction, such as:
                \begin{enumerate}
                    \item Talking to or looking at the moderator. This can be identified if participant's face is not towards the camera for long periods of time.
                    \item Walking out of frame
                    \item Using phone
                    \item Drinking water or eating
                    \item Engaging in any other activity that takes their attention away from the conversation
                \end{enumerate}
            \item Active Participation: Consider if both participants are actively participating in the conversation, which means either responding to each other's questions and statements, or showing interest in the discussion like nodding, smiling, acknowledging, etc. There could be instances where only one participant is talking and another is listening, but they are both engaged in the conversation.
        \end{enumerate}
    Answer Criteria:
    \begin{enumerate}
        \item If both participants are hundred percent engaged in the interaction and show no to little signs of distraction, answer 'Yes'.
        \item If one or both participants show signs of distraction or lack of engagement for a long period, answer 'No'.
        \item If you are uncertain about the participants' engagement, answer 'Unsure'.
    \end{enumerate}
\end{enumerate}
\textbf{VLM QA Prompt:} 
\textit{For the following video, use the context provided and answer for all the QA categories in the format of a key value pair like:}

[participant\_visibility: Yes, audio\_comprehension: No, recording\_artifact\_presence: Yes, audio\_video\_sync\_check: Yes, participant\_video\_sync\_check: No, offensive\_material\_check: No, sensitive\_material\_mention: No, participant\_engagement\_check: Yes]

\textit{Do not add any other info in your answer, just stick to the format provided.}

\subsubsection{Text-LLM Transcript QA Prompts}
\textbf{Context criteria provided to the system:}
\textit{You are a reviewer to check the quality of the audio through transcripts. In the transcript, 2 participants will interact with each other and discuss a topic given a prompt. we would like to capture any quality related issues related to content or audio itself. Answer each question with this format without explanation. Your answer only needs to contain the correct option ID such as a or b or c or d etc}

\textbf{Prompt to flag sensitive content:} Answer the question: {Did you observe any explicit description of sensitive material about speakers themselves (not other people in their story)? Choose one answer 'a. No Sensitive info' when you are not certain about other answers
\begin{enumerate}[label=(\alph*)]    
    \item No Sensitive info 
    \item Explicit Ethnicity about the speaker and their family (such as black, Ukrainian etc)
    \item Explicit mentioned speaker's Financial Identifiers (eg. credit card number, Social Security number, not general description such as ``rich'' or ``poor'')
    \item Membership of a Trade Union 
    \item Explicitly mentioned speaker's Political affiliation (such as republican, democrat, etc)
    \item Explicitly mentioned Speaker's severe or embarrising disease (terminal illnesses, or chronic diseases that significantly impact daily life, not mental health disorders nor general mention such as ``doctor'' or ``hospital'' or ``medication'')
    \item Explicit Biometrics number/color about the speaker themselves (not general description such as ``tall'' or ``skinny'')
    \item Explicit mentioned spleakers Faith-based affiliation (such as I am a muslim, catholic, etc)
    \item Personal \& government identification number or image (e.g. picture of passport, passport number, driver’s license number, SSN, Tax ID) 
    \item Severe Felony Criminal charges (eg. violent offenses, sex crimes, or serious property crimes)
    \item Explicit Citizenship 
    \item Precise address of the speaker such as street and house number (such as 123 Main St) 
    \item Transgender/Intersex/Non-binary Gender
\end{enumerate}

\textbf{Prompt to flag offensive content:} 
Answer the question: Did you observe any strong and explicit offensive material? Choose answer 'b. No' when you are not certain
\begin{enumerate}[label=(\alph*)]  
    \item Yes
    \item No
\end{enumerate}

\subsection{Workflow for Internal State, Rationale, and Visual Behavior Annotations}
\label{section:annotation-workflow}
The annotation workflow follows the below steps:
\begin{enumerate}
    \item In-person collection. Participants come in for a session to be recorded.
    \item Trained annotators identify all conspicuous visual elements, and timestamp the recordings accordingly.
    \item Approximately 24 hours later, and no more than 36 hours:
    \begin{enumerate}
        \item Participants come back to look at all the timestamped sections and try to annotate their internal states at those moments (1P-IS annotations). If they have nothing to annotate, they can move on to the next timestamp. 
        \item At the same time, participants also provide behavior rationales (1P-R) annotations for the same timestamped sections.
    \end{enumerate}
    \item Once the 1P annotation round is finished, trained annotators annotate:
Perceived internal state (3P-IS)
Perceived behavior rationale or perceived theory of mind (3P-R)
on at least all the moments annotated as 1P by the participants (and possibly more identified timestamps, if they think they can do it).
    \item Finally, trained annotators annotate visual descriptions (3P-V) for all timestamped moments, whether or not 1P annotations have been provided.
    \item External quality assurance: A portion of the deliverable (minimum 20\% of all annotations) is reviewed for quality and compliance with the guidelines. 
    \item Internal quality assurance, feedback, and continuous improvement: Annotation files are additionally inspected for compliance with the guidelines. Additional feedback is provided to annotation vendors. 
\end{enumerate}

\section{Human Evaluation Protocols}
\label{section:protocols}
\subsection{Dyadic Body Protocol (DBP)}
\subsubsection{\textbf{Introduction}}
In this study, you'll watch short video clips of real people and digital avatars talking. After each clip, you will be asked a few questions about how the avatar moved while speaking and while listening. We are interested in your personal impressions. \textbf{There are no right or wrong answers.}

Your feedback will help our team improve how AI platforms generate realistic and expressive virtual characters. This technology may be used in video games, virtual assistants, and other interactive tools where natural movement matters.

Please note:
\begin{itemize}
\item The avatar's face will remain static and is not a focus of the current study.
    \item The lower body will remain mostly stationary and is not a focus of the current study.
    \item We ask that you focus on the visual behavior from the waist up, which may include movement in the head, neck, shoulders, arms, wrists, hands, and torso.
    \item Some of the avatar’s movements may contain some shakiness and jitteriness. Please do your best to disregard these and focus on the quality and appropriateness of the gestures.
\end{itemize}

The people in the videos will be talking and include actual human footage (always presented on the left) and a 3D animated representation (always presented on the right).
\begin{figure}
    \centering
    \includegraphics[width=0.75\linewidth]{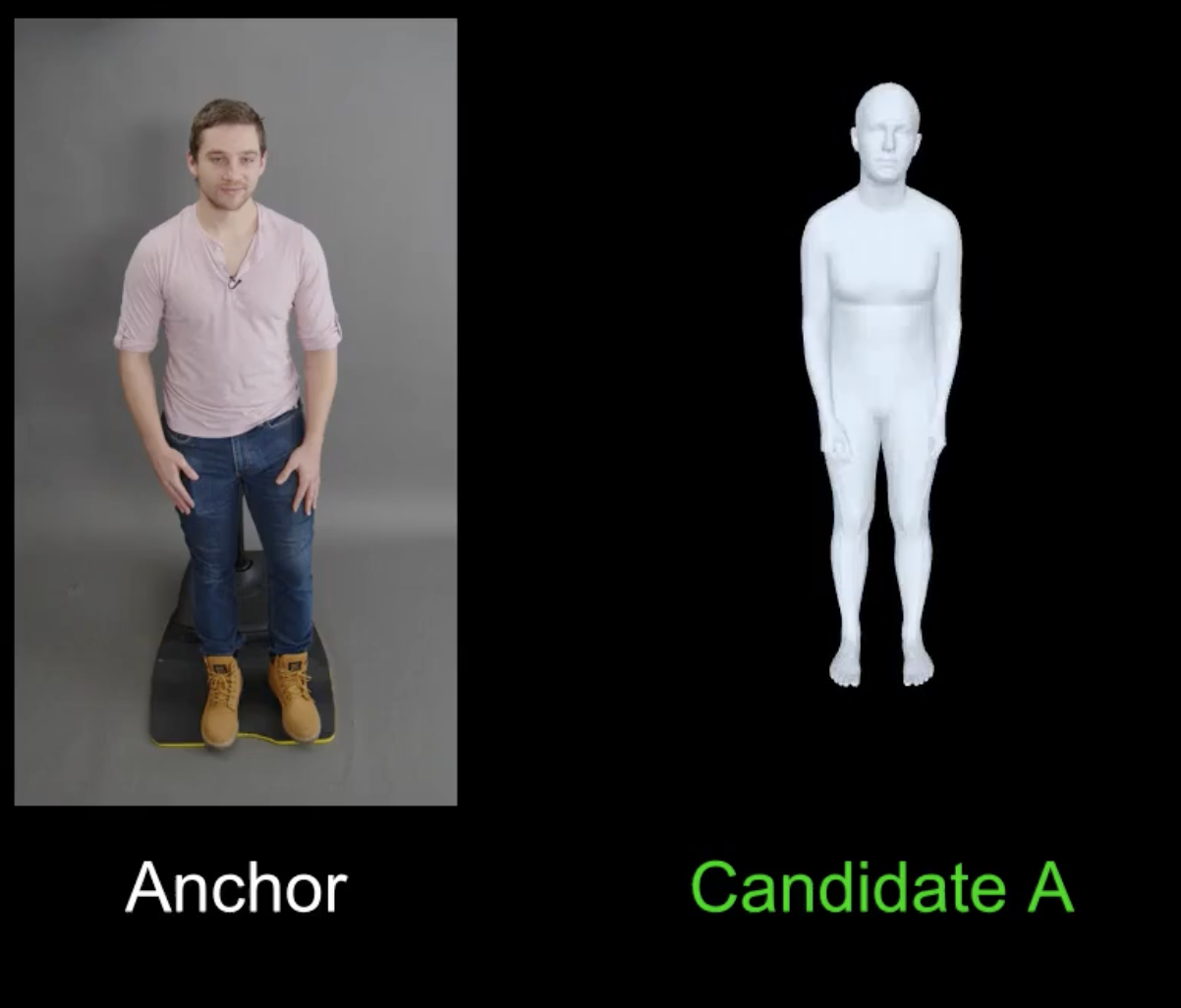}
    \caption{DPB stimuli.}
    \label{fig:enter-label}
\end{figure}

\subsubsection {\textbf{The Task}}
On a given trial, you will be presented with two video clips:
\begin{figure}
    \centering
    \includegraphics[width=1\linewidth]{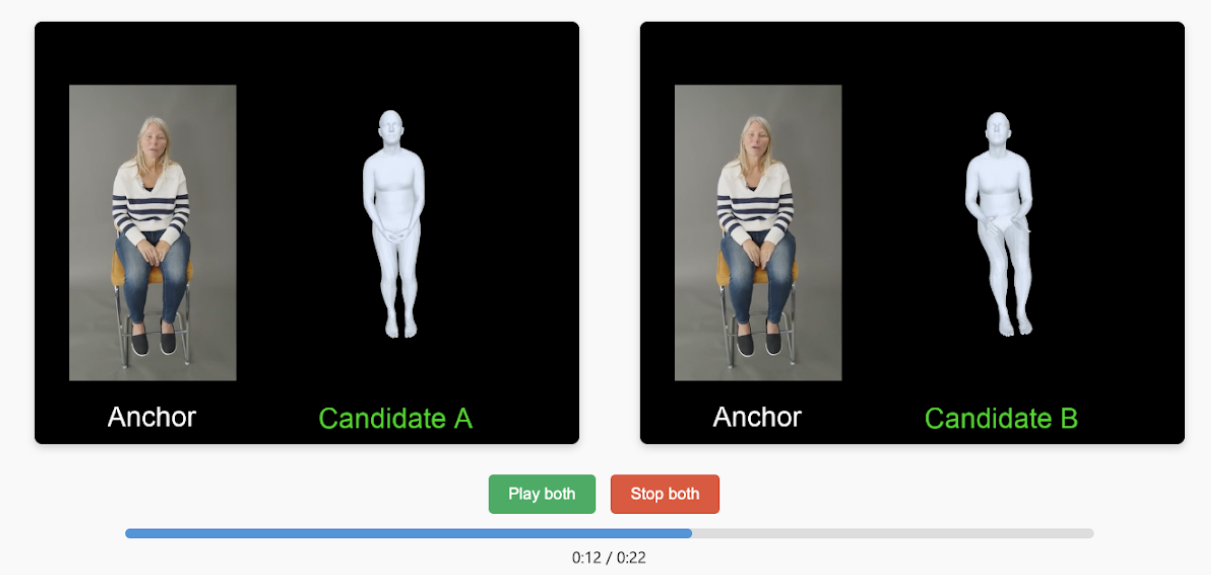}
    \caption{DPB pairwise comparison.}
    \label{fig:enter-label}
\end{figure}

The \textbf{Anchor} \textbf{is the actual human footage, which is the same in both videos.} You are not asked to provide ratings about Anchor directly, but we include Anchor, so that you have a visual reference and the context of the conversation. 

The left video includes a 3D animated rendering of \textbf{Candidate A} who is a dialogue partner for the Anchor. The right video includes a 3D animated rendering of \textbf{Candidate B }who is a dialogue partner for Anchor. 

We provide a visual indicator to help you quickly identify who is speaking. When the \textbf{Anchor} is speaking, the video label ``Anchor'' will be \textbf{highlighted green}. Similarly, When \textbf{Candidate A} and  \textbf{Candidate B }are speaking their role labels will be highlighted green. 

\textbf{Your task} is to assess which video,\textbf{ Candidate A or Candidate B }is a better dialogue partner for the Anchor along several dimensions. 

You can play each video individually and you can also watch the video clips simultaneously. 

\textbf{Your task is not} to assess the quality of \textbf{Anchor}; the \textbf{Anchor }video is provided only for context. All your judgments should be about whether \textbf{Candidate A} or \textbf{Candidate B} are higher quality dialogue partners for \textbf{Anchor.} 

\textbf{Definitions} 
Anchor - Is a human participant. You are not to rate this video, it is there to provide you with a visual reference and context for the conversation. 
Candidate A - Candidate dialogue partner for Anchor. You will be asked to rate this video compared to Candidate B. 
Candidate B - Candidate dialogue partner for Anchor. You will be asked to rate this video compared to Candidate A. 

\textbf{Roles} 
Speaking - The state in which a person is generating words and content in the conversation. 
Listening - The state in which a person is taking in the words and content in the conversations. 
Cross-talk - A state in which both people are speaking. 

\textbf{Actions} 
Turn-taking - the process of moving from speaking to listening and also from listening to speaking.

\subsubsection {\textbf{Flagging}}
As part of this annotation task, you are given the chance to flag and skip certain clips if you encounter technical issues or safety concerns. Occasionally, you may see broken or incomplete clips or audio. Additionally, the generated avatars might make gestures that could be interpreted as inappropriate, even if that was not the intent of the AI platform. 

When flagging, consider the following guidelines. 

Flag as broken or incomplete audio/video: 
\begin{itemize}
    \item When audio cuts out or distorts during a key moment leaving parts of the conversation unclear.
    \item When video freezes or skips making it impossible to evaluate the gesture accurately.

\end{itemize}

Flag for safety concerns or inappropriateness: 
\begin{itemize}
    \item When an avatar may appear to make lewd/sexual gestures.
    \item When an avatar shows gestures mimicking violent actions.
    \item When an avatar makes a gesture that could be interpreted as a hate symbol or is associated with harmful ideologies.
\end{itemize}

To flag and skip an item, click the following icon  to provide more information about the clip. Clicking the icon will route you to the following screen: 

Please provide your reason for flagging (select all that apply): 
$\square$ Audio is distorted 
$\square$ Audio is out of sync 
$\square$ Audio is cut out 
$\square$ Video freezes and/or skips 
$\square$ Avatar displays gestures that could be interpreted as lewd/sexual 
$\square$ Avatar shows violent gestures or actions 
$\square$ Avatar uses hate symbols or gestures associated with harmful ideologies 
$\square$ Other (Any other issue that impacts audio/video or makes the clip unsafe, uncomfortable, or inappropriate for evaluation): Please provide justification for why you are flagging this clip.

\subsubsection {\textbf{Item 1 (Lifelike)}}
\textbf{1. Overall, which candidate’s (A or B) visual behaviors are more lifelike? }\textit{By “lifelike,” we mean that the Candidate behaves in a way that looks, acts, or seems very similar to something that's real and alive. Visual behaviors can also include non-verbal actions that we use to communicate and express ourselves through our body language.}

$\circ$ Candidate A is much more lifelike 
$\circ$ Candidate A is slightly more lifelike 
$\circ$ Tie 
$\circ$ Candidate B is slightly more lifelike 
$\circ$ Candidate B is much more lifelike

\subsubsection {\textbf{Item 2 (Clarity of Intent)}}
\textbf{2. Which candidate (A or B) most clearly demonstrates an intent with their visual behaviors? } \textit{By “intent,” we mean visual behaviors that are deliberate, non-repetitive, and convey a specific thought, idea, or emotion. }

$\circ$ Candidate A appears to be much more intentional 
$\circ$ Candidate A appears to be slightly more intentional  
$\circ$ Tie 
$\circ$ Candidate B appears to be slightly more intentional 
$\circ$ Candidate B appears to be much more intentional

\subsubsection {\textbf{Item 3 (Turn-Taking)}}
\textbf{3. Which candidate (A or B) appears to have better turn-taking behavior? }\textit{By “turn-taking behavior,” we mean gestures to indicate the intent to speak (such as by raising a hand, palm-up, just before speaking) or an intent to prompt a response from the dialogue partner (such as by raising both arms towards the listener or by nodding their head toward the listener). }

$\circ$ Candidate A appears to have much better turn-taking behavior 
$\circ$ Candidate A appears to have slightly better turn-taking behavior  
$\circ$ Tie 
$\circ$ Candidate B appears to have slightly better turn-taking behavior
$\circ$ Candidate B appears to have much better turn-taking behavior

\subsubsection {\textbf{Items 4 - 7 (Listening)}}
Now please rate the Candidates as they are listening:
\begin{figure}
    \centering
    \includegraphics[width=1\linewidth]{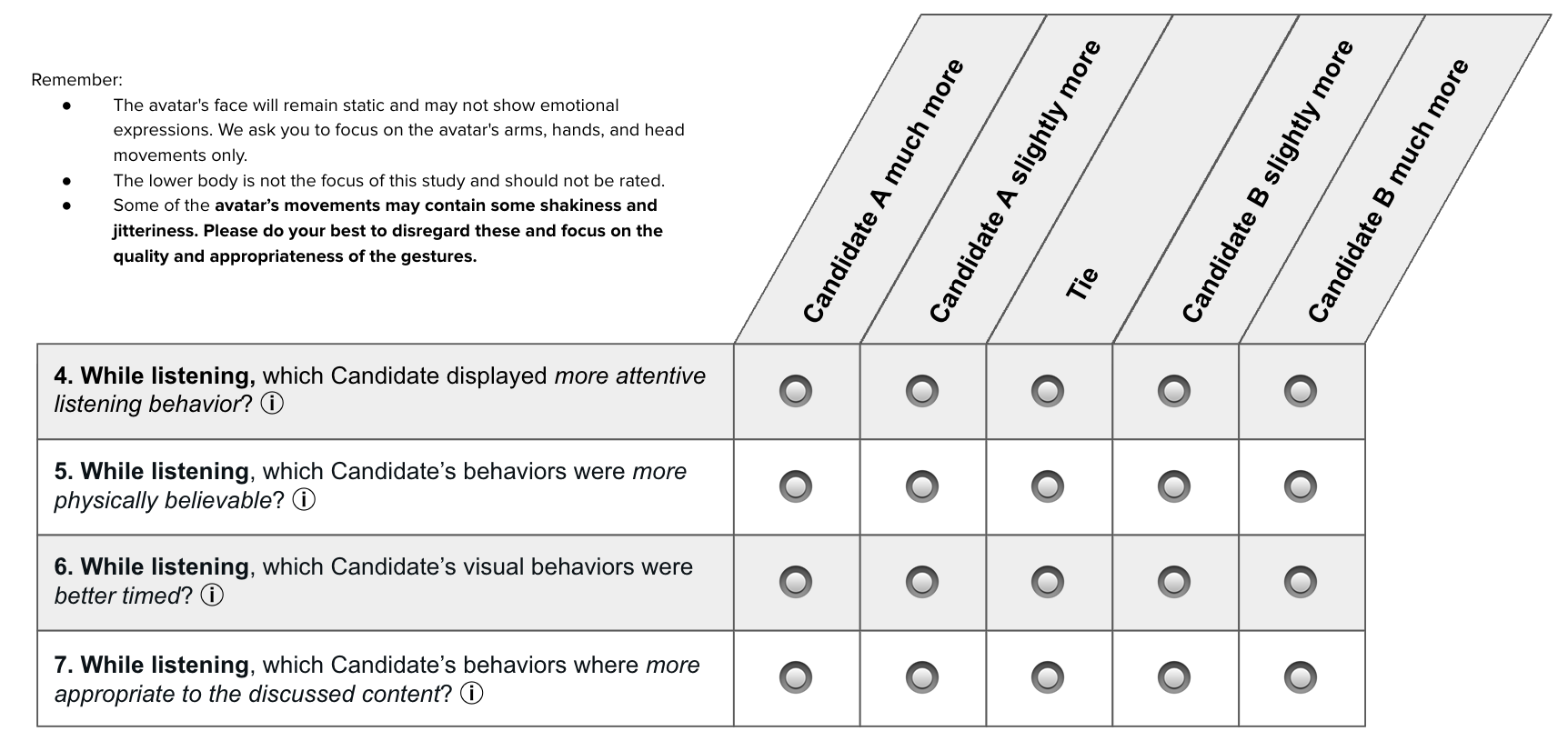}
    \caption{DBP listening items.}
    \label{fig:enter-label}
\end{figure}
4. \textbf{While listening, which Candidate displayed more attentive listening behavior? }
Tooltip content\textbf{ } : \textit{By “attentive” we mean behaviors like leaning forward, nodding or shaking their head in agreement, mirroring the hand or arm gestures of the Anchor, and visually indicating engagement and understanding.}
\textbf{5. }\textbf{While listening, which Candidate’s behaviors were more physically believable?}
Tooltip content\textbf{ } : \textit{By ``more physically believable'' we mean that the motion of the Candidate is humanly possible and does not exhibit impossible movements that defy human physiology.}
\textbf{6. While listening, which Candidate’s visual behaviors were better timed?}
Tooltip content\textbf{ } : \textit{By “better timed” we mean that the Candidate's listening movements and reactions are synchronized with the Anchor's speech.}
\textbf{7. While listening, which Candidate’s behaviors where more appropriate to the discussed content?}
Tooltip content\textbf{ } : \textit{By “more appropriate to the content discussed” we mean that the Candidate's behaviors, such as head nods and body language, are consistent with the emotional tone and subject matter of the conversation. For example, if the Anchor is discussing a serious topic, the Candidate's behavior should reflect a corresponding level of gravity or concern, rather than appearing overly casual or dismissive.}

\subsubsection {\textbf{Items 8 - 10  (Speaking)}}
Now please rate the Candidates as they are speaking:

\begin{figure}
    \centering
    \includegraphics[width=1\linewidth]{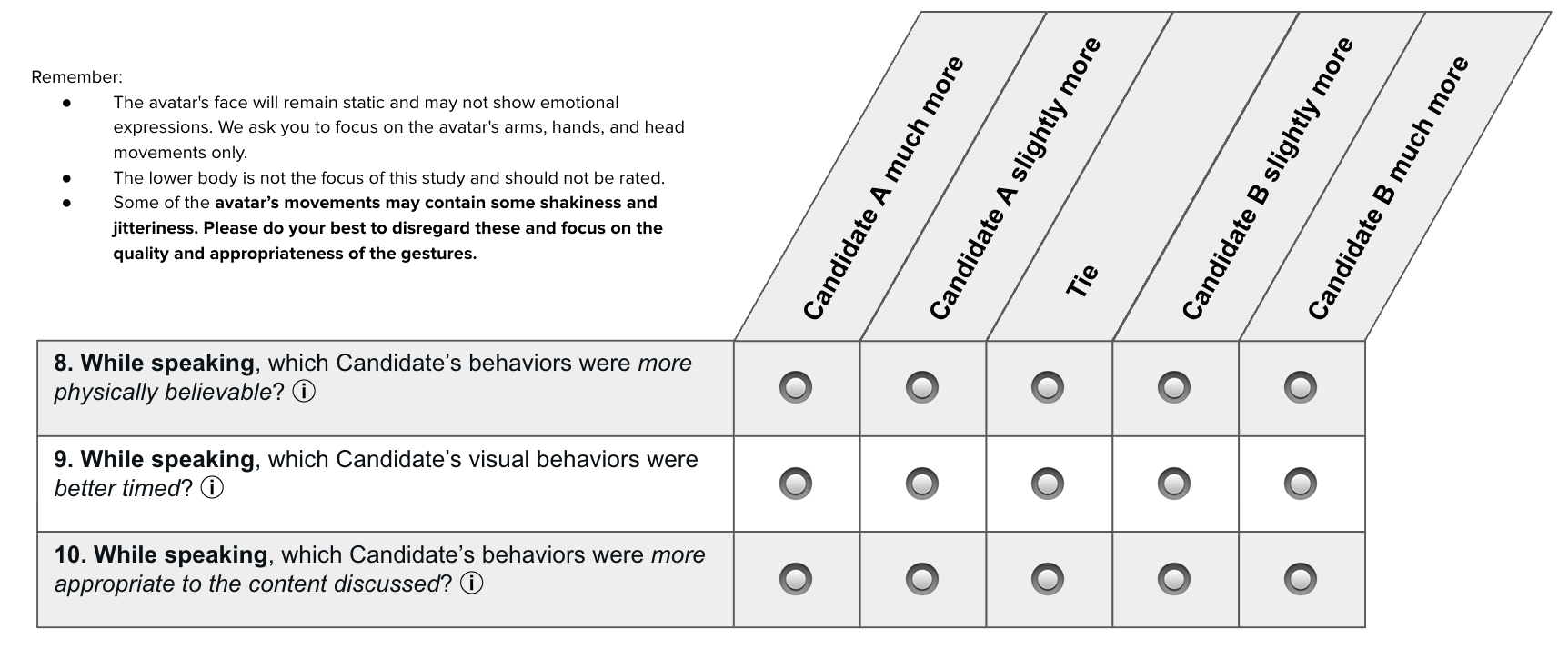}
    \caption{DBP speaking items.}
    \label{fig:enter-label}
\end{figure}
\textbf{8. While speaking, which Candidate’s behaviors were more physically believable?}
Tooltip content: \textit{By ``more physically believable'' we mean that the motion of the Candidate is humanly possible and does not exhibit impossible movements that defy human physiology.}
\textbf{9. While speaking, which Candidate’s visual behaviors were better timed?}
Tooltip content: \textit{By “better timed” we mean that the Candidate's speaking movements and gestures are synchronized with the Candidate's speech.}
\textbf{10. While speaking, which Candidate’s behaviors were more appropriate to the content discussed?}
Tooltip content: \textit{By “more appropriate to the content discussed” we mean that the Candidate's behaviors, such as head nods and body language, are consistent with the emotional tone and subject matter of the conversation. For example, if the Candidate is discussing a serious topic, the Candidate's behavior should reflect a corresponding level of gravity or concern, rather than appearing overly casual or dismissive.}

\subsection{Dyadic Face Protocol (DFP)}
\subsubsection{\textbf{Introduction}}
In this study, you'll watch short video clips of digital avatars and humans talking. After each clip, you will be asked a few questions about how one of the avatars moved while speaking and while listening. We are interested in your personal impressions. \textbf{There are no right or wrong answers. }

Your feedback will help our team improve how AI platforms generate realistic and expressive virtual characters. This technology may be used in video games, virtual assistants, and other interactive tools where natural movement matters.

\subsubsection {\textbf{The Task}}
On a given trial, \textbf{you will be presented with two video clips}:
\begin{figure}
    \centering
    \includegraphics[width=1\linewidth]{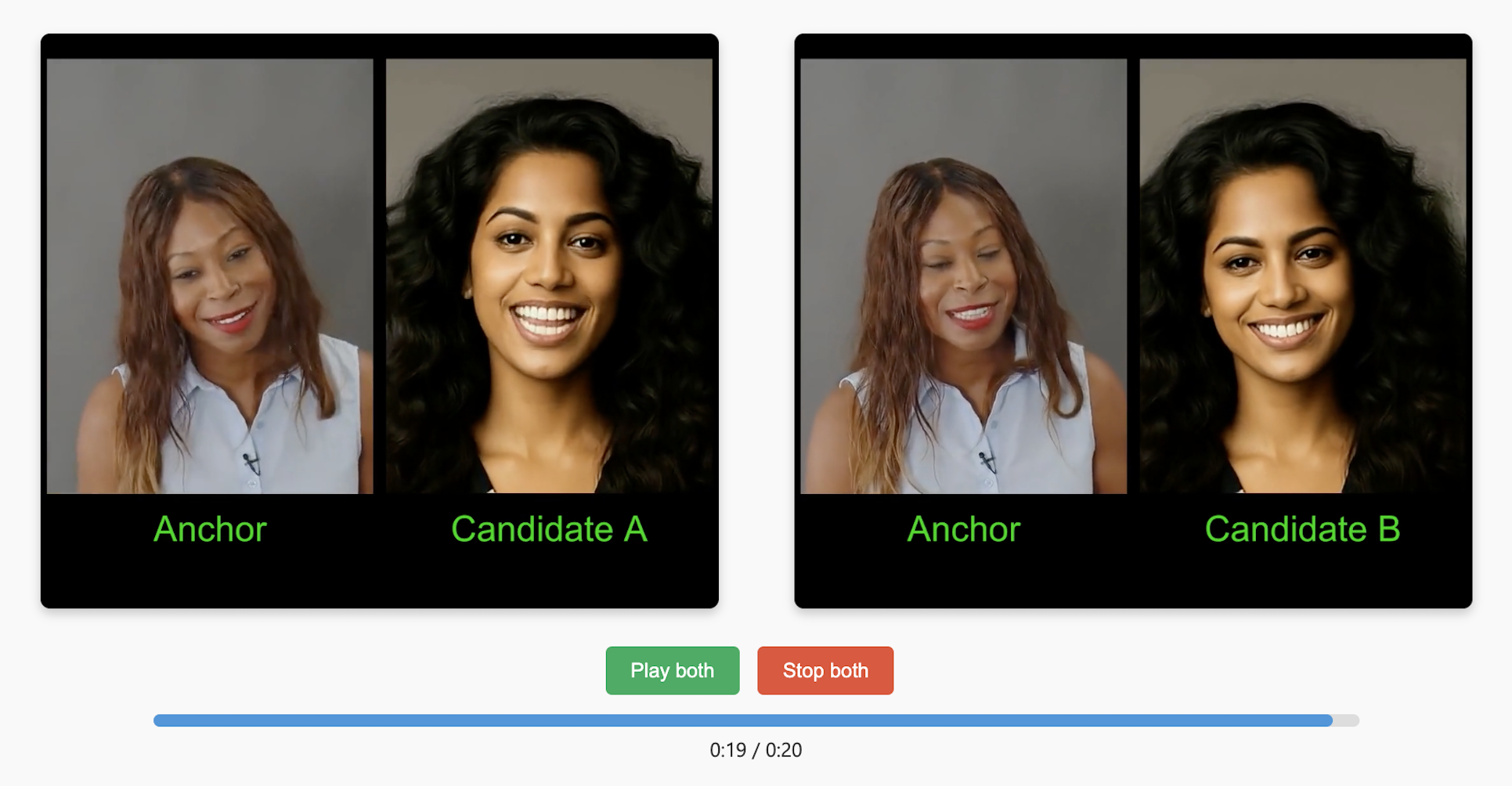}
    \caption{DFP stimuli.}
    \label{fig:enter-label}
\end{figure}
\textbf{Your task} is to assess which video,\textbf{ Candidate A or Candidate B} is a better dialogue partner for the \textbf{Anchor} along several dimensions. 

The \textbf{Anchor is the same in both videos. }You are not asked to provide ratings about Anchor directly, but we include Anchor, so that you have a visual reference and the context of the conversation. 
\begin{itemize}
    \item The left video includes Candidate A who is a dialogue partner for the Anchor.
    \item The right video includes Candidate B who is also a dialogue partner for Anchor.
    \item Candidates A and B behave differently while speaking and listening.
\end{itemize}

You can play each video individually and you can also watch the video clips simultaneously. 

Again, your task is not to assess the quality of Anchor; the Anchor video is provided only for context. All your judgements should be about whether Candidate A or Candidate B are better quality dialogue partners for Anchor. 

\textbf{Definitions} 
Anchor - Is a human participant. You are not to rate this video, it is there to provide you with a visual reference and context for the conversation. 
Candidate A - Candidate dialogue partner for Anchor. You will be asked to rate this video compared to Candidate B. 
Candidate B - Candidate dialogue partner for Anchor. You will be asked to rate this video compared to Candidate A. 

\textbf{Roles} 
Speaking - The state in which a person is generating words and content in the conversation. 
Listening - The state in which a person is taking in the words and content in the conversations. 
Cross-talk - A state in which both people are speaking. 

\textbf{Actions} 
Turn-taking - the process of moving from speaking to listening and also from listening to speaking.

\subsubsection {\textbf{Flagging}}
As part of this annotation task, you are given the chance to flag and skip certain clips if you encounter technical issues or safety concerns. Occasionally, you may see broken or incomplete clips or audio. Additionally, the generated avatars might make gestures that could be interpreted as inappropriate, even if that was not the intent of the AI platform. 

When flagging, consider the following guidelines. 

Flag as broken or incomplete audio/video: 
\begin{itemize}
    \item When audio cuts out or distorts during a key moment leaving parts of the conversation unclear.
    \item When video freezes or skips making it impossible to evaluate the gesture accurately.

\end{itemize}

Flag for safety concerns or inappropriateness: 
\begin{itemize}
    \item When an avatar may appear to make lewd/sexual gestures.
    \item When an avatar shows gestures mimicking violent actions.
    \item When an avatar makes a gesture that could be interpreted as a hate symbol or is associated with harmful ideologies.
\end{itemize}

To flag and skip an item, click the following icon to provide more information about the clip. Clicking the icon will route you to the following screen: 

Please provide your reason for flagging (select all that apply): 
$\square$ Audio is distorted 
$\square$ Audio is out of sync 
$\square$ Audio is cut out 
$\square$ Video freezes and/or skips 
$\square$ Avatar displays gestures that could be interpreted as lewd/sexual 
$\square$ Avatar shows violent gestures or actions 
$\square$ Avatar uses hate symbols or gestures associated with harmful ideologies 
$\square$ Other (Any other issue that impacts audio/video or makes the clip unsafe, uncomfortable, or inappropriate for evaluation): Please provide justification for why you are flagging this clip.

\subsubsection {\textbf{Item 1 (Lifelike)}}
\textbf{1. Overall, which candidate (A or B) was more life-like? }\textit{By “life-like,” we mean that the Candidate behaves in a way that looks, acts, or seems very similar to something that's real and alive. Visual behaviors can also include non-verbal actions and facial expressions that we use to communicate and express ourselves through our body language.}

$\circ$ Candidate A is much more lifelike 
$\circ$ Candidate A is slightly more lifelike 
$\circ$ Tie 
$\circ$ Candidate B is slightly more lifelike 
$\circ$ Candidate B is much more lifelike

\subsubsection {\textbf{Item 2 (Facial, Eye, and Lip Movement)}}
\textbf{2. Which candidate (A or B) had better facial expressions, eye movement, and lip movement? }

$\circ$ Candidate A appears to be much better
$\circ$ Candidate A appears to be slightly better
$\circ$ Tie 
$\circ$ Candidate B appears to be slightly better 
$\circ$ Candidate B appears to be much better

\subsubsection {\textbf{Item 3 (Clarity of Intent)}}
\textbf{3. Which candidate (A or B) most clearly demonstrated intent with their facial expressions? } \textit{By “intent,” we mean facial expressions and head gestures that are deliberate, non-repetitive, and convey a specific thought, idea, or emotion. }

$\circ$ Candidate A appears to be much more intentional 
$\circ$ Candidate A appears to be slightly more intentional  
$\circ$ Tie 
$\circ$ Candidate B appears to be slightly more intentional 
$\circ$ Candidate B appears to be much more intentional

\subsubsection {\textbf{Item 4 (Turn-Taking)}}
\textbf{4. Which candidate (A or B) appears to have better turn-taking behavior? }\textit{By ``turn-taking behavior,'' we refer to facial expressions and head gestures that signal the intention to speak or encourage a response from the dialogue partner. Examples of turn-taking behaviors are: a raised eyebrow, a subtle tilt of the head, slight opening of the mouth, a nod, and other nonverbal cues}. 

$\circ$ Candidate A appears to have much better turn-taking behavior 
$\circ$ Candidate A appears to have slightly better turn-taking behavior  
$\circ$ Tie 
$\circ$ Candidate B appears to have slightly better turn-taking behavior
$\circ$ Candidate B appears to have much better turn-taking behavior

\subsubsection {\textbf{Items 5 - 8 (Listening)}}
Now please rate the Candidates as they are listening:
\begin{figure}
    \centering
    \includegraphics[width=1\linewidth]{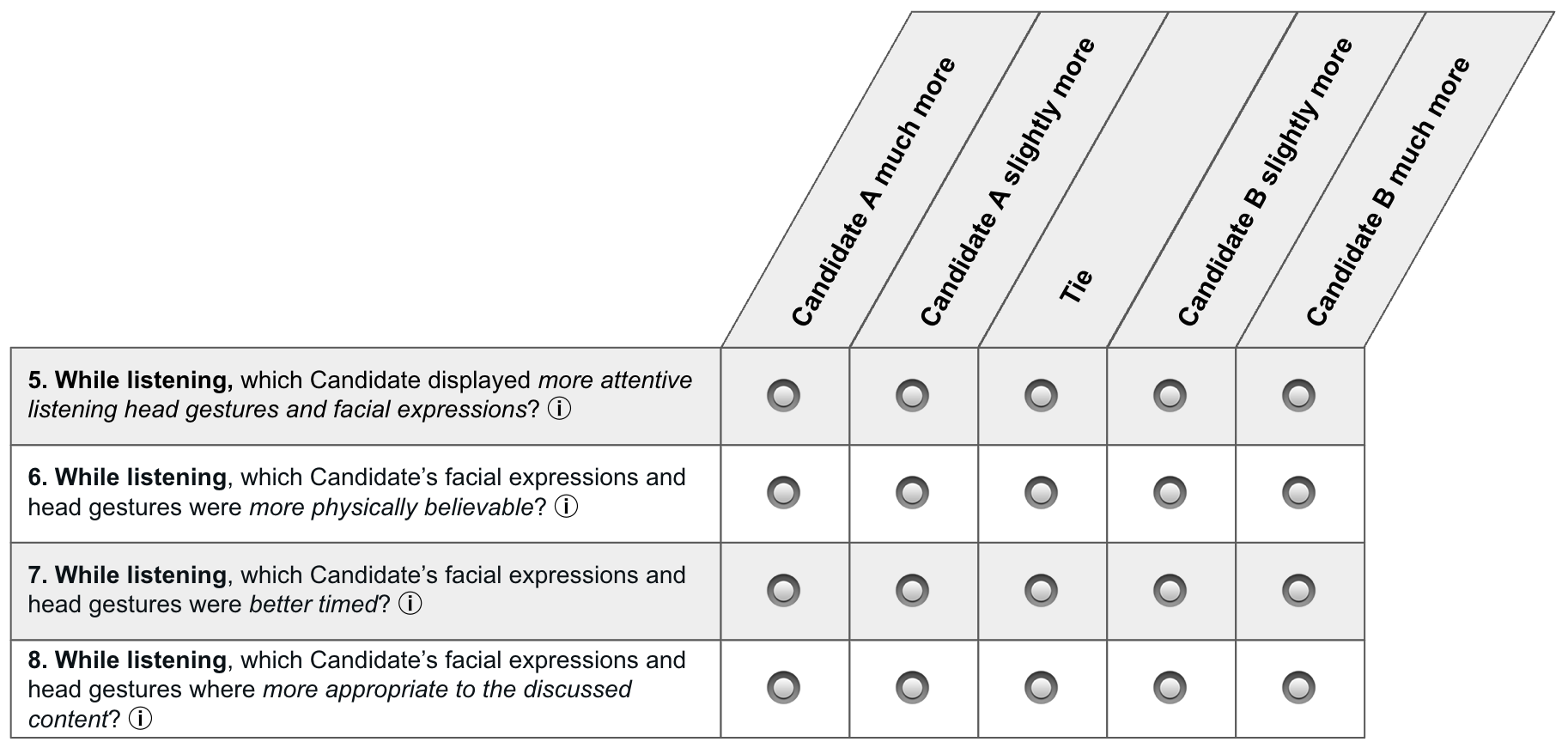}
    \caption{DFP listening items.}
    \label{fig:enter-label}
\end{figure}
5. \textbf{While listening, which Candidate displayed more attentive listening head gestures and facial expressions? }
Tooltip content\textbf{ } : \textit{By “attentive” we mean behaviors like leaning forward, nodding or shaking their head in agreement, visually indicating engagement and understanding.}
\textbf{6. }\textbf{While listening, which Candidate’s facial expressions and head gestures were more physically believable?}
Tooltip content\textbf{ } : \textit{By ``more physically believable'' we mean that the facial expressions and gestures of the Candidate are humanly possible and do not exhibit impossible movements that defy human physiology.}
\textbf{7. While listening, which Candidate’s facial expressions and head gestures were better timed?}
Tooltip content\textbf{ } : \textit{By “better timed” we mean that the Candidate's listening movements and reactions are synchronized with the Anchor's speech.}
\textbf{8. While listening, which Candidate’s facial expressions and head gestures where more appropriate to the discussed content?}
Tooltip content\textbf{ } : \textit{By “more appropriate to the content discussed” we mean that the Candidate's behaviors, such as head nods and body language, are consistent with the emotional tone and subject matter of the conversation. For example, if the Anchor is discussing a serious topic, the Candidate's behavior should reflect a corresponding level of gravity or concern, rather than appearing overly casual or dismissive.}

\subsubsection {\textbf{Items 8 - 10  (Speaking)}}
Now please rate the Candidates as they are speaking:

\begin{figure}
    \centering
    \includegraphics[width=1\linewidth]{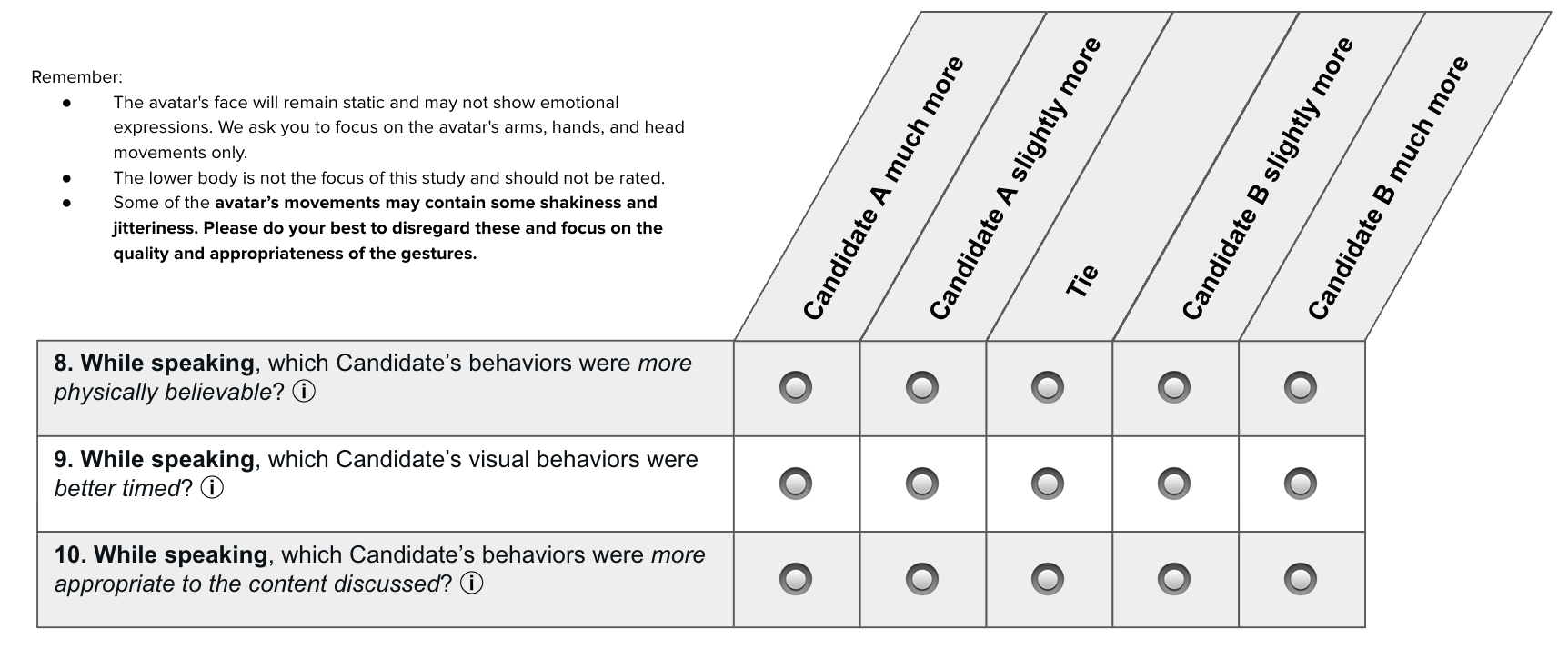}
    \caption{DFP speaking items.}
    \label{fig:enter-label}
\end{figure}
\textbf{9. While speaking, which Candidate’s facial expressions and head gestures were more physically believable?}
Tooltip content\textbf{ } : \textit{By ``more physically believable'' we mean that the head movements and facial expressions of the Candidate are humanly possible and do not exhibit impossible movements that defy human physiology.}
\textbf{10. While speaking, which Candidate’s facial expressions and head gestures were better timed?}
Tooltip content\textbf{ } : \textit{By “better timed” we mean that the Candidate's speaking movements (head nods and facial gestures) are synchronized well with the Candidate's speech.}
\textbf{11. While speaking, which Candidate’s facial expressions and head gestures were more appropriate to the content discussed?}
Tooltip content\textbf{ } : {\textit{By “more appropriate to the content discussed” we mean that the Candidate's behaviors, such as head nods, facial expression, and body language, are consistent with the emotional tone and subject matter of the conversation. For example, if the Candidate is discussing a serious topic, the Candidate's behavior should reflect a corresponding level of gravity or concern, rather than appearing overly casual or dismissive.}

\section{Activity-Based Prompts Examples}
\label{section:appendix_activity_based_examples}
\subsection{Language-Grounded Gesture Game}
\label{section:appendix_gesture_game}
In \Cref{tab:prompt_example_gesture_game}, we provide an example of language-grounded gesture game prompts.

\begin{table}[h]
    \centering
    \begin{tabular}{|m{7cm}|m{7cm}|}
        \toprule
        \textbf{Participant A} & \textbf{Participant B} \\ \midrule
        You're going to play a game with your partner. This game is going to test your acting skills! You are each given different lists of sentences below.
        
        Notice that one of the words is bolded in each sentence. When it is your turn you will read one of the sentences out loud AND act-out a corresponding gesture of the bolded word, emphasizing it. Your partner will react by responding with a word or phrase out-loud and making a corresponding expression or gesture.
        
        Example:
        You: (Sentence is “I’m feeling *confused*.”) You say it out loud while frowning and raising your eyebrows at the word “confused.”
        
        Partner: Responds saying “Why?” and acts with a concerned facial expression.
        
        Alternate with your partner until you’ve completed your list or the moderator asks you to move on.
        
        Your partner will go first.
        
        Here's your list:
        
        It's a *triangle*/It's *straight* line./It's a *square*.
        
        The traffic is *slow* due to construction.
        
        The ice cream is *cold* and delicious.
        
        The old man is *weak* and frail.
        
        I felt *helpless* as I watched the accident happen.
        
        He is *hesitant* to make a decision.
        
        I am *impatient* to finish this project.
        
        The car is *small* but fuel-efficient.
        
        I am *interested* in learning more about the topic.
        
        I am *irritated* by the constant interruptions.
        
        The alleyway is *narrow* and dark.
        
        The pillow is *soft* and comfortable. & You're going to play a game with your partner. This game is going to test your acting skills! You are each given different lists of sentences below.
        
        Notice that one of the words is bolded in each sentence. When it is your turn you will read one of the sentences out loud AND act-out a corresponding gesture of the bolded word, emphasizing it. Your partner will react by responding with a word or phrase out-loud and making a corresponding expression or gesture.
        
        Example:
        You: (Sentence is “I’m feeling *confused*.”) You say it out loud while frowning and raising your eyebrows at the word “confused.
        
        Partner: Responds saying “Why?” and acts with a concerned facial expression.
        
        Alternate with your partner until you’ve completed your list or the moderator asks you to move on.
        
        You will go first. Here's your list:
        
        The river is *wide* and deep.
        
        I am *nervous* about my presentation.
        
        I am *overwhelmed* by the amount of work.
        
        She is *heartbroken* over the breakup.
        
        Can you *pull* the curtain closed?
        
        The kitten is *playful* with its toy.
        
        I'm *furious* about the injustice.
        
        I feel *guilty* for not helping my friend.
        
        The athlete is *strong* and muscular.
        
        I am *happy* to see you.
        
        I am *frustrated* with the slow internet connection.
        
        The sun is *bright* in the sky.

  \\ \bottomrule
    \end{tabular}
    \caption{Example of the ``Language grounded gesture game'' prompt pair.} \label{tab:prompt_example_gesture_game}
\end{table}

\subsection{Collaborative Story-Telling Game}
\label{section:appendix_collaborative_story_telling}

\Cref{tab:prompt_example_collaborative_storytelling} shows an example of collaborative story-telling prompts.

\begin{table}[h]
    \centering
    \begin{tabular}{|m{7cm}|m{7cm}|}
        \toprule
        \textbf{Participant A} & \textbf{Participant B} \\ \midrule
        You're going to play a game with your partner. Together you will take turns weaving a fictional tale. It can be about whatever you like. 
        
        The rule is that you have to build the story together, alternating every sentence or two.
        
        For example, a story might look like the following:
        
        You: Once upon a time there was a little girl..
        Partner: ...who lived way way up high on a mountain in a cozy cabin in the sky...
        You: ...she had a friend who was an eagle who were a tiny little top-hat...
        
        The story can be about whatever you want! The point is to try to make it as interesting and fun as you can, creating it collaboratively. Be as expressive as you can with your voice and in your gestures.
        
        Your partner will begin with the words ``Once upon a time...'' & You're going to play a game with your partner. Together you will take turns weaving a fictional tale. It can be about whatever you like. 
        
        The rule is that you have to build the story together, alternating every sentence or two.
        
        For example, a story might look like the following:
        
        You: Once upon a time there was a little girl...
        Partner: ...who lived way way up high on a mountain in a cozy cabin in the sky...
        You: ...she had a friend who was an eagle who were a tiny little top-hat...
        
        The story can be about whatever you want! The point is to try to make it as interesting and fun as you can, creating it collaboratively. Be as expressive as you can with your voice and in your gestures.
        
        Whenever you're ready, begin with the four words, ``Once upon a time...''
  \\ \bottomrule
    \end{tabular}
    \caption{Example of the ``Collaborative Story-telling prompt'' pair for an interaction.} \label{tab:prompt_example_collaborative_storytelling}
\end{table}